\documentclass[lettersize,journal]{IEEEtran}
\usepackage{amsmath,amsfonts}
\usepackage{algorithmic}
\usepackage{algorithm}
\usepackage{array}
\usepackage[caption=false,font=normalsize,labelfont=sf,textfont=sf]{subfig}
\usepackage{textcomp}
\usepackage{stfloats}
\usepackage{url}
\usepackage{verbatim}
\usepackage{graphicx}
\usepackage{booktabs}
\usepackage{cite}
\usepackage{hyperref}
\usepackage{url}
\usepackage{booktabs}       
\usepackage{amsfonts}       
\usepackage{nicefrac}       
\usepackage{microtype}      
\usepackage{xcolor}         

\usepackage{amsmath,amsfonts,amssymb,amsthm,mathtools}
\usepackage{array}
\usepackage{textcomp}
\usepackage{stfloats}
\usepackage{float}
\usepackage{url}
\usepackage{verbatim}
\usepackage{graphicx}
\usepackage{booktabs}
\usepackage{enumitem}
\usepackage{hyperref}
\usepackage{subcaption}

\usepackage{threeparttable}
\usepackage{color}
\usepackage{xcolor}
\usepackage{amsmath}

\usepackage{tabularx}
\usepackage{bm}
\usepackage[capitalize,noabbrev]{cleveref}
\usepackage{bbm}

\usepackage{algorithm}
\usepackage{algorithmic}

\usepackage{multicol}
\usepackage{multirow}
\usepackage{array}
\newcolumntype{?}{!{\vrule width 1pt}}

\usepackage{balance}

\theoremstyle{plain}

\theoremstyle{definition}

\theoremstyle{remark}

\usepackage[textsize=tiny]{todonotes}

\definecolor{wine}{RGB}{114,47,55}

\usepackage{wrapfig}
\usepackage{makecell}

\begin{document}

\title{Continual Learning of Achieving Forgetting-free and Positive Knowledge Transfer}

\author{Zhi~Wang,~Zhongbin~Wu,~Yanni~Li,~Bing~Liu,~\IEEEmembership{Fellow,~IEEE,}~Guangxi~Li,~Yuping~Wang,{\color{black}~\IEEEmembership{Senior Member,~IEEE}}

\thanks{Z. Wang, Z. Wu, Y. Li, G. Li and Y. Wang are with the School of Computer Science and Technology, Xidian University, Xi'an, China, 710071. E-mails: zhiwang, wuzb@stu.xidian.edu.cn, yannili@mail.xidian.edu.cn, lgx, ywang@xidian.edu.cn.}
\thanks{B. Liu is with the Department of Computer Science, University of Illinois at Chicago. E-mail: liub@uic.edu.}
\thanks{Z. Wang, Z. Wu, Y. Li and B. Liu contribute equally to this work.}
\thanks{Y. Li and G. Li are co-corresponding authors.} 
}



\maketitle

\begin{abstract}
Existing research on continual learning (CL) of a sequence of tasks focuses mainly on dealing with catastrophic forgetting (CF)  to balance the learning plasticity of new tasks and the memory stability of old tasks. However, an ideal CL agent should not only be able to overcome CF, but also encourage positive forward and backward knowledge transfer (KT), i.e., using the learned knowledge from previous tasks for the new task learning (namely FKT), and improving the previous tasks' performance with the knowledge of the new task (namely BKT). To this end, this paper first models CL as an optimization problem in which each sequential learning task aims to achieve its optimal performance under the constraint that both FKT and BKT should be positive. It then proposes a novel Enhanced Task Continual Learning (ETCL)\footnote{\noindent The source code of ETCL is available at https://github.com/ETCLalg/ETCL} method, which achieves forgetting-free and positive KT. Furthermore, the bounds that can lead to negative FKT and BKT are estimated theoretically. Based on the bounds, a new strategy for online task similarity detection is also proposed to facilitate positive KT. To overcome CF, ETCL learns a set of task-specific binary masks to isolate a sparse sub-network for each task while preserving the performance of a dense network for the task. At the beginning of a new task learning, ETCL tries to align the new task's gradient with that of the sub-network of the previous most similar task to ensure positive FKT. By using a new bi-objective optimization strategy  and an orthogonal gradient projection method, ETCL updates only the weights of previous similar tasks at the classification layer to achieve positive BKT. Extensive evaluations demonstrate that the proposed ETCL markedly outperforms strong baselines on dissimilar, similar, and mixed task sequences. 
\end{abstract}


\section{Introduction}
\label{sec.intro}
\IEEEPARstart{C}{ontinual} learning (CL) using deep neural networks (DNNs) to learn a sequence of tasks is a challenging problem. Two key issues are overcoming \textit{Catastrophic Forgetting} (CF) \cite{3.1,3.2}, a phenomenon resulting in DNNs forgetting the knowledge learned in the past tasks upon learning new ones, and transferring knowledge across tasks, namely \textit{knowledge transfer} (KT). Existing approaches can be generally divided into network expansion methods and non-expansion methods. For example, LwF \cite{4.10}, CGN \cite{4.1}, DEN \cite{4.3}, APD \cite{4.7}, and BNS \cite{4.8} are representative expansion methods. These methods expand the network for each task to overcome CF, but they suffer from memory explosion with more tasks learned. While the basic idea of non-expansion methods is to constrain the gradient update of the network weights towards less harmful directions to protect the previously learned knowledge, e.g., orthogonal gradient projection (OG-based for short) methods, OWM \cite{7.2.1} and GPM \cite{7.2.4}, or to train task-specific masks to protect the knowledge learned from previous tasks to overcome CF, e.g., HAT \cite{9.2}, Piggyback \cite{n91} and SupSup \cite{9.12}. Among the non-expansion methods, the OG-based methods and mask-based methods have been shown to be effective in overcoming CF, yet are limited by scalability and KT ability. Unlike the above CL methods which learn all tasks with a single learner (model), recent research \cite{n98,n99,n101} suggests that an ensemble of multiple CL learners  brings huge benefits in balancing the learning plasticity of new tasks and memory stability of old tasks as compared with the CL methods by a single CL learner. Although a variety of the above representative CL methods have emerged,  most existing methods have a major limitation: they focus only on dealing with CF but ignore KT, while KT ability is a major goal of CL \cite{9.3}.

Task-incremental learning (TIL) is one of the important settings of CL \cite{1.2, 1.3}. The other one is \textit{class-incremental learning} (CIL) \cite{1.4}. 
The key difference between TIL and CIL is that in TIL, the task identifier is provided in both training and testing, while in CIL, the task identifier is only provided in training. The two settings are suitable for different types of applications. In the TIL setting,  when learning a new task $t$, naturally some previously learned tasks may be similar to $t$, and then the knowledge from them should be leveraged to learn $t$ better (namely  \textit{forward knowledge transfer}, FKT). Conversely, the learning of $t$ may also improve those similar previous tasks (namely  \textit{backward knowledge transfer}, BKT). Thus, an ideal TIL agent should not only be able to overcome CF but also to encourage positive FKT and BKT \cite{9.3}. Although some existing TIL methods perform KT, e.g., CAT \cite{9.3} using an additional sub-model, TRGP \cite{7.2.5} and CUBER \cite{n71} using layer-wise scaling matrices, WSN \cite{n95} jointly learning the model weights and task-adaptively binary masks, they still have some major shortcomings: only having limited FKT but no BKT or no guaranteed positive BKT (that does not cause forgetting again during BKT). 

In a word, the existing CL methods focus mainly on dealing with CF to balance the learning plasticity of new tasks and memory stability of old tasks \cite{n100,1.3}, which leaves a large gap from the ideal goal of TIL. To overcome the weaknesses of existing TIL methods and to approach the ideal goal of TIL, this research first models CL as an optimization problem with constraints, in which each sequential learning task aims to achieve its optimal performance  with both positive FKT and BKT.
By theoretically analyzing KT, 
this research introduces a new online task similarity metric and a novel CL mechanism that achieves both forgetting-free and positive FKT and/or BKT. A novel TIL method called ETCL (Enhanced Task Continual Learning) is also presented. Specifically, this paper makes the following contributions:
\begin{itemize}
\item It first theoretically studies the KT problem and gives the bounds that can lead to negative FKT and BKT. Based on this, a new criterion for online detection of similar tasks is proposed, which follows the real-world scenario without using any old task data.
\item It proposes a novel non-expansion TIL method ETCL, which has two novel ideas for learning each task:  
\begin{itemize}
\item ETCL learns the model weights and task-specific binary masks to isolate a sparse sub-network while preserving the performance
of a dense network of each task, which enables ETCL to eliminate CF and to learn more tasks with the same network size. ETCL also actively reuses the learned knowledge of previous tasks similar to the current task to achieve strong positive FKT by initially aligning gradients among similar tasks, and 
\item ETCL updates the weights of previous similar tasks only at the classification layer to achieve positive BKT by using a new bi-objective optimization strategy and an OG-based method to deal with CF during BKT. 
\end{itemize}
\end{itemize}

Extensive experiments show that the proposed ETCL not only overcomes CF better than existing state-of-the-art (SOTA) baselines on dissimilar, similar, and mixed task sequences, but also, perhaps more importantly, performs KT dramatically better than them when similar tasks are learned.

\section{Related Work}
This paper focuses on task-incremental/continual learning (TIL/TCL)  without network expansion.  For details on CIL, please refer to \cite{n300}.

An ideal TIL method requires effective learning of incremental tasks without CF and achieving positive FKT and/or BKT \cite{9.3}. Existing non-expansion TIL methods can be divided into the following categories: \textit{Regularization based} methods, e.g., EWC \cite{6.1} and  UCL \cite{6.4}, penalize modifications to important weights of old tasks through regularizations. \textit{Experience-replay based} methods, e.g., RES \cite{7.1.1}, 
iCaRL \cite{7.1.3} and A-GEM \cite{7.1.6} (an improved version of GEM \cite{7.1.2}), overcome CF by replaying the data  (either samples  of the real data or generated data) of old tasks for learning the new task. \textit{Orthogonal-gradient based} (OG-based) methods, e.g., OWM \cite{7.2.1}, OGD \cite{7.2.2}, GPM \cite{7.2.4}, RGO \cite{7.2.7}, etc., update the weights with gradients in the orthogonal directions of old tasks. 
\textit{Parameter isolation based} methods like HAT \cite{9.2}, Piggyback \cite{n91} and SupSup \cite{9.12} isolate a sub-network for each task. There are also reinforcement learning-based \cite{5.2.2,5.2.3,4.8}, soft mask-based \cite{konishi2023parameter}, and meta-learning based methods \cite{5.1.1,5.1.2,5.1.3}. These methods either learn all tasks with a single model, which has to compromise the performance of each task to obtain a shared solution or allocate a parameter subspace for each task to prevent mutual interference. But they are limited by the scalability and KT capability of the model.

\textbf{Mask-based Methods.} These methods belong to the parameter isolation category. By network quantization and pruning, Piggyback \cite{n91} learns a binary mask for each task on the network. The learned masks are applied to unmodified weights to provide good performance on a new task. HAT \cite{9.2} uses hard attention to learn pseudo-binary masks to protect old models to overcome forgetting. SupSup \cite{9.12} finds that supermasks within a randomly initialized network for each task avoid CF. WSN \cite{n95} jointly trains the model and task-adaptive sub-networks by reusing prior task parameters to achieve forgetting-free and FKT. Unlike our proposed ETCL, Piggyback, HAT and SupSup have no explicit KT mechanism, while WSN has only limited FKT and no BKT. 

\textbf{KT-based Methods.} Several early non-neural network based methods have done KT among similar tasks using KNN \cite{8.1.1}, regression \cite{8.1.2}, and naive Bayes \cite{8.1.4,5.1.2}, but they do not deal with CF. A few DNN based methods like CAT \cite{9.3}, TRGP \cite{7.2.5}, CUBER \cite{n71}, WSN \cite{n95} and ARI \cite{n93}  simultaneously deal with both CF and KT. CAT uses binary masks of neurons in HAT to achieve CF prevention, and employs a separate model to perform task similarity detection for KT. The OG-based method TRGP  first selects the most related old tasks within the `trust region' for the new task, and then reuses the frozen weights in layer-wise scaling matrices to jointly optimize the matrices and the model to achieve FKT. On the basis of TRGP, CUBER first analyzes the conditions under which updating the learned model of old tasks could lead to BKT. It then proposes a new method for FKT and BKT. By characterizing ``task-parameter relationships", ARI  models the similarities between the optimal weight spaces of tasks and exploits this to enable KT across tasks. Unlike our ETCL, the main weaknesses of CAT, TRGP, CUBER, and ARI are that they suffer from their weak KT mechanisms, i.e., limited FKT and some negative BKT leading to CF (see Table  \ref{KT1} in Sec.~V).

\textbf{Ensemble Model-based Methods.} The Ensemble Model is powerful in improving generalization but is under-explored in CL. Recent research    \cite{n98,n99,n101} suggests that an ensemble of multiple CL models can bring a large benefit in balancing the learning plasticity of new tasks and memory stability of old tasks as compared with the CL methods by a single model.  \cite{n98} and \cite{n99} theoretically analyze  the generalization error for learning plasticity and memory stability in CL. Then, inspired by the robust biological learning systems that process sequential experiences with multiple parallel compartments, two recent TIL methods, CoSCL \cite{n98} (Cooperation of Small Continual Learners) and CAF \cite{n99} (Continual learners with Active Forgetting), are proposed as general strategies for TIL. Extensive experimental results demonstrate that with a fixed parameter budget, CoSCL and CAF can improve a variety of representative CL methods. However, CoSCL and CAF do not explicitly deal with KT. Although their results show that CoSCL and CAF can improve the backward transfer (BWT) and/or forward transfer (FWT) performance of existing typical CL methods, they markedly underperform our ETCL method. 

The innovation of the proposed ETCL is three-fold: 1) \textit{More reasonable CL optimization goal}. The optimization goal of most existing CL methods is to balance the learning plasticity of new tasks and memory stability of old tasks with a compromised performance of each task, while ETCL aims to achieve both CF elimination and positive KT (including FKT and BKT) for each task in CL (see Eq.~(1)); 2) \textit{Stronger strategies for  preventing CF and performing KT.} Although the masks in ETCL are similar to those existing ones for dealing with CF,  the existing mask-based methods are limited by their scalability and KT capability. However, based on LTH (Lottery Ticket Hypothesis)\cite{n94}, ETCL effectively deals with the scalability issue. And ETCL aligns the new task's initial gradient with that of the sub-network of the most similar previous task to guide the learning of the new task to achieve a strong positive FKT. Furthermore, by using a new bi-objective optimization and an OG-based method, ETCL updates only the weights of previous similar tasks at the classification layer to achieve positive BKT; 3) \textit{Better theoretical bounds}. For learning plasticity and memory stability in CL, CoSCL \cite{n98} and CAF \cite{n99} theoretically analyze the generalization errors, which can be uniformly upper-bounded by three items: (1) discrepancy between task distributions, (2) flatness of the loss landscape and (3) cover of the parameter space. But our theoretical bounds are devoted to deriving what is necessary and sufficient to achieve positive FWT and BWT, with  FWT and BWT uniformly upper-bounded by the three items: (1) empirical error of the task, (2) discrepancy between task distributions and (3) the absolute difference of the empirical results of tasks (see Eq.~(\ref{neq6})). 

\section{Exploration of Positive KT}
\subsection{Formulation of TIL and KT}

\textbf{Task Incremental Learning (TIL).} Let $\mathbf{X}$ be the input space, $\mathbf{Y}$ the label space of $\mathbf{X}$, and  $\mathbb{T}=\{t\}^T_{t=1}$ the tasks,  which are learned sequentially. Each task has a training dataset with its task descriptor $t$, $\mathbb{D}_t = \{((\bm{x}_{t,i},t),\bm{y}_{t,i})\}^{N_t}_{i=1}$, where $\bm{x}_{t,i} \in \mathbf{X}$ is the input data and $\bm{y}_{t,i} \in \mathbf{Y}_t \subset \mathbf{Y}$ is its class label. The goal of TIL is to construct a predictor $h$: $\mathbf{X} \times \mathbb{T} \rightarrow \mathbf{Y}$ to predict the class label $\hat{\bm{y}}_{t,i} \in \mathbf{Y}_t$ for $(\hat{\bm{x}}_i,t)$ (a given test instance $\hat{\bm{x}}_i$ from task $t$).

\textbf{Knowledge Transfer} \textbf{(KT}). Let $\mathbb{T}_{sim}$ / $\mathbb{T}_{dis}$  be a set of similar/dissimilar tasks of the current task $t$ ($\mathbb{T}_{sim}, \mathbb{T}_{dis} \subseteq \mathbb{T}$, $\mathbb{T}_{dis}=\mathbb{T}-\mathbb{T}_{sim}$). A TIL learner should transfer the knowledge learned in the past \textbf{forward} and leverages it to learn $t$ better (i.e., FKT), and additionally, the learning of $t$ should also improve the previously learned tasks in $\mathbb{T}_{sim}$ by \textbf{backward} knowledge transfer (i.e., BKT) under the assumption that the system has the ability to detect tasks' similarity online when a new task $t$ comes.

Ke et al. \cite{9.3} suggested that \textit{an ideal TIL model/method should satisfy two basic requirements: (1) overcoming CF and (2) performing forward and/or backward KT to improve the performance of the TIL model across similar tasks.} Thus, for a supervised TIL model $h(\mathbf{X}; \mathbf{W})$ of a CNN parameterized by its  weights $\textbf{W}$, 
we introduce an ideal TIL optimization objective: pursuing the optimal performance of each task while ensuring forgetting-free and positive KT across tasks (if similar tasks exist) by a single CL learner. We formalize this idea as follows: 
\begin{equation}
\label{iTIL}
\small
\begin{matrix}
\textbf{W}^*=\underset{\textbf{W}}{\textup{argmin}}\frac{1}{N_t}\sum _{i=1}^{N_t}\mathcal{L}(h(\bm{x}_{t,i};\textbf{W}),\bm{y}_{t,i}), t \in [1,T] \\
s.t. \;\; FWT \geq 0, BWT \geq 0
\end{matrix}
\end{equation}
where $\mathcal{L}(.)$ is the classification loss of task $t$, such as cross-entropy loss or mean square error loss. BWT (Backward Transfer), also called \textit{forgetting rate}, and FWT (Forward Transfer) are performance metrics shown in Eq.~(\ref{metrics}) to measure BKT and FKT for sequential learning tasks, respectively.

\subsection{Exploration of Positive FKT and BKT}

We first introduce some definitions and then explore what factors cause positive or negative FKT/BKT in TIL.

\noindent \textbf{Forward Negative KT Margin (FNM).} Given two similar tasks $i$ and $t$ ($i<t$) in the TIL setting, let $\epsilon_{t}(\cdot)$ be the test error of task $t$. $g(i, t)$ denotes that task $t$ performs its learning with the help of the knowledge of the previous similar task $i$, and $g(t)$ otherwise. Then, negative FKT happens when $\epsilon_{t}(g(i,t))>\epsilon_{t}(g(t))$.

\noindent \textbf{Backward Negative KT Margin (BNM).} Given two similar tasks $i$ and $t$ ($i<t$) in the TIL setting, let $\epsilon_{i}(g(i))$ be the test error of task $i$ before task $t$ learning, and $\epsilon_{i}'(g(i,t))$ be the test error of task $i$ after task $t$ is learned, then negative BKT happens when $\epsilon_{i}'(g(i,t)) > \epsilon_{i}(g(i))$. 

Thus, the negative FKT and BKT margins are  defined as   
\vspace{-1mm}
\begin{equation}
\label{F-B NKM}
\begin{array}{l}
\begin{aligned}
&FNM=\epsilon_{t}(g(i,t)) - \epsilon_{t}(g(t))\\
&BNM=\epsilon_{i}'(g(i,t)) - \epsilon_{i}(g(i))
\end{aligned}
\end{array}
\end{equation}
\noindent \textbf{Proposed FNM/BNM-based KT Metrics.} From Eq.~(\ref{F-B NKM}), it is clear that the degree of forward/backward negative KT can be evaluated by FNM/BNM, and negative KT occurs when the FNM/BNM is positive. As $\epsilon_{t}$ is inversely proportional to the test accuracy of task $t$ (denoted by $A_t$) and FNM/BNM may not always be computable, the degrees of FKT and BKT across similar tasks $i$ and $t$ in TIL (denoted by FWT and BWT, respectively) can be evaluated as follows:  

\begin{equation}
\label{FWT-BWT}
\begin{array}{l}
\begin{aligned}
&FWT=A_t(g(i,t)) - A_t(g(t))\\
&BWT=A_{i}'(g(i,t)) - A_{i}(g(i))
\end{aligned}
\end{array}
\end{equation}
where $A'_{i}$ is the test accuracy of task $i$ after task $t$ is learned. The greater the positive/negative value of FWT/BWT, the greater the quantity of positive/negative FWT/BWT.

\noindent \textbf{Theoretical Bound for KT.} Given two similar tasks $i$ and $t$ ($i<t$), we now analyze the theoretical bounds for KT in TIL so as to investigate the factors that lead to forward or backward positive/negative KT between them. 

Recall that the mapping function of a DNN for classification in TIL is the hypothesis or predictor $h: \mathbf{X} \times \mathbb{T} \rightarrow \mathbf{Y}$. According to the test data distribution $\mathcal{D}'_t$ of task $t$,  the test error showing that the hypothesis $h$ disagrees with its labeling function $l_t$ (which can also be a hypothesis) is defined as
\begin{equation}
\label{eq5}
\epsilon_t(h,l_t)=\mathbb{E}_{\mathbf{x} \sim  \mathcal{D}'_{t}} \left [|h(\mathbf{x}, t)-l_t(\mathbf{x})|\right ], \mathbf{x} \in \mathbf{X}
\end{equation}
For simplicity, we also denote the \textit{risk} or \textit{error} of hypothesis $h$ on task $t$ by $\epsilon_t(h)$ ($=\epsilon_t(h,l_t)$). Let the divergence of the test data distributions of $\mathcal{D}'_{i}$ and $\mathcal{D}'_{t}$ be $d(\mathcal{D}'_{i},\mathcal{D}'_t)$ of tasks $i$ and $t$, where $d(.)$ can be calculated by a similarity/distance metric. Then we can derive and prove the following theorem.

\textbf{Theorem 1.} The theoretical bounds for FWT and BWT of tasks $i$ and $t$ ($i < t$)  in TIL are given by
\begin{equation}
\label{neq6}
\small
\begin{array}{l}
\begin{aligned}     
&\epsilon_{t}(h)\leq \epsilon_{i}(h)+d(\mathcal{D}'_{i},\mathcal{D}'_{t}) + \textup{min}\{\mathbb{E}_{\mathbf{x} \sim  \mathcal{D}'_{i}}(\mathbb{S}),\mathbb{E}_{\mathbf{x} \sim  \mathcal{D}'_{t}}(\mathbb{S})\} \\
&\epsilon'_{i}(h)\leq \epsilon_{t}(h)+d(\mathcal{D}'_{i},\mathcal{D}'_{t}) + \textup{min}\{\mathbb{E}_{\mathbf{x} \sim  \mathcal{D}'_{i}}(\mathbb{S}),\mathbb{E}_{\mathbf{x} \sim  \mathcal{D}'_{t}}(\mathbb{S})\}
\end{aligned} 
\end{array}
\end{equation}
where $\epsilon_{i}(h)$ ($\epsilon'_{i}(h)$) is the test error of task $i$ before (after) task $t$ learning (is learned), and  $\mathbb{S}=|l_{i}(\mathbf{x})-l_{t}(\mathbf{x})|$ represents the absolute difference between the test results on data $\mathbf{x}$ of tasks $i$ and $t$.  The proof is given in Appendix {\color{black} A}.

From Theorem 1, we observe the following: (1) In the forward/backward KT process, two additional losses are introduced, (i) the loss due to the divergence of the test data distributions of tasks $i$ and $t$ (the second term on the right side of Eq.~(\ref{neq6})), and (ii) the loss due to the difference of data classification results of the tasks (the third term). (2) \textit{It is clear that the necessary and sufficient conditions for the elimination of negative forward/backward KT are that the errors introduced by the above two terms should be zero.} (3) The less the two additional losses above are, the greater the gain of FWT or BWT will be. 

It is worth noting that Eq.~(\ref{neq6}) may not always be computable in practice as it is impossible to get the test data during model training. Thus, with the assumption that the training and test data are i.i.d (independently identically distributed), we can employ the empirical errors $\hat{\epsilon}_{t}(h)$ and $\hat{\epsilon}_{i}(h)$/$\hat{\epsilon}_{i}'(h)$ to approximate $\epsilon_{t}(h)$ and $\epsilon_{i}(h)$/$\epsilon'_{i}(h)$ using the training data.

Moreover, related KT researches \cite{n70, n72, n74, 5.1.2, 7.2.5, n71} have proven the following Theorem 2.

\textbf{Theorem 2.} Low similarity or negatively correlated tasks will result in negative KT. Only high similarity tasks or positively correlated tasks can achieve positive KT.

\begin{figure*}[h]
    \centering
    \includegraphics[scale=0.6]{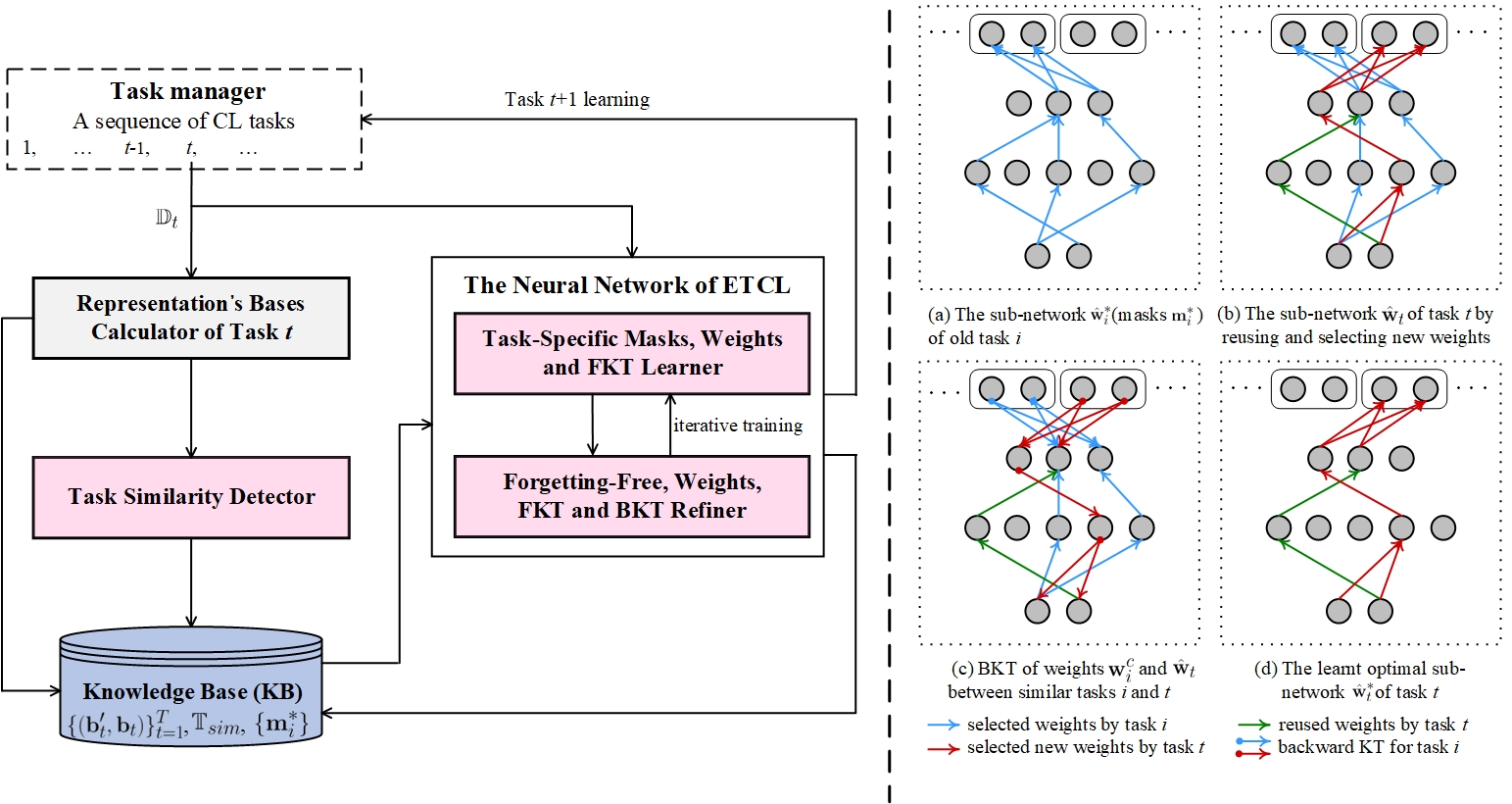}
    \caption{The architecture and pipeline of the proposed ETCL (on the left), where the proposed new techniques are embedded in the pink  components. To the right of the dotted line separation, let $i$ and $t$ be two tasks similar to each other ($i<t$). (a) The selected sub-network $\hat{\mathbf{w}}_i^*$ (indicated by masks $\mathbf{m}^*_{i}$) of the previous task $i$ represented with blue arrows. (b) {\color{black}The selected initial sub-network $\hat{\mathbf{w}}_t$ (masks $\mathbf{m}_{t}$)} of task $t$ represented by the selected new or unused weights by previous tasks (red arrows) and reused weights of previous similar task $i$ (green arrows) leading to automatic forward KT. (c) During task $t$ training, the weights corresponding to $\mathbf{m}_{t}$ are constantly updated and optimized. With the bi-objective optimization of the classification layer, the knowledge from task $t$ is backward transferred to previous task $i$ (those arrows with a circular point at the tails). (d) The optimized sub-network {\color{black}$\hat{\mathbf{w}}_t^*$ (masks $\mathbf{m}^*_{t}$)} of task $t$ with newly selected and reused weights.}
    \label{fig1}
    \vspace{-3mm}
\end{figure*}

\section{ETCL: Continual Learning of achieving Forgetting-free and Positive FKT and BKT}
To achieve the ideal TIL goal (Eq.~(\ref{iTIL})), the proposed ETCL introduces three new mechanisms: 1) task-specific binary masks to isolate a sparse sub-network (which also preserves the performance of a dense network) for each task to eliminate CF and to learn more sequential tasks, 2) optimized masks with the initial gradient alignment and bi-objective optimization for positive FKT and BKT, and 3) an online task similarity detector without using any old task data. The proposed mechanisms are performed by the three pink components of ETCL shown in Figure \ref{fig1}(a), whose details are presented below.
 
\subsection{Forgetting-free CL with Task-specific Masks and Orthogonal Weight Updating}
Lottery Ticket Hypothesis (LTH) \cite{n94} demonstrates the existence of sparse sub-networks, which preserves the performance of a dense network. Inspired by the LTH, we propose a new type of mask that sequentially learns and selects a sparse optimal sub-network in the whole network for each task, which achieves forgetting-free and overcomes the limited scalability of existing mask-based methods \cite{9.2,9.12,n91}.

As the DNNs are often over-parameterized to allow room for learning new tasks, we can find sub-networks that achieve on-par or even better performance. Given the   parameters/weights \textbf{W} of a DNN, a set of binary masks  (denoted by $\mathbf{m}^*_t$) corresponding to an optimal sub-network for task $t$ with a value less than the model capacity $C$ is learned as follows:

\vspace{-5mm}
\begin{equation}
\begin{split}
\label{nm}
\mathbf{m}^*_t=\underset{\mathbf{m}_t \in \{0,1\}^{|\textbf{W}|}}{\textup{argmin}} \; \frac{1}{N_t}\sum _{i=1}^{N_t} \{\mathcal{L}(h(\bm{x}_{t,i};\textbf{W}\odot \mathbf{m}_t),\bm{y}_{t,i})\\ 
- \mathcal{L}(h(\bm{x}_{t,i};\textbf{W}),\bm{y}_{t,i}) \}, \;
\textup{s.t.} \;|\mathbf{m}^*_t| << C = |\textbf{W}|
\end{split}
\end{equation}
where $\textbf{m}_t$ is a set of un-optimized masks for the sub-network of task $t$, and $\odot$ means element-wise multiplications of two matrices with the same dimensions. In what follows, we describe how to learn $\textbf{m}^*_t$ and at the same time how to minimize the loss of  task $t$.

Let each weight in a CNN be associated with a learnable parameter, called \textit{weight score} $\mathbf{s}$, which numerically determines the importance of the weight to task $t$. That is, the larger the weight score $\mathbf{s}$, the more important the weight is to task $t$. Based on LTH, we find a sparse sub-network $\hat{\mathbf{w}}_t$, i.e., we select a small set of weights to be activated by reusing weights of the prior sub-networks and also selecting those weights that have not been chosen/used by previous tasks, and assign them to task $t$ as its sub-network (see Figure \ref{fig1}(b)). This has two benefits: (1) each learning task has its own independent weight sub-network in the whole weight space resulting in no forgetting, and (2) the sub-network requires less capacity than the full network avoiding network capacity exploding as the number of learning tasks increases. Thus, we find $\hat{\mathbf{w}}_t$ by selecting $c\%$ of the network weights with the highest weight scores $\mathbf{S}_{s}=\{\textbf{s}$\}, where $c$ is the target layer-wise capacity ratio. The selection of weights is represented by the task-specific {\color{black}binary weight masks  $\mathbf{m}_t$} where a value of 1 in the mask denotes that the weight is selected during the forward pass and a value of 0 otherwise. Formally, $\mathbf{m}_t$ is obtained by applying an indicator function $\mathbbm{1}_c$ on $\mathbf{s}$ where $\mathbbm{1}_c(\mathbf{s})=1$ if $\mathbf{s}$ belongs to the top-c\% scores and otherwise $\mathbbm{1}_c(\mathbf{s})=0$. Thus, for the sub-network of task $t$, we can obtain  $\hat{\mathbf{w}}_t=\mathbf{W} \odot \mathbf{m}_t$.

To jointly learn the model weights 
and the binary masks $\mathbf{m}_t$ of task $t$, given the cross entropy loss $\mathcal{L}(.)$, we optimize $\mathbf{W}$ and $\mathbf{S_s}$\footnote{As the gradient of $\mathbf{s} \in \mathbf{S_s}$ based on the indicator function always has the value of 0, its updating employs  Straight-through Estimator \cite{n96,n97} to deal with the issue.} to obtain its optimal sub-network $\hat{\mathbf{w}}^*_t$ as follows: 

\begin{equation}
\label{nw}
\hat{\mathbf{w}}^*_t=\underset{\mathbf{W},\mathbf{S_s}}{\textup{argmin}} \; \mathcal{L}(\mathbf{W} \odot \mathbf{m}_t; \mathcal{D}_t), t \in [1,T] 
\end{equation}
where $\mathbf{W}$ and $\mathbf{S_s}$  are updated by the following equations:
\begin{equation}
\label{nws1}
\mathbf{W} \leftarrow \mathbf{W}-\eta \left (  \partial \mathcal{L}/\partial \mathbf{W}\odot (\mathbbm{I}-\mathbf{M}_{t-1})\right )
\end{equation}
\begin{equation}
\label{nws2}
\mathbf{S_s} \leftarrow \mathbf{S_s}-\eta \left ( \partial \mathcal{L}/\partial \mathbf{S_s} \right )
\end{equation}
where $\mathbbm{I}$ is a set of the all-ones matrix with the same dimensions as matrix $\mathbf{M}_{t-1}$, $\eta$ is the learning rate, and $\mathbf{M}_{t-1}=\{\mathbf{m}^*_i\}^{t-1}_{i=1}$, which  is the accumulated binary masks of the previously learned $(t-1)$ tasks in learning task $t$. 

Note that to ensure forgetting-free along with improving  model classification performance, ETCL introduces the following two new mechanisms: (1) After learning task $t$, its $\hat{\mathbf{w}}^*_t$ is frozen, i.e., the gradients of the weights corresponding to the masks  $\mathbf{m}^*_t$ of task $t$ will be set to zero in future tasks learning to ensure that each task has its own independent sub-network $\hat{\mathbf{w}}^*_t$, which is inherently immune to CF as each sub-network does not interfere with the other sub-networks; (2)  The weights from the classification layer of the model determine the classification of task $t$, while the weights between $\mathbf{M}_{t-1}$ and $\hat{\mathbf{w}}^*_t$ in the layer inevitably overlap with each other. To achieve enhanced CF resistance and improved classification performance, when learning task $t$, we only update {\color{black}the weights of the classification layer 
of the previous similar tasks by borrowing an OG-based method GPM \cite{7.2.4} to overcome CF (see Figure 1(c)).}

\subsection{Positive FKT and BKT CL with 
Gradient Alignment and Bi-objective Optimization}

As the existing mask-based methods like HAT, SupSup, and Piggyback are mainly designed to overcome CF, 
additional KT mechanisms are needed. Before going further, we note that anti-forgetting and positive KT are inherently contradictory because the goal of the former is to ensure that knowledge of the learned tasks does not interfere with each other, while the goal of the latter is to encourage to reuse the knowledge of one task A to help learn another task B (FKT) and in learning B to improve A at the same time (BKT). We would like to learn a novel mask that is versatile in the sense that it enables {\color{black}both forgetting-free and positive KT (including positive FKT and BKT)} effectively and efficiently. To this end, we propose the following strategies:

\textbf{Str-1: Decoupling Problems.} 
{\color{black}First,} in learning each task, we decouple the learning of weights and the learning of its weight scores $\mathbf{S_s}$ corresponding to its masks shown {\color{black}in Eqs.~(\ref{nws1}) and (\ref{nws2})}, which yields two benefits: (1) as each task $t$ learns its own sub-network (i.e., $\hat{\mathbf{w}}^*_t$, $t \in [1,T]$) that is independent of  other task sub-networks, the forgetting-free is achieved.  (2) When learning a new task $t$, as its sub-network $\hat{\mathbf{w}}^*_t$ can reuse some weights that have been used by previous ($t-1$) task sub-networks (see $\mathbf{W} \odot \mathbf{m}_t$ in Eq.~(\ref{nw}) and Figure 1(b)), FKT is achieved. 

\textbf{Str-2: Aligning Initial Gradients.} Although the above Str-1 can naturally perform FKT, it has two issues: (1) The FKT is often sub-optimal due to random blind searching for useful previous knowledge, and (2) the searching is also very inefficient for complex backbone architectures. To address the issues, we  present an initial gradient alignment strategy to maximize FKT from the previously learned tasks that are similar to $t$ and to accelerate the convergence of performing FKT of the task. Specifically, when learning task $t$, if $\mathbb{T}_{sim} \neq \emptyset$ (see the next subsection for the calculation of $\mathbb{T}_{sim}$), we first feed the training dataset $\mathbb{D}_t$ of task $t$ into mask $\mathbf{m}^*_i$ of task $i$ to obtain its corresponding weight scores $\mathbf{S}^i_s$, where the task $i$ ($\in \mathbb{T}_{sim}$) is the most similar previous task, and then feed $\mathbb{D}_t$ to its model to obtain the initial weight scores $\mathbf{S}^t_s$ of task $t$, and then perform the following initial $\mathbf{S}^t_s$'s gradient alignment as follows: 

\begin{equation}
\label{G-Align}
\partial \mathcal{L}/\partial \mathbf{S}^t_s \leftarrow \left ( \partial \mathcal{L}/\partial \mathbf{S}^t_s + \partial \mathcal{L}/\partial \mathbf{S}^i_s \right ), \; i<t, i \in \mathbb{T}_{sim}
\end{equation}
where $\mathcal{L}$ is $\mathcal{L}_{sim}$ (see Eq. (\ref{Lsim})), and  ($\partial \mathcal{L}/\partial \mathbf{S}^t_s + \partial \mathcal{L}/\partial \mathbf{S}^i_s$) means that the gradient $\partial \mathcal{L}/\partial \mathbf{S}^t_s$ aligns to gradient $\partial \mathcal{L}/\partial \mathbf{S}^i_s$ resulting in $\mathbf{S}^t_s$ approaching $\mathbf{S}^i_s$, so as to overcome the above issues. Note that the initial gradient alignment only performs once at the beginning of task $t$ learning. {\color{black}With Eqs.~(\ref{nws1})- (\ref{nws2}), the continued model update with task $t$ gradient will eventually lead to its task-specific weights and weight scores.} 


\textbf{Str-3: Bi-objective Optimisation.} To achieve positive BKT with maximal transfer and minimal interference, unlike all existing KT-based methods, we propose a strategy that performs orthogonal gradient weights updating across previous similar tasks of task $t$ only in the classification layer.\footnote{ETCL adopts multiple classification heads, that is, one classification head is assigned to a learning task.} {\color{black}That is, in learning a new task $t$, if ETCL finds $\mathbb{T}_{sim} \neq \emptyset$, it would update the weights of previous similar tasks in $\mathbb{T}_{sim}$ in the classification layer with the following bi-objective training loss $\mathcal{L}_{sim}$ and an OG-based method GPM to achieve BKT  (see Figure 1(c)). The OG-based method, which is known for effective CF prevention, helps overcome CF that may be caused by BKT. 
Note that ETCL uses the cross-entropy loss $\mathcal{L}(.)$ for its training in Eqs.~(\ref{nws1}) and  (\ref{nws2}).
\begin{equation}
\label{Lsim}
\small
\mathcal{L}_{sim} = -\frac{1}{N_t} \sum_{i=1}^{N_t} \bm{y}_{t,i} \log(\hat{\bm{y}}_{t,i})+ \frac{1}{N_1} \sum_{j=1}^{N_1} \left ( 1- \frac{\mathbf{w}^c_j \cdot \mathbf{w}^c_t}{|\mathbf{w}^c_j| |\mathbf{w}^c_t|} \right )
\end{equation}
where $\bm{y}_{t,i}$ is the ground-truth label for a given test instance $\hat{\bm{x_i}}$ from task $t$, while $\hat{\bm{y}}_{t,i}$ is the predicted label of $(\hat{\bm{x_i}}, t)$.  $\mathbf{w}^c_j$ and $\mathbf{w}^c_t$ are the weights of tasks $j$ and $t$ in the classification layer, respectively, $j \in \mathbb{T}_{sim}$  and $N_1=|\mathbb{T}_{sim}|$. The first term in the formula is the cross-entropy loss for classification, and the second term aims to make the similar tasks have similar weights, i.e., the more similar the tasks, the smaller the difference of their weights.


The algorithm for the proposed ETCL is summarized as \textbf{Algorithm 1 ETCL}.  

\begin{algorithm}[h]
   \caption{ETCL}
   \label{alg:etcl}
   \leftline{\textbf{Input:} Training datasets $\{\mathbb{D}_t\}^T_{t=1}$; the model weights $\mathbf{W}$;} 
   \leftline{the layer-wise capacity $c$;}
\begin{algorithmic}[1]
   \STATE Randomly initialize $\mathbf{W}$ and $\mathbf{S}_{s}$;
   \FOR{{\bfseries each} task $t \in [1, T]$}
   \FOR{{\bfseries each} batch data $\mathbf{d}_{t} \subset \mathbb{D}_t$}
   \STATE Obtain mask $\mathbf{m}_t$ of top-$c$\% scores $\mathbf{S}_{s}$ at each layer
   \IF{$t$ == 1 or $\mathbb{T}_{sim}$ == $\emptyset$}
   \STATE Compute Eq. (\ref{nw});
   \STATE Update $\mathbf{W}$ and $\mathbf{S_s}$ by Eq. (\ref{nws1}) and Eq. (\ref{nws2});
   \ELSE
   \STATE Compute $\mathcal{L}_{sim}$ by Eq. (\ref{Lsim}); 
   \STATE Compute Eq. (\ref{G-Align}) once; // Gradient alignment
   \STATE Compute Eq. (\ref{nw});
   \STATE Update $\mathbf{W}$ and $\mathbf{S_s}$ by Eq. (\ref{nws1}) and Eq. (\ref{nws2});
   \FOR{{\bfseries each} task $i \in \mathbb{T}_{sim}$}
   \STATE update $\mathbf{w}^c_i$ with $\mathcal{L}_{sim}$ and the method GPM;
   \ENDFOR
   \ENDIF
   \ENDFOR
   \STATE $\mathbf{M}_{t} \leftarrow \mathbf{M}_{t-1} \cup \mathbf{m}_t^*$; // Accumulated binary masks
   \ENDFOR
\end{algorithmic}
\end{algorithm}

\vspace{-3mm}
\subsection{Online Task Similarity Detection}
Theorems 1 and 2 reveal: (1) an accurate measure of task similarity is essential for positive KT, (2) to ensure positive KT, the divergence of data distributions and the difference of the data classification results must be considered together, and (3) it is possible for only similar/positive related tasks to achieve positive KT. 
Meanwhile, we note that after the model has learned ($t-1$) tasks, all the knowledge learned is recorded in the weight matrix $\mathbf{W}$ of the model. We also observed in experiments that for a new task $t$ learning, if $t$ gains improved performance on $\mathbf{W}$ compared with the performance on $\mathbf{W}$ with randomly initialized weights and no training, it indicates that there is shared knowledge in $\mathbf{W}$ for task $t$, i.e., there must be previous similar tasks to $t$; otherwise here is no similar previous task.

As the embeddings/representations of input data represent the input data and determine their  test results, following the necessary and sufficient conditions for guaranteeing positive KT as revealed by Theorems 1 and 2, we propose a new online task similarity detection criteria based only on the distance of the representation bases of input data without using any previous task data. Specifically, given a model denoted by $model_{ori}$ with the same architecture as the ETCL model (denoted by $model_{CL}$) and randomly initialized weights without training, using some training data $\mathbb{D}'_t$ randomly sampled with a rate of 5\% from $\mathbb{D}_t$ of task $t$, ETCL performs the following steps:  

\textbf{Step 1:} Before starting to learn a new task $t$, feeding $\mathbb{D}'_t$ into $model_{ori}$ and $model_{CL}$, respectively, so as to obtain their bases $\mathbf{b}'_t$ and $\mathbf{b}_t$ of the representations of task $t$ corresponding to $model_{ori}$ and $model_{CL}$, where the bases $\mathbf{b}'_t$ and $\mathbf{b}_t$ can be calculated by the component of ETCL \textbf{Representation Bases Calculator of Task $t$} (see Figure 1(a) and below).

\textbf{Step 2:} Calculating the distances between a previously learned task $i$ ($i\in [1,t-1]$) and new task $t$ with respect to their bases $\mathbf{b}'_i$/$\mathbf{b}'_t$ and $\mathbf{b}_i$/$\mathbf{b}_t$, where the bases $\mathbf{b}'_i$ and $\mathbf{b}_i$ of the previously learned task $i$ can be retrieved from the KB of ETCL (see Figure~\ref{fig1}(a)).
\begin{equation}
\small
\label{eq-dis}
\begin{array}{l}
\begin{aligned}
&{dis}'=dis({\mathbf{b}_i}',{\mathbf{b}_t}')/\sum\nolimits_{i=1}^{t-1}dis({\mathbf{b}_i}',{\mathbf{b}_t}')\\ 
&dis\;=dis(\mathbf{b}_i,\mathbf{b}_t)/\sum\nolimits_{i=1}^{t-1}dis(\mathbf{b}_i,\mathbf{b}_t)
\end{aligned}
\end{array}
\end{equation}
where $dis'$ represents the original/true Bases Distance (BD) of the two tasks as there is no knowledge of any task in $model_{ori}$, and $dis$ denotes the BD of the two tasks based on some learned knowledge in $model_{CL}$. 

Based on the above observations and Eq.~(\ref{eq-dis}), we can infer that if $dis < dis'$, it indicates that task $i$ and task $t$  have some shared knowledge in $model_{CL}$, i.e., they have some similarities, so their BD is going to be closer than their initial BD value $dis'$, and vice versa. Thus, we propose a simple yet accurate metric of similar tasks, namely \textit{SDM} (\textit{Similarity or Dissimilarity Metric}) to measure the similarity/dissimilarity of tasks $i$ and $t$.
\begin{equation}
\label{SDM}
\small{SDM}=\left\{\begin{matrix}
  i,t \in \mathbb{T}_{sim}\; &if \;dis < {dis}',  |dis-{dis}'| \ge \delta \\
 i,t \in \mathbb{T}_{dis}\;& otherwise
\end{matrix}\right. 
\end{equation}
where $i<t, t \in [2,T]$, and $\delta$ is a distance threshold. Based on Theorems 1 and 2, $\delta$ should take an empirical value based on the training dataset of task $t$ to ensure positive KT. 

\textbf{Step 3:} Calculating the similarity between tasks $i$ (an old task) and $t$ by Eq. (\ref{SDM}). If tasks $i$ and $t$ are dissimilar, $\mathbb{T}_{dis} \leftarrow \mathbb{T}_{dis} \cup i$; otherwise, $\mathbb{T}_{sim} \leftarrow \mathbb{T}_{sim} \cup i$.
  
There is still an issue that we do not know the specific representation's bases of each task, and different tasks may have different base distributions, which makes it impossible to calculate accurately the distance of the bases $dis/{dis}'$ by the Euclidean distance or KL divergence. \cite{12.1} and \cite{12.2} pointed out that the Wasserstein distance (the schematic diagram is shown in Figure~\ref{W-D}) has some advantages over the Euclidean distance and others in this case. That is, the Wasserstein distance needs no assumptions on the distribution of the data and does not need to know the type of the distribution, and it takes into account not only the distance, but also the shape/geometry of the data, which makes it suitable for computing the distance between two distributions. Therefore, we use the Wasserstein distance in our work. 

\begin{figure}[h]
  \centering
  \includegraphics[width=0.4\textwidth]{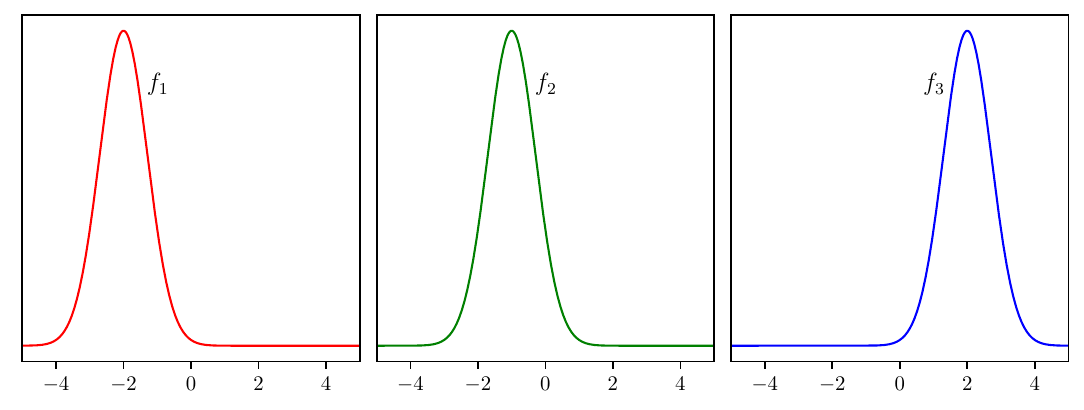}
  \caption{The schematic diagram of the difference between Euclidean distance and Wasserstein distance. The figure shows three distributions $f_1$(red), $f_2$(green) and $f_3$(blue). Each pair has the same distance in the Euclidean space. But in the Wasserstein space, $f_1$ and $f_2$ are closer as the shapes/geometries of $f_1$ and $f_2$ are more similar overall.}
  \label{W-D}
\end{figure}
\vspace{-8mm}
\subsection{Representation's Bases Calculation of Task $t$}
Through the following steps, ETCL can calculate the representation's bases $\mathbf{b}_t$ of task $t$ on dataset $\mathbb{D}'_t$.

\textbf{Step 1:} Feed $\mathbb{D}'_t$ into the corresponding model to get its  representation $\mathbf{R}_t$ of task $t$;

\textbf{Step 2:} Perform SVD (Singular Value Decomposition) \cite{13.2} on $\mathbf{R}_t$ as follows
\vspace{-2mm}
\begin{equation}
\label{svd}
\mathbf{R}_t=\mathbf{U}_t\mathbf{\Sigma}_t(\mathbf{V}_t)^T
\end{equation}
where $\mathbf{U}_t$  and $\mathbf{V}_t$ are left and right singular value matrices, respectively, which are orthogonal to each other, and $\mathbf{\Sigma}_t$ contains the singular values along the main diagonal of $\mathbf{R}_t$.

\textbf{Step 3: } With the vector approximation method based on Euclidean Distance \cite{13.2}, take the lower rank $k$-rank approximation $(\mathbf{R}_t)_k$ of $\mathbf{R}_t$ than that of $\mathbf{R}_t$, according to the following criterion for a given threshold $\epsilon_{th}$.

\begin{equation}
\label{approxi-R}
||(\mathbf{R}_t)_k||^2 \ge ||\mathbf{R}_t||^2
\end{equation}
where $||.||$ is 2-norm, and $\epsilon_{th}$ takes 0.99.

\textbf{Step 4:} Obtain $\mathbf{b}_t=\{\mathbf{u}_1,\mathbf{u}_2,...,\mathbf{u}_k \}$, where $\{\mathbf{u}_1$, $\mathbf{u}_2$, ... , $\mathbf{u}_k \}$ are the first $k$ vectors in $\mathbf{U}_t$ (see Eq.~(\ref{svd})) as \textit{the space bases of significant representation $(\mathbf{R}_t)_k$ for task $t$}.

\section{Experiments}
\subsection{Experiment Setup}

\noindent\textbf{Datasets.} In order to fully verify the ability of our ETCL and to compare with baselines in CF prevention and KT (FKT and/or BKT), a total of 11 dissimilar/similar/mixed task  datasets are used in our experiments, which are as follows:

1) \textbf{Dissimilar Task Datasets.} For this set of experiments, we use five benchmark image classification datasets: (1) PMNIST (10 tasks), (2) CIFAR 100 (10 tasks), (3) CIFAR 100 Sup (20 tasks), (4) MiniImageNet (20 tasks), and (5) 5-Datasets (5 tasks). We regard the tasks in each dataset as dissimilar as each task has different/disjoint classes. Note that two datasets CIFAR 100 Sup and 5-Datasets (consisting of 5 datasets of different tasks) are datasets with “difficult” tasks \cite{7.2.4}.

2) \textbf{Similar Task Datasets.} (1) F-EMNIST-1 (10 tasks), (2) F-EMNIST-2 (35 tasks), (3) F-CelebA-1 (10 tasks), and (4) F-CelebA-2 (20 tasks). We consider tasks in F-EMNIST and F-CelebA to be similar as each task in F-EMNIST contains one writer’s written digits/characters and each task in F-CelebA contains images of one celebrity labeled by whether he/she is smiling or not.

3) \textbf{Mixed Task Datasets.} (1) (EMNIST, F-EMNIST-1) (20 tasks) and (2)  (CIFAR 100, F-CelebA-1) (20 tasks). Each of them is a sequence of combined tasks from the similar task dataset F-EMNIST-1 (or F-CelebA-1) and the dissimilar task dataset  EMNIST (or CIFAR 100) with tasks randomly mixed. 

\vspace{+2mm}
\noindent\textbf{Baselines.} We compare ETCL with 18 SOTA baselines of 4 categories: (1) \textit{Network expansion methods.} LwF \cite{4.10}, DEN \cite{4.3}, and APD \cite{4.7}. (2) \textit{Non-network expansion methods.} (2.1) \textbf{Experience-replay/OG/Regularization-based methods}: A-GEM \cite{7.1.6}, OWM \cite{7.2.1}, OGD \cite{7.2.2}, GPM \cite{7.2.4}, EWC \cite{6.1}, UCL \cite{6.4} and CAF-MAS \cite{n99} (the best-performing combined model of CoSCL or CAF with MAS \cite{6.3}, in which the mechanisms of CAF are embedded within the representative experience-replay method MAS). 
 (2.2) \textbf{Mask-based methods}: HAT \cite{9.2}, SupSup \cite{9.12} and Piggyback \cite{n91}. (2.3) \textbf{KT-based methods.} CAT \cite{9.3}, WSN \cite{n95}, TRGP \cite{7.2.5}, CUBER \cite{n71} and ARI \cite{n93}. We use the official codes of these baselines.
 
Refer to Appendix B, C, and D for additional details about the datasets, baselines, and implementation details.

\vspace{+2mm}
\noindent\textbf{Performance Metrics.} Three metrics: 1) \textbf{Average accuracy} (ACC) of all tasks after the last task has been learned. 2) \textbf{Backward transfer} (BWT) \cite{7.1.2}: also called \textit{forgetting rate}, which indicates how much the new task affects the old tasks. A negative BWT value indicates forgetting or CF and a positive value represents positive BKT. 3) \textbf{Forward transfer} (FWT) indicates how much the old tasks affect a new task learning, which can be calculated using Eq.~(\ref{FWT-BWT}), which is also used in \cite{9.3}. A positive FWT value indicates positive FKT, otherwise negative FKT. 
\begin{equation}
\label{metrics}
\begin{array}{l}
\begin{aligned}
&ACC=\frac{1}{T}\textstyle \sum_{i=1}^{T}A_{T,i}\\
&FWT=\frac{1}{T-1}\textstyle \sum_{i,t}(A_t(g(i,t)) - A_t(g(t)))\\
&BWT=\frac{1}{T-1}\textstyle \sum_{i=1}^{T-1}(A_{T,i}-A_{i,i})
\end{aligned}
\end{array}
\end{equation}
where $i<t, t\in [2, T]$, $T$ is the total number of tasks, $A_{i,i}$ is the accuracy of task $i$ right after learning task $i$, and $A_{T,i}$ is the accuracy of the model on $i^{th}$ task after learning the last task $T$. For other notations, see Eq.~(\ref{FWT-BWT}).}

\begin{table*}[ht]
\small
\centering
\setlength{\tabcolsep}{1.6mm}{
\caption{ACC and BWT performances with standard deviations over 5 different runs of the proposed ETCL and 18 strong baselines of the 4 categories on five dissimilar benchmark datasets.}
\label{ACC+BWT}
\renewcommand\arraystretch{1.4}
\resizebox{\linewidth}{!}{
\begin{threeparttable}
\begin{tabular}{|c|c?c|c?c|c?c|c?c|c?c|c?c|c|}
\Xhline{1px}
\multicolumn{2}{|c?}{\multirow{1}{*}{\textbf{Datasets}}} & \multicolumn{2}{c?}{\textbf{PMNIST (10 Tasks)}}   & \multicolumn{2}{c?}{\textbf{CIFAR 100 (10 Tasks)}} & \multicolumn{2}{c?}{\textbf{CIFAR 100 Sup (20 Tasks)}} & \multicolumn{2}{c?}{\textbf{MiniImageNet (20 Tasks)}} & \multicolumn{2}{c|}{\textbf{5-Datasets (5 Tasks)}}& \multicolumn{2}{c|}{\textbf{Average}} \\\Xhline{1px}
\textbf{Type} & \textbf{Methods} & { ACC(\%)}   & { BWT}  & { ACC(\%)}  & { BWT}    & { ACC(\%)}      & { BWT}    & { ACC(\%)}     & { BWT}   & { ACC(\%)}     & { BWT}& { ACC(\%)}     & { BWT} \\\hline
\uppercase\expandafter{} & {\color{purple}\textbf{ONE}} & {\color{purple}\textbf{96.70}} & None        & {\color{purple}\textbf{79.58}} & None  & {\color{purple}\textbf{61.00}} & None & {\color{purple}\textbf{69.46}} & None  &{\color{purple}\textbf{93.58}}  & None &{\color{purple}\textbf{80.06}}& None\\
\Xhline{1px}

\multirow{3}{*}{\uppercase\expandafter{(1)}} 
& LwF & 85.72 {\tiny$\pm$ 0.47} & -0.11 {\tiny$\pm$ 0.01} & 67.70 {\tiny$\pm$ 0.37} & -0.08 {\tiny$\pm$ 0.01} & 51.55 {\tiny$\pm$ 0.49} & -0.03 {\tiny$\pm$ 0.01} & 60.51 {\tiny$\pm$ 0.32} &-0.03 {\tiny$\pm$ 0.01} & 89.10 {\tiny$\pm$ 0.57} & -0.02 {\tiny$\pm$ 0.01} & 70.92& -0.05\\\cline{2-14} 
 & DEN & 91.17 {\tiny$\pm$ 0.49} & -0.03 {\tiny$\pm$ 0.01} & 68.84 {\tiny$\pm$ 0.25} & -0.03 {\tiny$\pm$ 0.01} & 51.10 {\tiny$\pm$ 0.41} & -0.03 {\tiny$\pm$ 0.01} & 56.58 {\tiny$\pm$ 0.42} & -0.04 {\tiny$\pm$ 0.01} & 79.75 {\tiny$\pm$ 0.53} & -0.01 {\tiny$\pm$ 0.01}&69.49&-0.03\\\cline{2-14}
 & APD & 92.48 {\tiny$\pm$ 0.59} & -0.03 {\tiny$\pm$ 0.01} & 72.49 {\tiny$\pm$ 0.43} & -0.03 {\tiny$\pm$ 0.01} &56.81 {\tiny$\pm$ 0.44} & -0.02 {\tiny$\pm$ 0.01} & 58.73 {\tiny$\pm$ 0.51} & -0.03 {\tiny$\pm$ 0.01} & 83.72 {\tiny$\pm$ 0.54} & -0.07 {\tiny$\pm$0.01}&72.86&-0.04\\\Xhline{1px}

\multirow{8}{*}{\uppercase\expandafter{(2.1)}}
& A-GEM  & 83.56 {\tiny$\pm$ 0.16} & -0.13 {\tiny$\pm$ 0.01} & 63.98 {\tiny$\pm$ 1.22} & -0.15 {\tiny$\pm$ 0.02} & 42.78 {\tiny$\pm$ 0.89} & -0.13 {\tiny$\pm$ 0.05} & 57.24 {\tiny$\pm$ 0.72}  & -0.12 {\tiny$\pm$ 0.01} & 84.04 {\tiny$\pm$ 0.33} & -0.12 {\tiny$\pm$ 0.01} &66.33&-0.13\\\cline{2-14}
& OWM  & 90.71 {\tiny$\pm$ 0.11} & -0.02 {\tiny$\pm$ 0.01} & 50.94 {\tiny$\pm$ 0.60} & -0.03 {\tiny$\pm$ 0.01} & -- & --  & -- & -- & -- & -- &70.83&-0.03\\\cline{2-14}
& OGD  & 82.50 {\tiny$\pm$ 0.13} & -0.14 {\tiny$\pm$ 0.01} & 47.12 {\tiny$\pm$ 0.87} & -0.04 {\tiny$\pm$ 0.01} & 36.92 {\tiny$\pm$ 0.57} & -0.03 {\tiny$\pm$ 0.04} & 44.89 {\tiny$\pm$ 0.49} & -0.04 {\tiny$\pm$ 0.02} & 57.12 {\tiny$\pm$ 0.41} & -0.04 {\tiny$\pm$ 0.01}&53.71&-0.06\\\cline{2-14}
& GPM  & 93.91 {\tiny$\pm$ 0.16} & -0.03 {\tiny$\pm$ 0.01} & 72.48 {\tiny$\pm$ 0.40} & -0.03 {\tiny$\pm$ 0.01} & 57.10 {\tiny$\pm$ 0.38} & -0.03 {\tiny$\pm$ 0.01} & 60.41 {\tiny$\pm$ 0.01} & -0.03 {\tiny$\pm$ 0.04} & 91.22 {\tiny$\pm$ 0.22} & -0.01 {\tiny$\pm$ 0.00}&75.02&-0.03\\\cline{2-14}
& EWC  & 89.97 {\tiny$\pm$ 0.57} & -0.04 {\tiny$\pm$ 0.01}   & 68.80 {\tiny$\pm$ 0.88} & -0.02 {\tiny$\pm$  0.01} & 41.49 {\tiny$\pm$ 0.79} & -0.03 {\tiny$\pm$ 0.02} & 52.01 {\tiny$\pm$ 2.53} & -0.12 {\tiny$\pm$ 0.03} & 86.61 {\tiny$\pm$ 0.20} & -0.05 {\tiny$\pm$ 0.01} &64.18&-0.05\\\cline{2-14}
& UCL  & 89.53 {\tiny$\pm$ 0.22} & -0.05 {\tiny$\pm$ 0.01} & 64.08 {\tiny$\pm$ 0.46} & -0.06 {\tiny$\pm$ 0.02} & 47.22 {\tiny$\pm$ 0.53} & -0.09 {\tiny$\pm$ 0.02} & 45.85 {\tiny$\pm$  0.41} & -0.10 {\tiny$\pm$ 0.04} & 88.54 {\tiny$\pm$ 0.38} & -0.05 {\tiny$\pm$ 0.02} &67.04&-0.07\\\cline{2-14}
& CAF-MAS  & 92.85 {\tiny$\pm$ 0.17} & -0.03 {\tiny$\pm$ 0.01} & 69.22 {\tiny$\pm$ 0.41} & -0.01 {\tiny$\pm$ 0.02} & 59.71 {\tiny$\pm$ 0.46} & -0.01 {\tiny$\pm$ 0.02} & 70.81 {\tiny$\pm$  0.39} & -0.02 {\tiny$\pm$ 0.04} & 89.54 {\tiny$\pm$ 0.35} & -0.05 {\tiny$\pm$ 0.02} &76.43&-0.03\\\hline
 
\multirow{3}{*}{\uppercase\expandafter{(2.2)}} & HAT  & 90.35 {\tiny$\pm$ 0.32} & \color{blue}\textbf{0.00} {\tiny$\pm$ 0.00} & 72.06 {\tiny$\pm$ 0.30} & 0.00 {\tiny$\pm$ 0.00} & 55.85 {\tiny$\pm$ 0.37} & \color{blue}\textbf{0.00} {\tiny$\pm$ 0.00} & 59.78  {\tiny$\pm$ 0.47} &-0.03 {\tiny$\pm$ 0.01} & 91.32  {\tiny$\pm$ 0.18} & -0.01 {\tiny$\pm$ 0.00} &73.87&-0.01\\\cline{2-14}
 & SupSup  & 96.03 {\tiny$\pm$ 0.12} & \color{blue}\textbf{0.00} {\tiny$\pm$ 0.00} & 74.63 {\tiny$\pm$ 0.36} & 0.00 {\tiny$\pm$ 0.00} & 61.53 {\tiny$\pm$ 0.23} & \color{blue}\textbf{0.00} {\tiny$\pm$ 0.00} & 70.55 {\tiny$\pm$ 0.20} & 0.00  {\tiny$\pm$ 0.00} & 92.30 {\tiny$\pm$ 0.19} & \color{blue}\textbf{0.00} {\tiny$\pm$ 0.00}&79.08&0.00\\\cline{2-14}
 & Piggyback  & 95.73 {\tiny$\pm$ 0.17} & \color{blue}\textbf{0.00} {\tiny$\pm$ 0.00} & 69.82 {\tiny$\pm$ 0.26} & 0.00 {\tiny$\pm$ 0.00} & 48.45 {\tiny$\pm$ 0.53} & \color{blue}\textbf{0.00} {\tiny$\pm$ 0.00} & \color{blue}\textbf{73.58} {\tiny$\pm$ 0.27} & 0.00{\tiny$\pm$ 0.00} & 93.26 {\tiny$\pm$ 0.59} & \color{blue}\textbf{0.00} {\tiny$\pm$ 0.00}&76.17&0.00\\\hline
 
\multirow{4}{*}{\uppercase\expandafter{(2.3)}} & CAT  &93.87 {\tiny$\pm$ 0.51} & -0.03 {\tiny$\pm$ 0.01} & 59.06 {\tiny$\pm$ 0.49} & -0.08 {\tiny$\pm$ 0.01} & 50.23 {\tiny$\pm$ 0.32} & -0.02 {\tiny$\pm$ 0.01} & 59.55 {\tiny$\pm$ 0.61} & -0.03 {\tiny$\pm$ 0.01} & 86.05 {\tiny$\pm$ 0.74} & -0.04 {\tiny$\pm$ 0.03}&69.75&-0.04\\\cline{2-14}
& WSN & 96.41 {\tiny$\pm$ 0.17} & \color{blue}\textbf{0.00} {\tiny$\pm$ 0.00} & \color{blue}\textbf{75.59} {\tiny$\pm$ 0.27} & 0.00 {\tiny$\pm$ 0.00} & \color{blue}\textbf{61.74} {\tiny$\pm$ 0.23} & \color{blue}\textbf{0.00} {\tiny$\pm$ 0.00} & 71.96 {\tiny$\pm$ 0.41} & 0.00 {\tiny$\pm$ 0.00} & \color{blue}\textbf{93.38} {\tiny$\pm$ 0.12} & \color{blue}\textbf{0.00} {\tiny$\pm$ 0.00}&\color{blue}\textbf{79.82}&0.00\\\cline{2-14}
 & TRGP  & 96.34 {\tiny$\pm$ 0.11} & -0.08 {\tiny$\pm$ 0.01} & 73.95 {\tiny$\pm$ 0.32} & -0.02 {\tiny$\pm$ 0.01} & 58.48 {\tiny$\pm$ 0.01} & -0.01 {\tiny$\pm$  0.00} & 60.73 {\tiny$\pm$ 0.60} & -0.02 {\tiny$\pm$ 0.06} & 92.82 {\tiny$\pm$ 0.10} & -0.04 {\tiny$\pm$ 0.01}&76.47&-0.03\\\cline{2-14}
 & CUBER & \color{blue}\textbf{97.04} {\tiny$\pm$ 0.11} & -0.02 {\tiny$\pm$ 0.01} & 74.67 {\tiny$\pm$ 0.22} & \color{blue}\textbf{0.01} {\tiny$\pm$ 0.01} & 58.51 {\tiny$\pm$ 0.01} & -0.01 {\tiny$\pm$  0.00} & 66.92 {\tiny$\pm$ 0.35} & \color{blue}\textbf{0.07} {\tiny$\pm$ 0.04} & 91.36 {\tiny$\pm$ 0.30} & -0.01 {\tiny$\pm$ 0.00}&77.70&\color{blue}\textbf{0.01}\\\cline{2-14}
  & ARI & 84.20 {\tiny$\pm$ 0.13} & \color{blue}\textbf{0.00} {\tiny$\pm$ 0.01} & 48.90 {\tiny$\pm$ 0.28} & -0.02 {\tiny$\pm$ 0.01} & - & - & - & - & - & -&66.55&-0.01\\\cline{2-12}
 \Xhline{1px}
\multicolumn{2}{|c?}{ETCL(\textbf{\textbf{Ours}})}   & \textbf{97.11} {\tiny$\pm$ 0.03} & \textbf{0.00 {\tiny$\pm$ 0.00}} & \textbf{77.41} {\tiny$\pm$ 0.11} & 0.00 {\tiny$\pm$ 0.00} & \textbf{62.28} {\tiny$\pm$ 0.01} &\textbf{0.00 {\tiny$\pm$ 0.00}} & \textbf{74.21} {\tiny$\pm$ 0.11} & 0.00{\tiny$\pm$ 0.00} & \textbf{93.46}  {\tiny$\pm$ 0.06} &\textbf{0.00} {\tiny$\pm$ 0.00}&\textbf{80.89}&0.00\\ \Xhline{1px}
\end{tabular}
\begin{tablenotes}
\item \textbf{ONE} -- building a model for each task independently using a separate neural network, which has no knowledge transfer and no forgetting involved (denoted as \textbf{None}). As CAT is bound to its specific network structure, its experimental results are run according to its network structure and source code. Other methods use the same backbone network on each dataset shown in Appendix D. ``--" indicates that the source codes are not provided by the baselines leading to no experimental results. The {\color{purple}red results} indicate the ONE’s results, and the {\color{blue}blue results} mean the best prior results.
\end{tablenotes}
\end{threeparttable}}}
\end{table*}

\begin{table*}[ht]
\centering
\setlength{\tabcolsep}{1.8mm} 
\caption{FWT and BWT performances with standard deviations of the proposed ETCL and 7 strong baselines with/without the KT capacity over 5 different runs on four similar task datasets.}
\label{KT1}
\renewcommand\arraystretch{1.4}
\resizebox{\linewidth}{!}{
\begin{threeparttable}
\begin{tabular}{|c|c|c|c|c|c|c|c|c|c|c|c|c|c|c|c|}
\hline
\multicolumn{1}{|c?}{\multirow{1}{*}{{\small\textbf{Datasets}}}}  & \multicolumn{3}{c?}{\small\textbf{F-EMNIST-1 (10 Tasks)} }   & \multicolumn{3}{c?}{\small\textbf{F-EMNIST-2 (35 Tasks)} } & \multicolumn{3}{c?}{\small\textbf{F-CelebA-1 (10 Tasks)}} & \multicolumn{3}{c?}{\small\textbf{F-CelebA-2 (20 Tasks)}} & \multicolumn{3}{c|}{\small\textbf{Average}} \\\hline
\textbf{ Methods}  & {ACC (\%)}   & {FWT}  & {BWT}   & {ACC (\%)}   & {FWT}  & {BWT}   & {ACC (\%)}   & {FWT}  & {BWT} & {ACC (\%)}   & {FWT}  & {BWT} & {ACC (\%)}   & {FWT}  & {BWT}  \\\hline
{\color{purple}\textbf{ONE}}& {\color{purple}\textbf{69.85}}& None & None  & { \color{purple}\textbf{71.55}}& None& None&{\color{purple}\textbf{75.55}}& None& None& {\color{purple}\textbf{76.09}}& None& None & {\color{purple}\textbf{73.26}} &None&None\\\hline
CAF-MAS& {62.87} {\tiny$\pm$  0.12}& {-0.0217}& {-0.0528}& { 77.52}  {\tiny$\pm$  0.26}& 0.0724& { -0.0127} &{72.10} {\tiny$\pm$  0.41}& { {-0.0349}}& {{0.0004}} & { {71.21}} {\tiny$\pm$  0.17}& { {-0.0549}}& \color{blue} \textbf{0.0061} & { 70.93 }& {{-0.0098}} & {-0.0148}\\\hline
SupSup& {66.92} {\tiny$\pm$  0.26}& {-0.0293}& {0.0000}& { 72.15}  {\tiny$\pm$  0.21}& 0.0060& { 0.000} &{70.46} {\tiny$\pm$  0.37}& { {-0.0509}}& {{0.0000}} & { {69.32}} {\tiny$\pm$  0.29}& { {-0.0677}}& {0.0000} & { 69.71 }& {{-0.0354}} & {0.0000}\\\hline
GPM& {75.18} {\tiny$\pm$  0.06}& {0.0372}& {0.0218}& { 79.20}  {\tiny$\pm$  0.40}& { \color{blue}\textbf{0.0782}}& { -0.0007} &{ \color{blue}\textbf{84.00}} {\tiny$\pm$  0.36}& { \color{blue}\textbf{0.0741}}& {\color{blue}\textbf{0.0104}} & { \color{blue}\textbf{77.39}} {\tiny$\pm$  0.30}& { \color{blue}\textbf{0.0176}}& {-0.0046} & { \color{blue}\textbf{78.94}}& {\color{blue}\textbf{0.0518}} & {0.0067}\\\hline
CAT& 61.90{\tiny$\pm$  0.21}& -0.1041& 0.0259& 63.00{\tiny$\pm$  0.25}&  -0.0964&  \color{blue}\textbf{0.0164}&  73.42{\tiny$\pm$  0.21}&  -0.0113&  -0.0100& 68.21{\tiny$\pm$  0.12}&  -0.0788& 0.0000 &  66.63 & -0.0727 & 0.0081\\\hline
WSN&  78.10{\tiny$\pm$  0.17}&  \color{blue}\textbf{0.0825}&  0.0000&  76.34{\tiny$\pm$  0.25}&  0.0479&  0.0000& 75.55{\tiny$\pm$  0.21}&  0.0000& 0.0000& 74.30{\tiny$\pm$  0.12}& -0.0179&  0.0000 & 76.07 & 0.0282 & 0.0000\\\hline
TRGP&  76.66{\tiny$\pm$  0.46}&  0.0469&  \textbf{0.0301} &  \color{blue}\textbf{79.54}{\tiny$\pm$  0.42}& 0.0715&  0.0100&  76.30{\tiny$\pm$  0.49}&  0.0075&  0.0000& 72.58{\tiny$\pm$  0.35}&  -0.0351&  0.0000&  76.27 & 0.0227&  \color{blue}\textbf{0.0100}\\\hline
CUBER&  \color{blue}\textbf{78.48}{\tiny$\pm$  0.47}& 0.0703&  0.0215&  76.80{\tiny$\pm$  0.53}& 0.0578&  -0.0126& 76.36{\tiny$\pm$  0.55}&  0.0076& 0.0005& 72.59{\tiny$\pm$  0.33}& -0.0350&  0.0000& 76.05 &  0.0252 & 0.0022\\\hline
ETCL(\textbf{Ours})& \textbf{ 80.32}{\tiny$\pm$  0.23}& \textbf{0.0948}& 0.0141& \textbf{82.57}{\tiny$\pm$  0.17}&  \textbf{0.1055}& 0.0080&  \textbf{87.27}{\tiny$\pm$  0.11}& \textbf{0.1202}& \textbf{0.0107}& \textbf{86.82}{\tiny$\pm$  0.12}&  \textbf{0.1033}& \textbf{0.0096} &  \textbf{84.25} &  \textbf{0.1060}&  \textbf{0.0106}\\\hline
\end{tabular}
\begin{tablenotes} 
\item The ResNet-18 backbone is used for the four similar task datasets F-EMNIST-1, F-EMNIST-2, F-CelebA-1 and F-CelebA-2 as most baselines and our ETCL except CAF-MAS and CAT, where CAF-MAS uses AlexNet  backbone while CAT uses 3-Layer FCN. As CAT is bound to their specific network structure 3-Layer FCN, its experimental results were run according to its network architecture and source code. The {\color{purple}red results} indicate the ONE's results, and the {\color{blue}blue results} mean the best prior results.
 \end{tablenotes}
\end{threeparttable}}
\end{table*}

\begin{table*}[ht]
\centering
\caption{KT performances of ETCL and 8  strong baselines with/without KT mechanisms over 5 different runs on two mixed task  datasets.}\label{KT2}
\renewcommand\arraystretch{0.96}
\resizebox{\linewidth}{!}{
\begin{threeparttable}
\begin{tabular}{|p{2cm}<{\centering}?p{1.5cm}<{\centering}|p{1.5cm}<{\centering}|p{1.5cm}<{\centering}?p{1.5cm}<{\centering}|p{1.5cm}<{\centering}|p{1.5cm}<{\centering}?p{1.5cm}<{\centering}|p{1.5cm}<{\centering}|p{1.5cm}<{\centering}|}
\hline
\multicolumn{1}{|c?}{\textbf{Datasets}} & \multicolumn{3}{c?}{\textbf{(EMNIST, F-EMNIST-1) (20 Tasks)}}  & \multicolumn{3}{c?}{\textbf{(CIFAR 100, F-CelebA-1) (20 Tasks)}} & \multicolumn{3}{c|}{\textbf{Average}}\\\hline
\footnotesize\textbf{Methods}  & \footnotesize ACC (\%)   & \footnotesize FWT  & \footnotesize BWT  &\footnotesize ACC (\%)   &\footnotesize FWT   &\footnotesize BWT   &\footnotesize ACC (\%) &\footnotesize FWT  &\footnotesize BWT  \\\hline
\footnotesize \textcolor{purple}{\textbf{ONE 1}}& \footnotesize\textcolor{purple}{\textbf{77.44}}& \footnotesize None&\footnotesize None&\footnotesize  \textcolor{purple}{\textbf{64.50}}&\footnotesize None &\footnotesize None &\footnotesize  \textcolor{purple}{\textbf{70.97}}&\footnotesize None&\footnotesize None\\\hline
\footnotesize SupSup&\footnotesize 69.48 {\tiny$\pm$  0.26}&\footnotesize -0.0796&\footnotesize {0.0000} &\footnotesize  \textcolor{blue}{\textbf{65.34} {\tiny$\pm$  0.14}}&\footnotesize 0.0084 &\footnotesize \textcolor{blue}{\textbf{0.0000}} &\footnotesize  67.41 {\tiny$\pm$  0.21}&\footnotesize -0.0356 &\footnotesize 0.0000\\\hline
\footnotesize GPM&\footnotesize 73.69 {\tiny$\pm$  0.32}&\footnotesize -0.0365&\footnotesize \textcolor{blue}{\textbf{0.0038}} &\footnotesize  64.28 {\tiny$\pm$  0.28}&\footnotesize 0.0013 &\footnotesize -0.0037 &\footnotesize  68.99 {\tiny$\pm$  0.28}&\footnotesize -0.0189 &\footnotesize \textcolor{blue}{\textbf{0.0001}}\\\hline
 \footnotesize HAT& \footnotesize70.70{\tiny$\pm$  0.18}& \footnotesize -0.0626& \footnotesize 0.0000 & \footnotesize 56.82{\tiny$\pm$  0.13}& \footnotesize -0.0768& \footnotesize \textcolor{blue}{\textbf{0.0000}} & \footnotesize 63.76{\tiny$\pm$  0.13}& \footnotesize -0.0677& \footnotesize 0.0000\\\hline
\footnotesize CAT&\footnotesize 74.61{\tiny$\pm$  0.19}&\footnotesize -0.0045&\footnotesize -0.0219&\footnotesize 61.94{\tiny$\pm$  0.16}&\footnotesize -0.0256&\footnotesize \textcolor{blue}{\textbf{0.0000}}&\footnotesize 68.28{\tiny$\pm$  0.16}&\footnotesize -0.0151&\footnotesize -0.0110\\\hline
\footnotesize WSN&\footnotesize 74.23{\tiny$\pm$  0.17}&\footnotesize -0.0230&\footnotesize 0.0000&\footnotesize 61.05{\tiny$\pm$  0.16}&\footnotesize -0.0345&\footnotesize \textcolor{blue}{\textbf{0.0000}} &\footnotesize 67.64{\tiny$\pm$  0.13}&\footnotesize -0.0288&\footnotesize 0.0000\\\hline
\footnotesize TRGP&\footnotesize 75.53{\tiny$\pm$  0.28}&\footnotesize 0.0012&\footnotesize -0.0136&\footnotesize 61.92{\tiny$\pm$  0.21}&\footnotesize -0.0117&\footnotesize -0.0155&\footnotesize 68.73{\tiny$\pm$  0.21}&\footnotesize -0.0048&\footnotesize -0.0146\\\hline
\footnotesize CUBER&\footnotesize \textcolor{blue}{\textbf{77.23}{\tiny$\pm$  0.28}}&\footnotesize \textcolor{blue}{\textbf{0.0053}}&\footnotesize -0.0074&\footnotesize 64.85{\tiny$\pm$  0.31}&\footnotesize \textcolor{blue}{\textbf{0.0109}} &\footnotesize -0.0073&\footnotesize \textcolor{blue}{\textbf{71.04}{\tiny$\pm$  0.31}}&\footnotesize \textcolor{blue}{\textbf{0.0081}}&\footnotesize -0.0073\\\hline
\footnotesize ETCL 1 (\textbf{Ours})&\footnotesize \textbf{78.81}{\tiny$\pm$  0.19}&\footnotesize \textbf{0.0126}&\footnotesize 0.0006&\footnotesize \textbf{68.31}{\tiny$\pm$  0.11}&\footnotesize \textbf{0.0154}&\footnotesize \textbf{0.0246}&\footnotesize \textbf{73.56}{\tiny$\pm$  0.12}&\footnotesize \textbf{0.0140}&\footnotesize \textbf{0.0126}\\\midrule \midrule
\footnotesize \textcolor{purple}{\textbf{ONE 2}}& \footnotesize\textcolor{purple}{\textbf{87.36}}& \footnotesize None&\footnotesize None&\footnotesize  \textcolor{purple}{\textbf{72.31}}&\footnotesize None &\footnotesize None &\footnotesize  \textcolor{purple}{\textbf{79.84}}&\footnotesize None&\footnotesize None\\\hline
\footnotesize CAF-MAS&\footnotesize 86.40 {\tiny$\pm$  0.27}&\footnotesize -0.0019&\footnotesize -0.0083 &\footnotesize 71.10 {\tiny$\pm$  0.16}&\footnotesize 0.0113 &\footnotesize -0.0364 &\footnotesize  78.75 {\tiny$\pm$  0.28}&\footnotesize 0.0047 &\footnotesize -0.0224\\\hline
\footnotesize ETCL 2 (\textbf{Ours})&\footnotesize \textbf{88.45}{\tiny$\pm$  0.23}&\footnotesize \textbf{0.0071}&\footnotesize \textbf{0.0043}&\footnotesize \textbf{74.59}{\tiny$\pm$  0.14}&\footnotesize \textbf{0.0145}&\footnotesize \textbf{0.0096}&\footnotesize \textbf{81.52}{\tiny$\pm$  0.13}&\footnotesize \textbf{0.0108}&\footnotesize \textbf{0.0070}\\\hline
\end{tabular}
  
     \begin{tablenotes}
     \footnotesize
        \item[1] AlexNet  is used for ONE 2, CAF-MAS and our ETCL 2, while 3-Layer FCN is used for ONE 1, our ETCL 1 and all other seven baselines (i.e., SupSup, GPM, HAT, CAT, WSN, TRGP and CUBER) except CAF-MAS on the two mixed task datasets as the six baselines perform poorly on AlexNet. The {\color{purple}red results} indicate the ONE's results, and the {\color{blue} blue results} are the best prior results.
        
     \end{tablenotes}
    \end{threeparttable}}
\end{table*}

\begin{figure*}[ht]
	\centering 
    \captionsetup[subfloat]{skip=2pt, position=bottom}
        \captionsetup[subfloat]{aboveskip=0pt, belowskip=-5pt}
	\subfloat[MiniImageNet (20 Tasks)]{
		\includegraphics[width=0.4\linewidth]{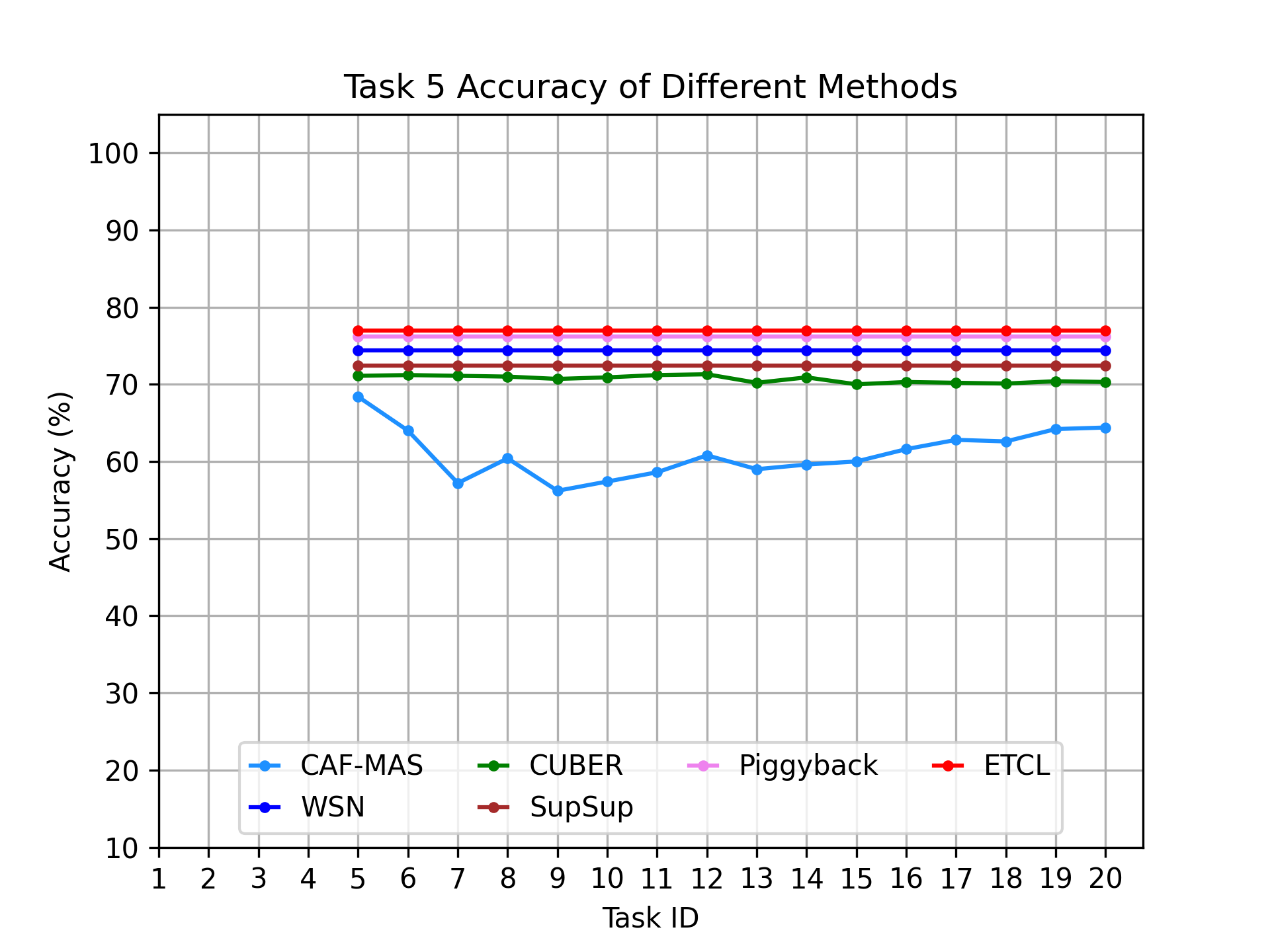}}
	\subfloat[F-EMNIST-1 (10 Tasks)]{
		\includegraphics[width=0.4\linewidth]{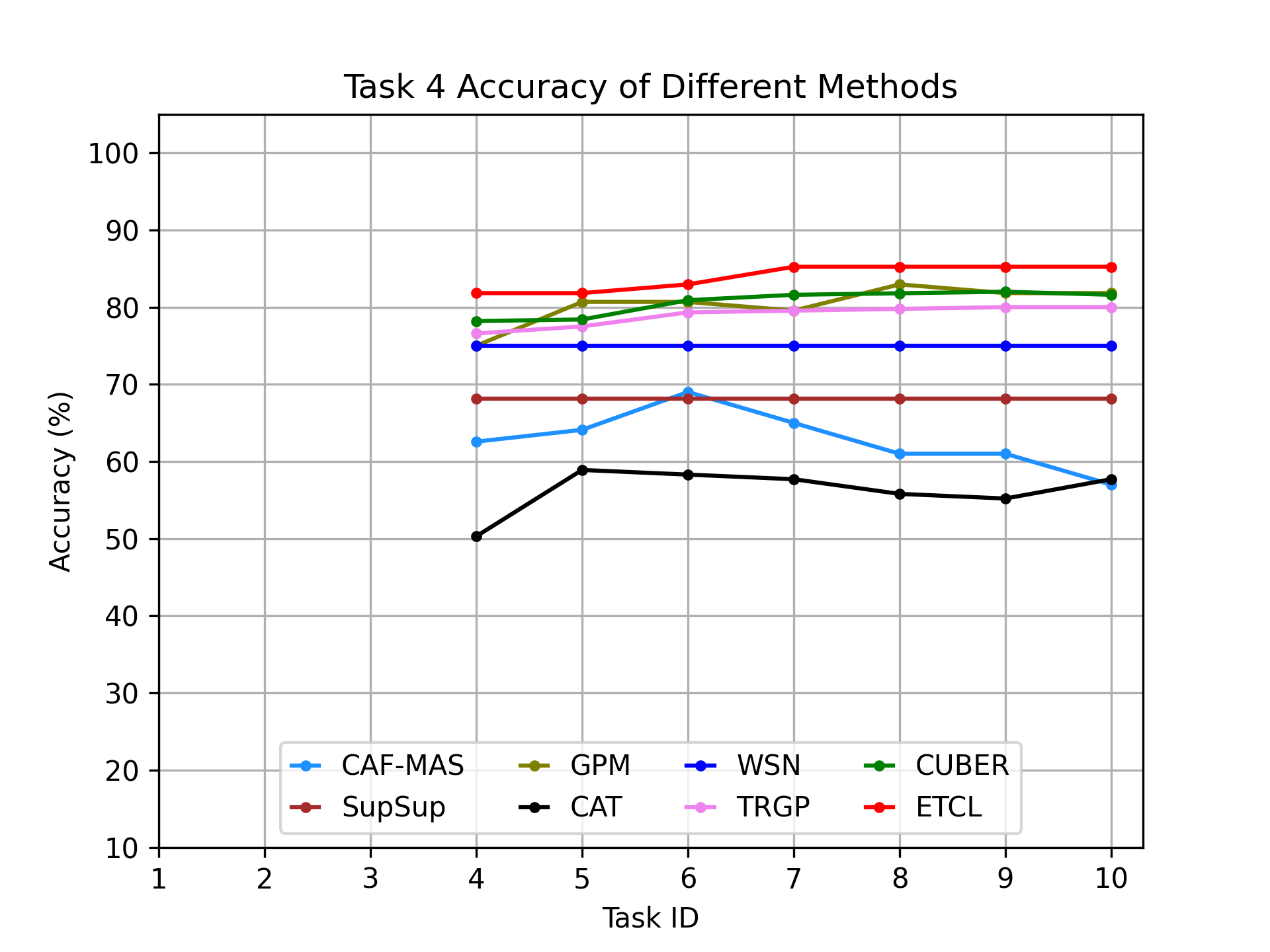}}
	  \\
	\subfloat[F-CelebA-2 (20 Tasks)]{
		\includegraphics[width=0.4\linewidth]{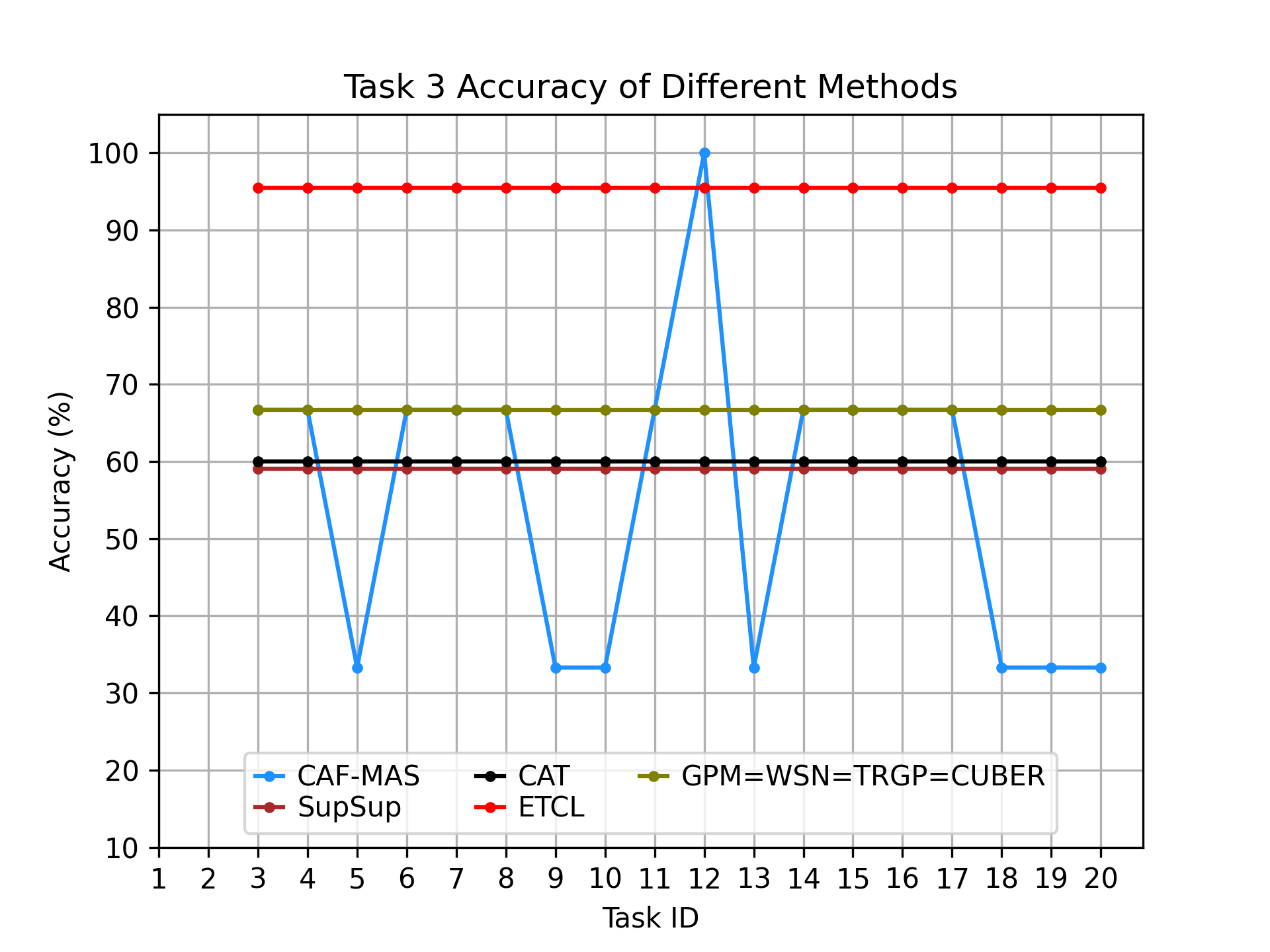}}
	\subfloat[(CIFAR 100, F-CelebA-1) (20 Tasks)]{
		\includegraphics[width=0.4\linewidth]{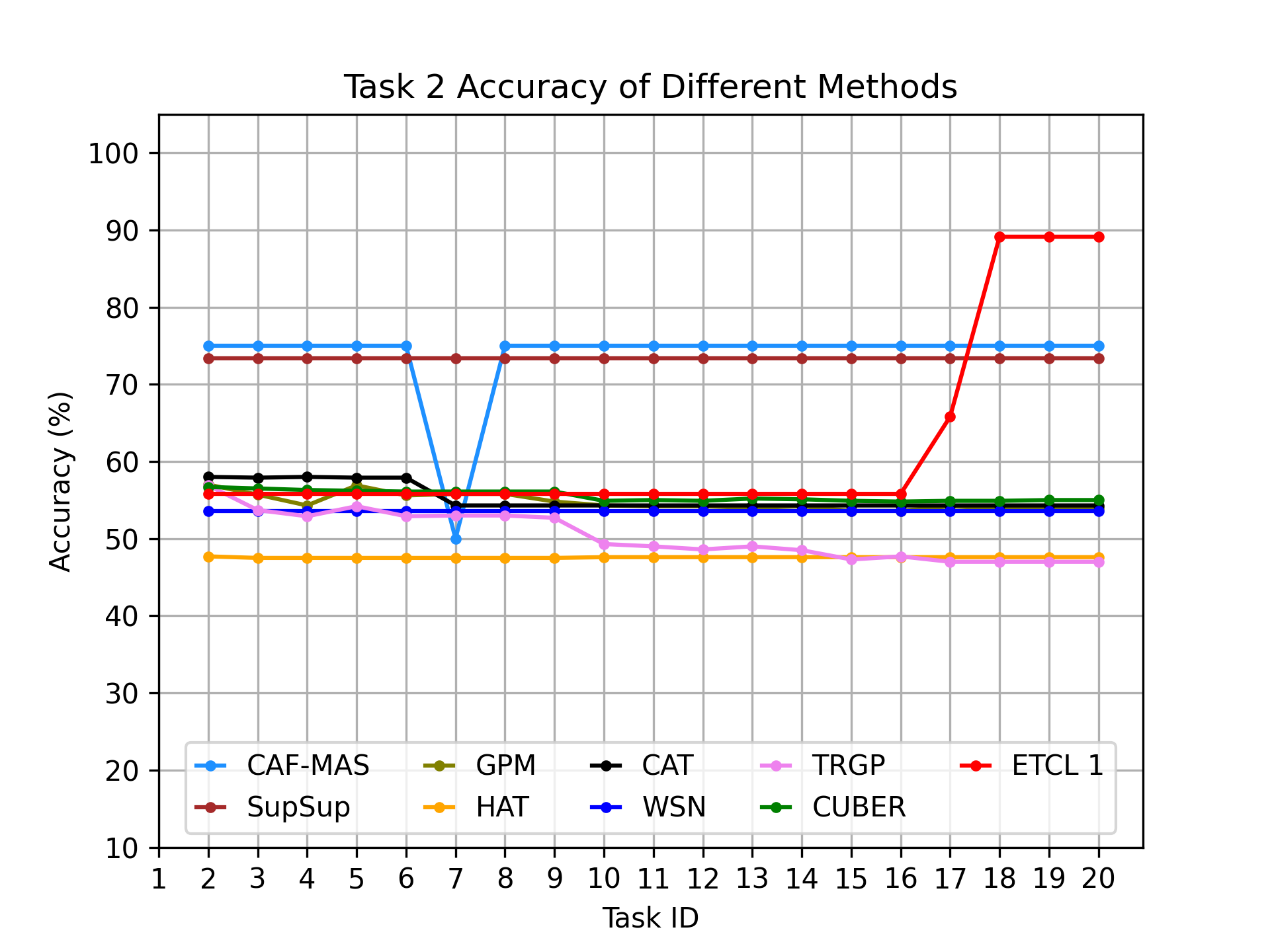}}
	\caption{The performances of  $A_{i,i}$ and  $A_{t,i}$, where (a) $t \in [6,20]$ and $i=5$ on the dissimilar task dataset MiniImageNet (20 tasks), (b) $t \in [5,10]$ and  $i=4$ on the similar task dataset F-EMNIST-1 (10 tasks), (c) $t \in [4,20]$ and $i=3$ on the similar task dataset F-CelebA-2 (20 tasks) and (d) $t \in [3,20]$ and $i=2$ on the mixed task dataset (CIFAR 100, F-CelebA-1) (20 tasks).}
\label{Per-Ati}
\end{figure*}

\subsection{Main Experimental Results and Analysis}
\noindent\textbf{Results of Dissimilar Tasks - Overcoming CF.} The task sequences here consist only of dissimilar tasks, which have little shared knowledge to transfer. We use ACC and BWT (forgetting rate) as the metrics to evaluate their average accuracy and CF prevention. Table \ref{ACC+BWT} reports the results, which shows that ETCL outperforms all 18 strong baselines in ACC and exceeds the average ACC (71.46\%) of all baselines by up to 9.43\%. We notice that WSN is only slightly weaker than our ETCL as it also has no forgetting. This is not surprising as the tasks are dissimilar and as long as there is no forgetting, the performance cannot be improved much. When similar tasks are used, WSN is much weaker than our ETCL (see Tables \ref{KT1} and \ref{KT2} below). Importantly, ETCL not only achieved zero forgetting (BWT=0.0) on all 5 datasets but also improved the average accuracy of each task by 0.41\%, 1.28\% and 4.75\% as compared with ONE respectively on 3 datasets PMNIST, CIFAR 100 Sup and MiniImageNet, which show some positive FKT, while the average BWT of all baselines is negative, -0.03 on average. And we notice that although the average BWT ($=0.01$)  of CUBER is positive, its BWT on five datasets has positive and negative oscillations. In addition, its average ACC on the five datasets is weaker than that of our ETCL with the average ACC margin of 3.19\% due to its limited FKT.

\noindent\textbf{Results of Similar Tasks - Knowledge Transfer (KT).} Similar task sequences contain more shared knowledge to transfer. Table~\ref{KT1} reports the FWT and BWT performances of the proposed ETCL and 7  strong baselines that were designed with/without the explicit KT capacity. CAF-MAS is the best-performing combined model of CoSCL or CAF with MAS, in which the mechanisms of CAF are embedded within the representative experience-replay method MAS. SupSup, CAT and WSN are mask-based methods, while TRGP and CUBER are built up on the OG-based GPM method. Table~\ref{KT1} shows that ETCL achieves all positive FWT and BWT in four similar tasks datasets, resulting in ACC gains of 10.47\%, 11.02\%, 11.72\% and 10.73\% respectively compared to ONE. Although GPM was not designed for KT, it actually performs very well, especially in its forward transfer capability. CAT is weak as it works only with 3-Layer FCN. Our ETCL is strong in both forward and backward transfer. The average results in the rightmost column show that ETCL is significantly better than the baselines. It is worth noting that with multiple continual learners and a fixed parameter budget, although CoSCL and CAF can improve a variety of representative continual learning methods' performances on ACC, FWT, and BWT, e.g., CAF-MAS, by a large margin, their FWT and BWT are both negative, resulting in their weak ACC performances.

\vspace{+2mm}
\noindent\textbf{Results for Mixed Tasks - CF prevention and KT}: At this point, because similar and dissimilar tasks appear randomly in the mixed task sequences, it becomes more challenging to achieve CF prevention and positive KT. However, unlike all baselines, Table \ref{KT2} clearly shows that our ETCL achieves both positive FKT and BKT. With backbone 3-Layer FCN, compared with the average ACC results (73.64\% and 62.31\%) of the seven baselines (i.e.,  SupSup, GPM, HAT, CAT, WSN, TRGP and CUBER) on the two mixed task datasets, our ETCL respectively obtains the gains of 5.17\% and 6.0\%, and achieves the improved ACC of 1.37\% and 3.81\% as compared with the corresponding ONE 1. And with backbone AlexNet, compared with the ACC results of the recent SOTA method CAF-MAS, our ETCL respectively obtains the gains of 2.05\% and 3.49\%, and achieves the improved ACC of 1.09\% and 2.28\% as compared with the corresponding ONE 2.
 
Moreover, CAF-MAS is bound to backbone AlexNet, while the other seven baselines perform poorly on AlexNet on the two mixed task datasets. However, our ETCL works well on AlexNet or 3-Layer FCN, which shows that our ETCL has a better model generalization than the baselines. It is worth noting that the ACC performance of CAF-MAS outperforms the other baselines on the two mixed datasets, which contributes to its well-balanced mechanism: balancing the flexibility of learning new tasks and the memory stability of old tasks, and collaboration with multiple continuous learners. The results of CAF-MAS suggest that the well-balanced mechanism and ensemble of multiple continual learners on mixed task sequences are promising means to improve the generalization and performance of a CL model.

\vspace{-3mm}
\subsection{Ablation Experiments}

The proposed ETCL achieves its positive FKT and BKT by relying on the proposed three new techniques: a new task similarity metric SDM (Eq.~(\ref{SDM}) based on Wasserstein distance,  Aligning Initial Gradients (AIG, Str-2 in Sec.~IV) to further guide and enhance FKT, and Bi-objective Optimisation (BIO) to achieve positive BKT with maximal transfer and minimal interference (Str-3 in Sec.~IV). The ablation experimental results are given in Table \ref{ETCL-AblT}. ``ETCL(-SDM)" denotes without using SDM task similarity metric but using Euclidean distance, ``ETCL(-AIG)" means without deploying the AIG strategy in ETCL, and ``ETCL(-BIO)" means removing the BIO in ETCL. 

\begin{table}[H]
\caption{{\color{black}Ablation experiments of the proposed ETCL. }}\label{ETCL-AblT}
\centering
\resizebox{\linewidth}{!}{
\begin{tabular}{|l?c?c?c?c|}
\hline
\multicolumn{1}{|c?}{\multirow{2}{*}{ \small\textbf{Datasets}}} & \multicolumn{1}{c?}{\small\textbf{ETCL(-SDM)}} & \multicolumn{1}{c?}{\small\textbf{ETCL(-AIG)}} & \multicolumn{1}{c?}{\small\textbf{ETCL(-BIO)}}  & \multicolumn{1}{c|}{\small\textbf{ETCL}} \\\cline{2-5}
\multicolumn{1}{|c?}{} & \multicolumn{1}{c?}{\small{ ACC(\%)}} & \multicolumn{1}{c?}{\small{ ACC(\%)}}  & \multicolumn{1}{c?}{\small{ ACC(\%)}}  & \multicolumn{1}{c|}{\small{ ACC(\%)}}\\\hline
\small F-EMNIST-1 & \small 74.22& \small 79.83 & \small 69.37  & \small \small \textbf{80.32}\\\hline
\small F-EMNIST-2& \small 72.71& \small 82.21  & \small 72.20 & \small \textbf{82.57} \\\hline
\small F-CelebA-1& \small 83.10& \small 84.77  & \small 74.77 & \small \textbf{87.27}\\\hline
\small F-CelebA-2& \small 78.89& \small 84.32 & \small 77.66 & \small \textbf{86.82} \\\hline
\small (EMNIST, F-EMNIST-1)& \small 77.07& \small 77.36  & \small 77.35  & \small \textbf{78.81} \\\hline
\small (CIFAR 100, CelebA-1) & \small 66.10& \small 65.30 & \small 63.15 & \small \textbf{68.31} \\\hline
\end{tabular}}
\vspace{-2mm}
\end{table}

The ablation results show that the full ETCL always gives the best ACC and every component, i.e., SDM, AIG or BIO, contributes to the model's performance. Particularly, on the more similar tasks datasets, i.e., EMNIST-2 (35 tasks) and F-CelebA-2 (20 tasks), if the SDM or BIO mechanism is removed from ETCL, the accuracy of ETCL will drop sharply, which shows the effectiveness of the proposed SDM and BIO. In addition, the results in the table also show that the ACC performance of ETCL with the AIG mechanism is improved by an average of 2.55\% on similar or mixed task sequences, which fully demonstrates the necessity and correctness of ETCL's AIG mechanism.

\vspace{-3mm}
\subsection{Additional  Performance Experimental Results}

\begin{figure*}[ht]  
    \centering  
    \captionsetup[subfigure]{skip=2pt}  
    \captionsetup[subfloat]{aboveskip=0pt, belowskip=-5pt}
    \subfloat[MiniImageNet (20 Tasks)]{ 
        \includegraphics[width=0.2\linewidth]{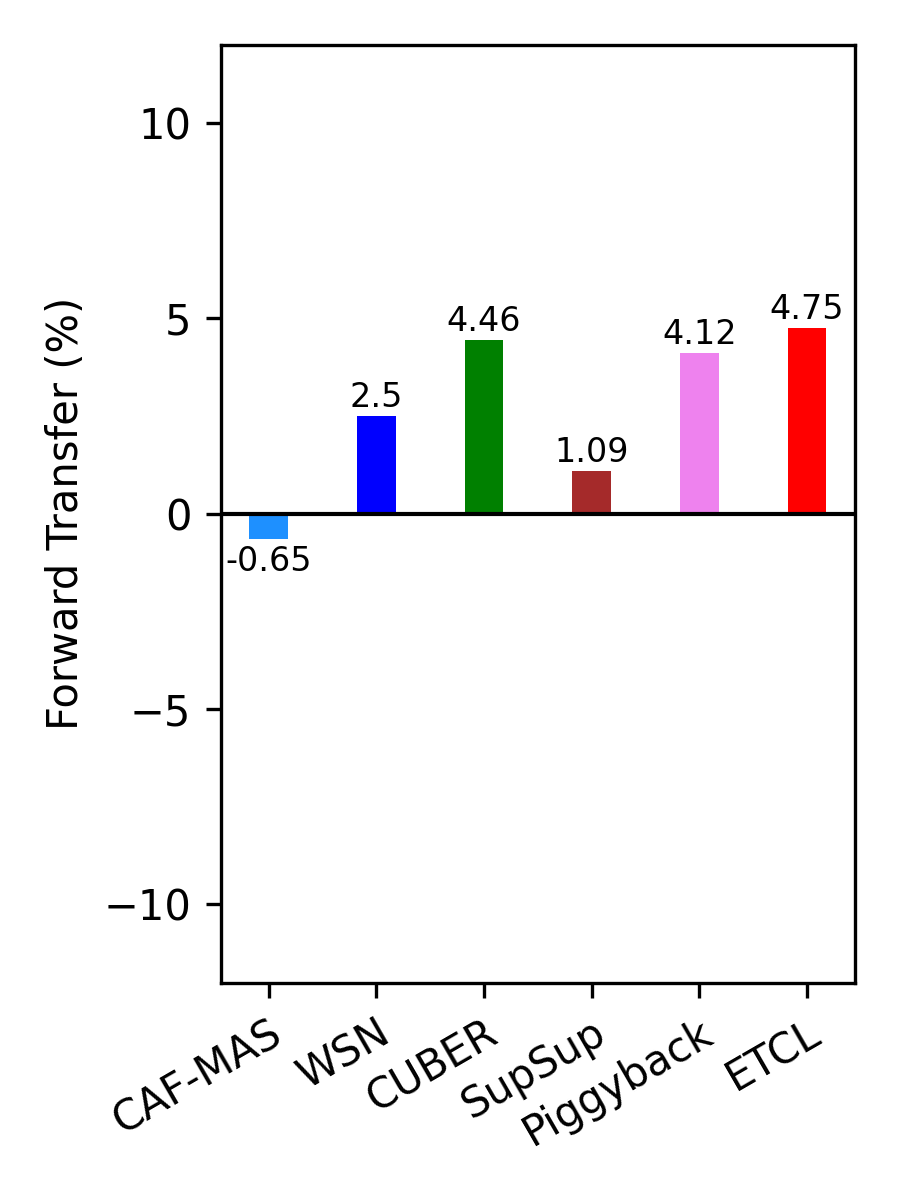}   
        \hfill
        \includegraphics[width=0.2\linewidth]{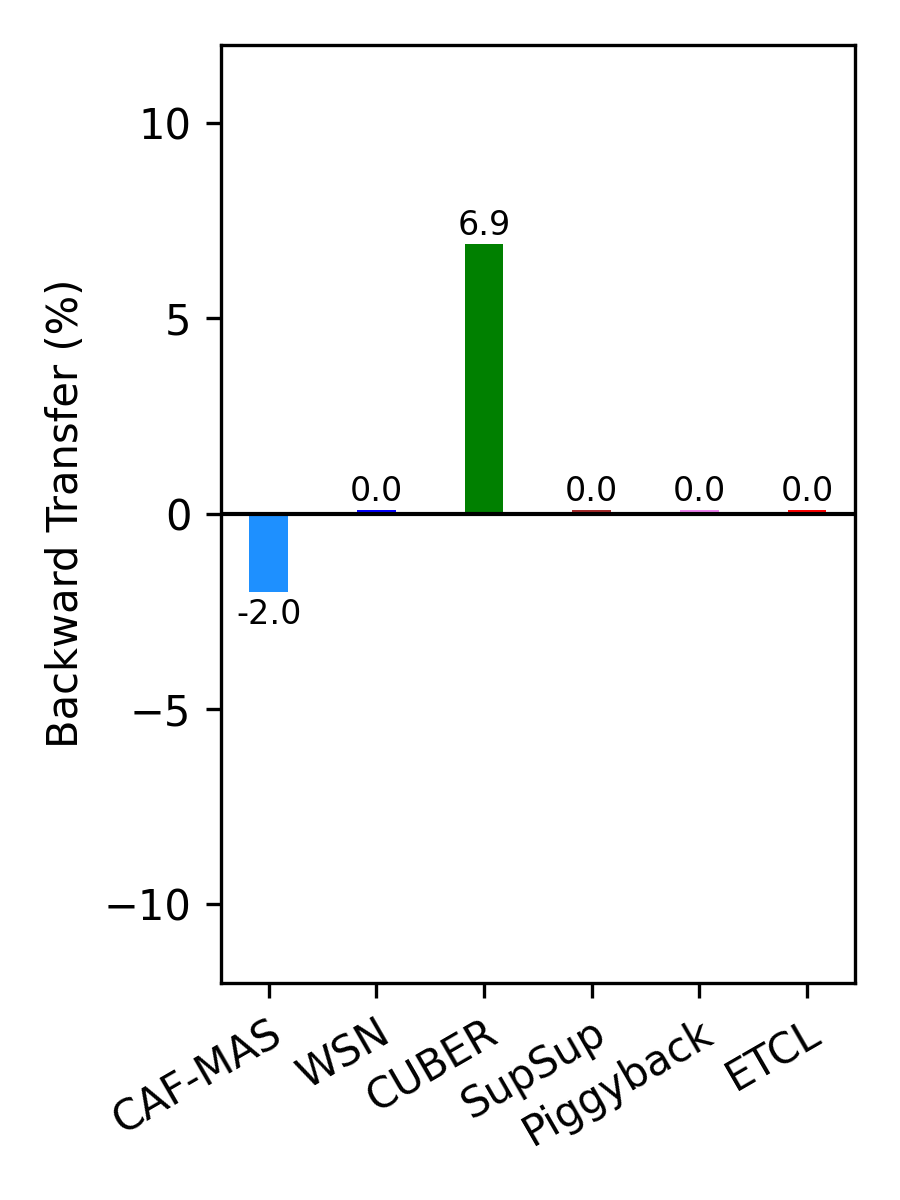}     
    }     
    \subfloat[F-EMNIST-1 (10 Tasks)]{ 
        \includegraphics[width=0.2\linewidth]{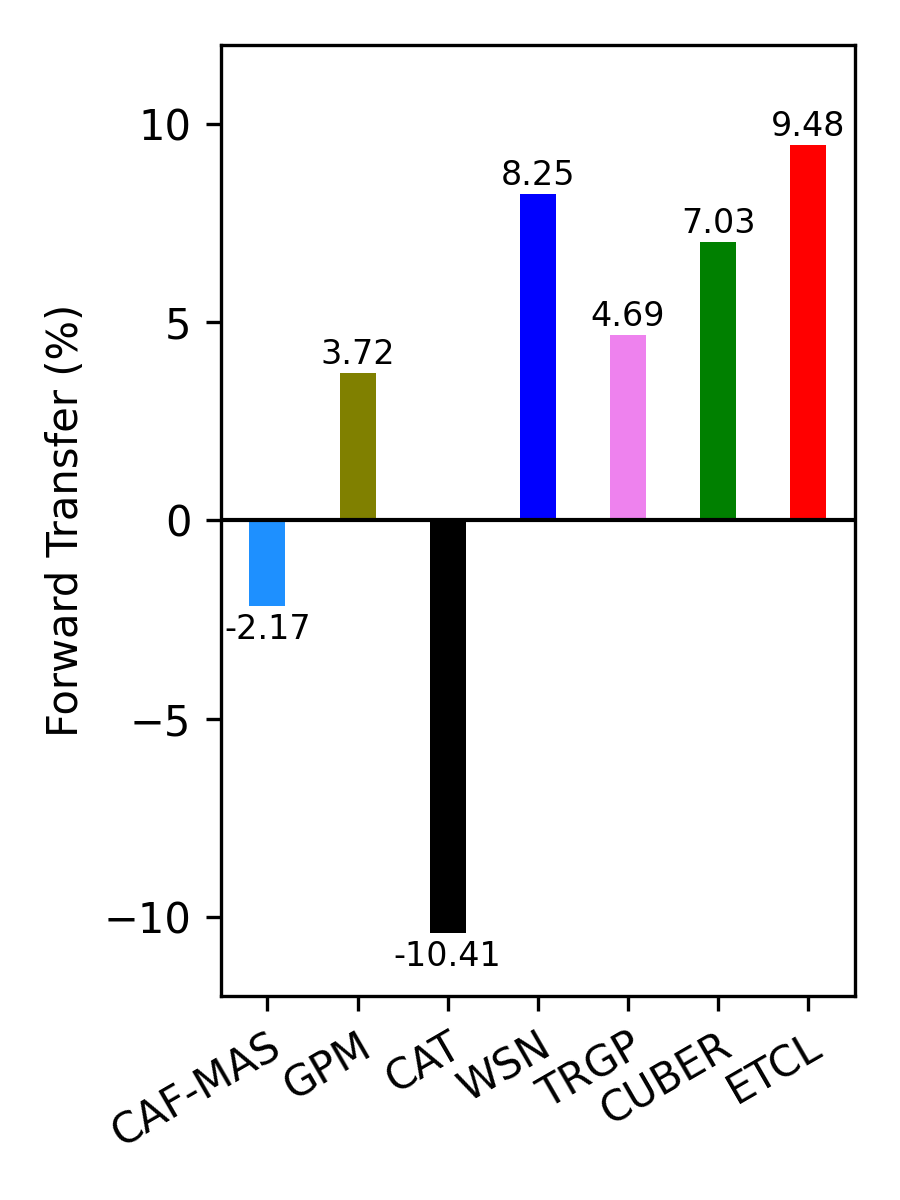}  
        \hfill 
        \includegraphics[width=0.2\linewidth]{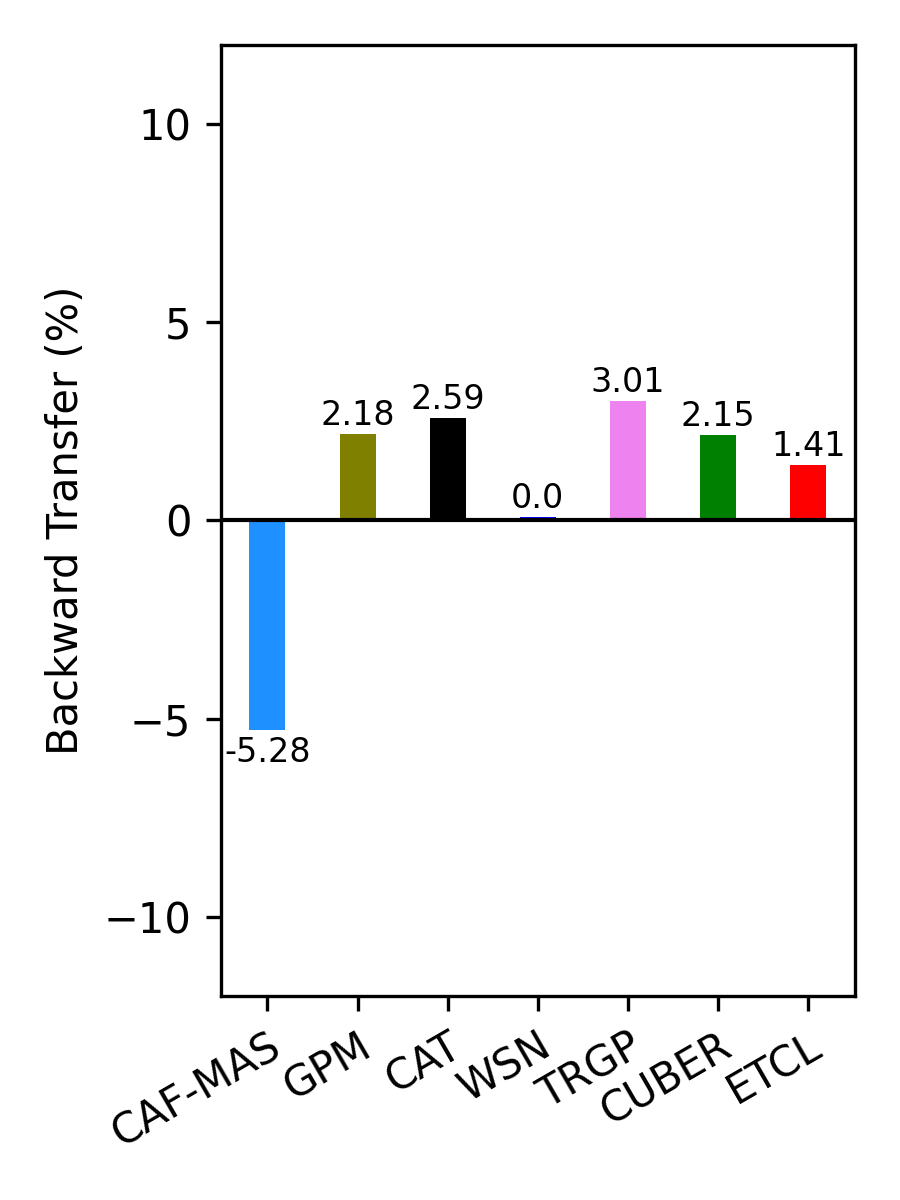}  
    }        
    \\ 
    \subfloat[F-CelebA-2 (20 Tasks)]{  
        \includegraphics[width=0.2\linewidth]{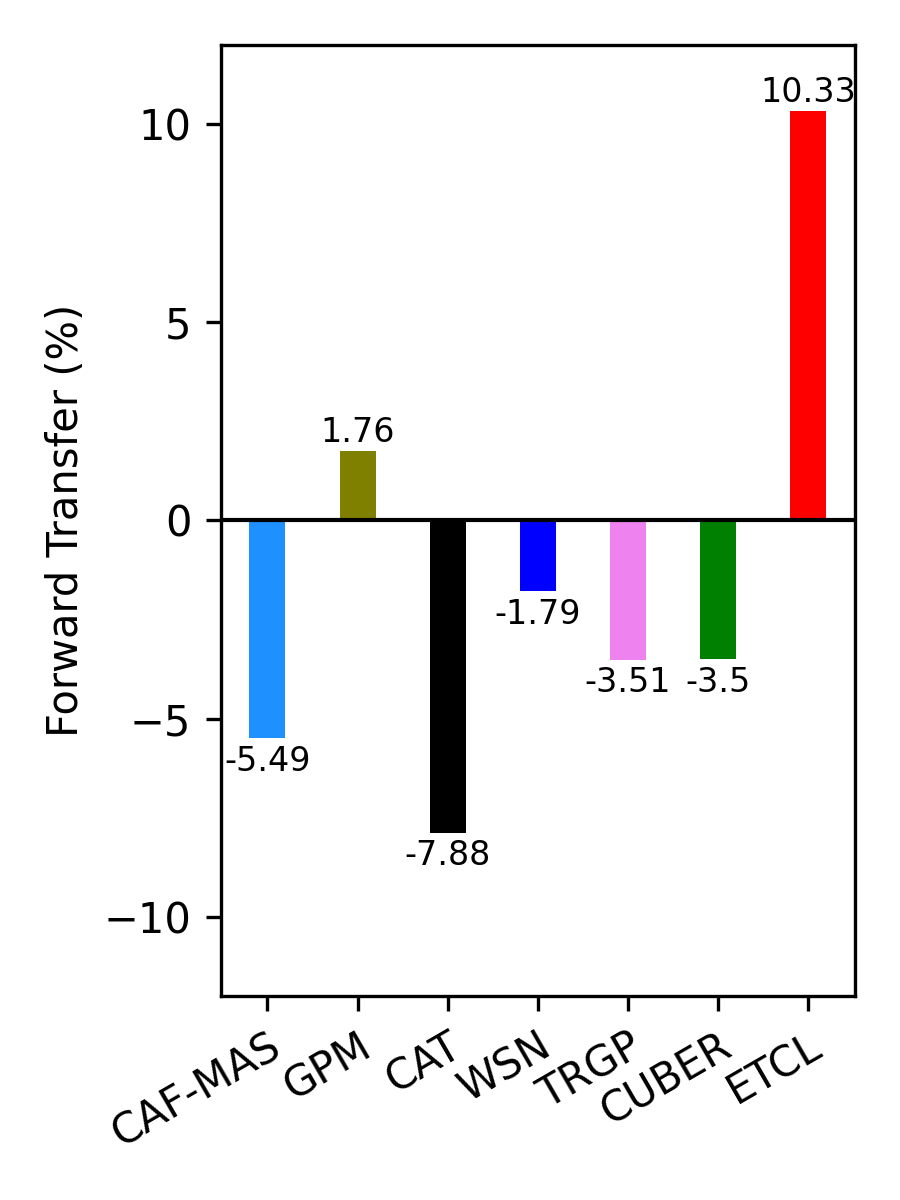}  
        \hfill 
        \includegraphics[width=0.2\linewidth]{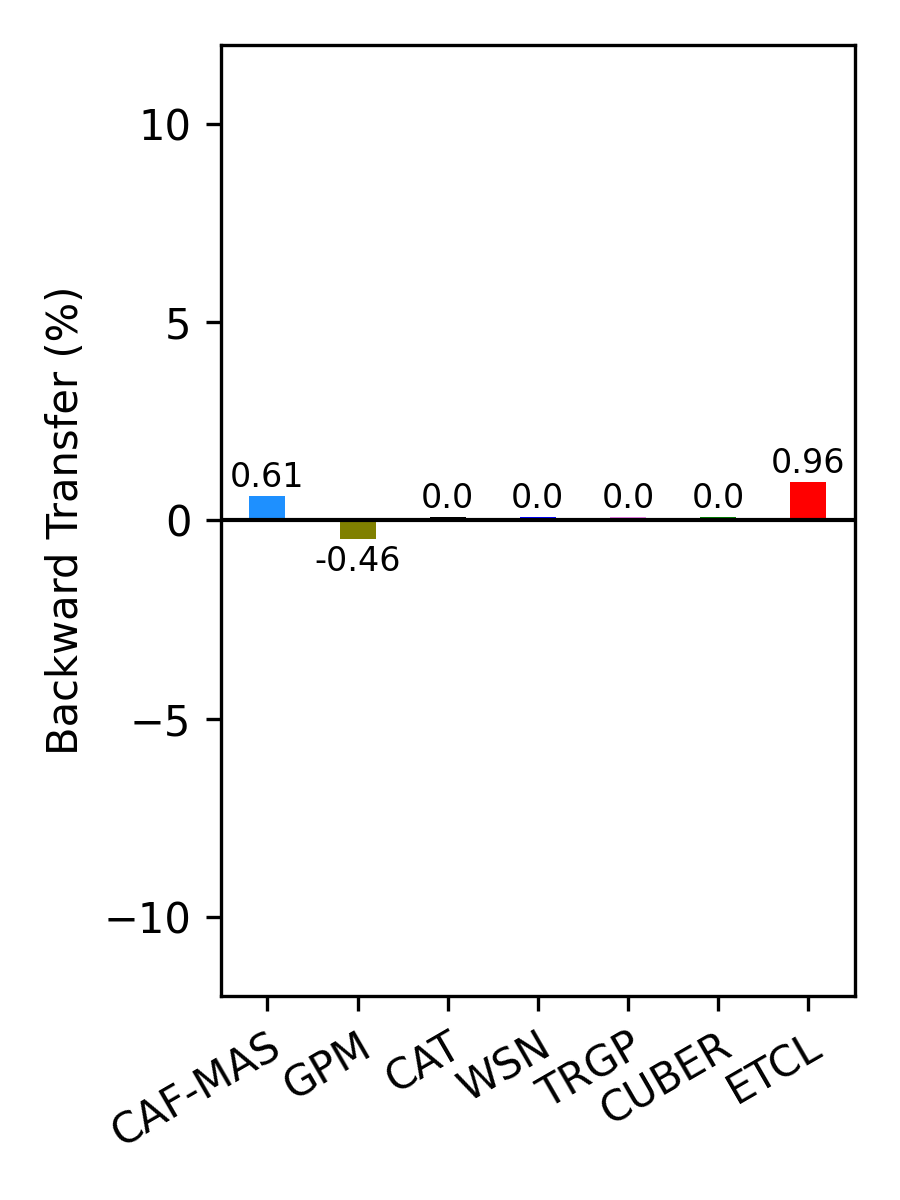}  
    }  
    \subfloat[(CIFAR 100, F-CelebA-1) (20 Tasks)]{  
        \includegraphics[width=0.2\linewidth]{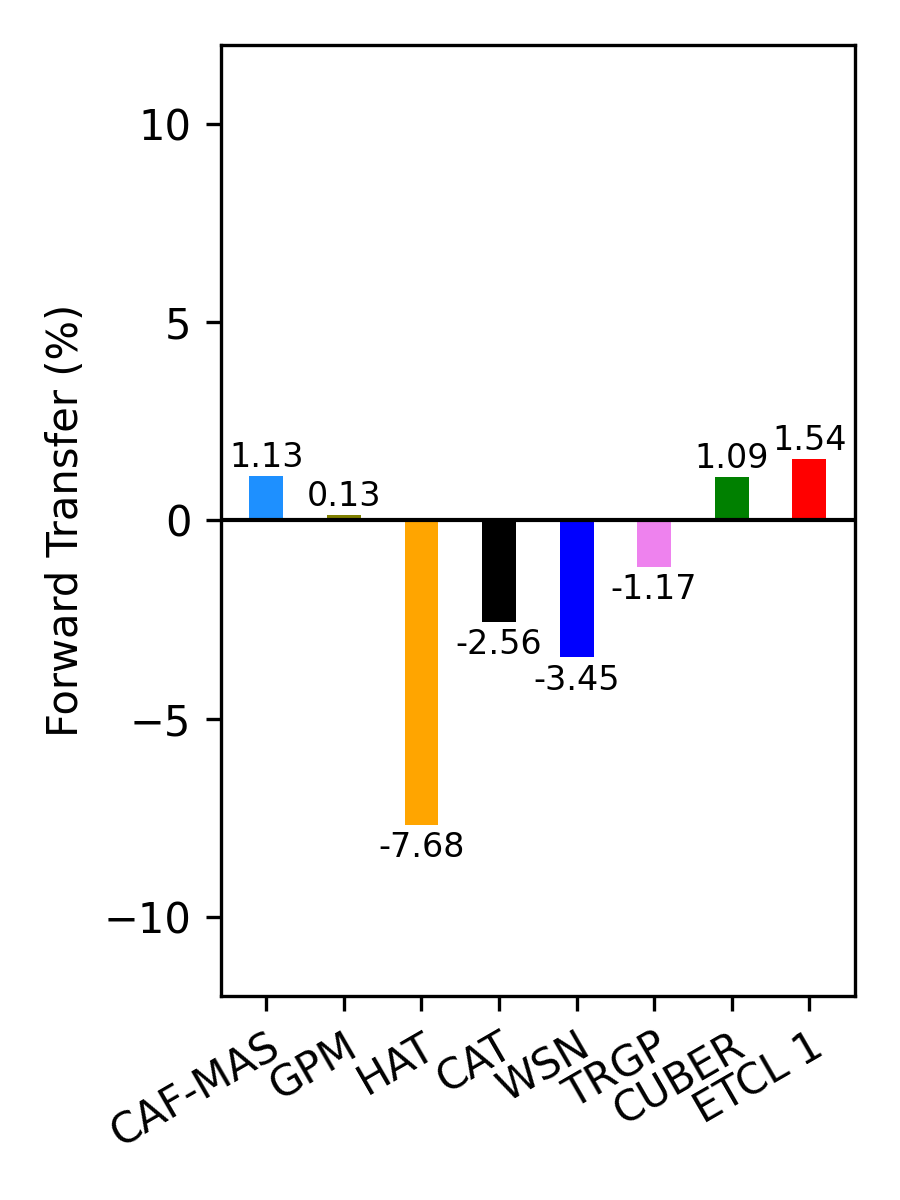}  
        \hfill 
        \includegraphics[width=0.2\linewidth]{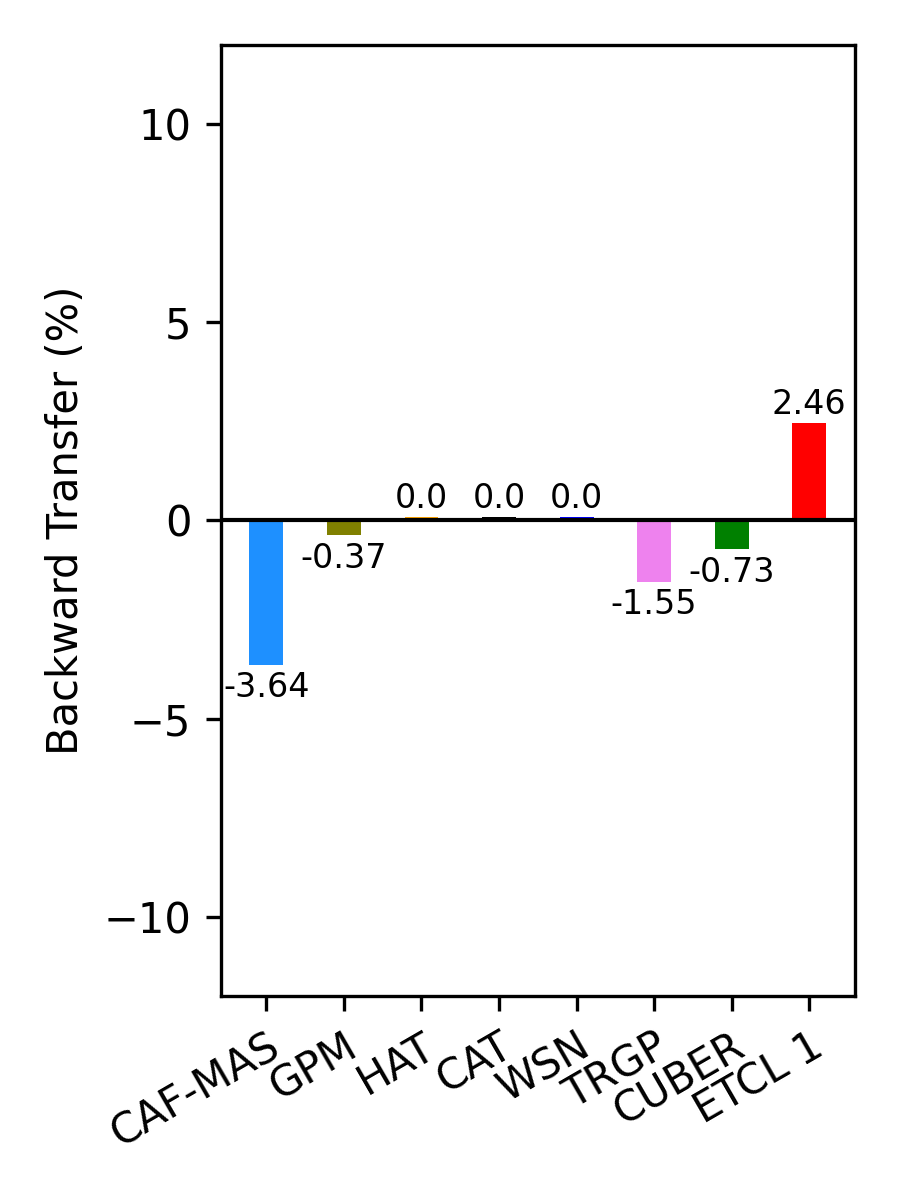}    
    }  
      
    \caption{The FWT and BWT performances of SOTA TIL methods on various dissimilar/similar/mixed task datasets.} 
\label{FWT-BWT bar} 
\end{figure*}

According to the performance metrics shown in Eq.~(\ref{metrics}), we have known that if a TIL method has a positive/negative high $A_{i,i}$  value, it is shown that the method has a strong positive/negative FWT; if the $A_{t,i}$ curve of a TIL method has a stable and upward or downward trend, it is shown that the method must have a strong positive or negative BWT respectively, where $A_{t,i}$ is the accuracy of task $i$ after learning a  new task $t$ ($i<t, t \in [i+1, T]$). Figure \ref{Per-Ati} shows  $A_{i,i}$ and $A_{t,i}$ experimental results of some SOTA TIL methods on various datasets, while Figure \ref{FWT-BWT bar} gives their corresponding BWT and FWT on the datasets, where the task $i$ is randomly selected in each dataset to make the case more convincingly. 

From  Figures \ref{Per-Ati} and \ref{FWT-BWT bar}, we can get the following observations: (1) The parameter isolation-based methods and KT-based methods generally have higher ACC, FWT, and BWT performances, e.g., Piggyback, WSN, and SupSup; (2) In the TIL, just achieving the goal: to balance the learning plasticity of new tasks and
the memory stability of old tasks is not enough, e.g., CAF-MAS, which inevitably causes instability and/or degradation of the performance ACC and negative KT; (3) On mixed task datasets, it is more challenging to achieve both forgetting-free and positive KT. If the KT mechanism is not well designed, the model performance will still deteriorate (see the performance of TRGP (with FKT mechanism) shown in Figure \ref{Per-Ati}~(d)); (4) When learning new tasks, the knowledge of similar old tasks existing in the network can be reused only if the conditions of both Theorem 1 and Theorem 2 are satisfied; otherwise the negative FKT will result. See the FWT performance of WAN (with automatic forward knowledge transfer without task similarity judgment) shown in Figure \ref{FWT-BWT bar}~(c)-(d); (5) The proposed ETCL markedly outperforms all strong baselines on dissimilar/similar/mixed task datasets, which validates the ideal optimization objective of TIL (see Eq.~(\ref{iTIL})) and the strategies of the CF prevention and positive KT proposed in this paper. 

Since the time and space consumption and scalability of a model are also important indicators of the quality of a model, we conducted comparative experiments on the two above performances of ETCL and some SOTA baselines. The experimental results show that ETCL has better time-space complexity and the best model scalability. The detailed comparative experimental results of the time-space complexity and model scalability are given in Appendix E.

\section{Conclusion}
To overcome the weakness of the existing CL methods in terms of KT and to achieve the ideal goal of CL, in this research, we theoretically study the KT problem and give the bounds that can lead to negative forward and backward KT. Equipped with the proposed new task similarity metric, and a new type of the mask which can overcome CF and perform positive KT simultaneously, we propose a novel TIL method ETCL. Extensive experimental results have shown that the proposed ETCL not only can achieve forgetting-free but also can perform significantly better positive FKT and BKT than various strong baselines on similar, dissimilar or mixed task sequences. Further theoretical research on KT and improving ETCL's accuracy are our future research directions.

\section{Acknowledgments}
This work of Z. Wang, Z. Wu, Y. Li, Y. Wang was supported by the National Natural Science Foundation of China (No.~62176202 and No.~62272367) and Joint Funds of the National Natural Science Foundation of China (U22A2098).


\bibliographystyle{IEEEtran}
\bibliography{main.bib}

@inproceedings{7.2.5,
  title={TRGP: Trust Region Gradient Projection for Continual Learning},
  author={Lin, Sen and Yang, Li and Fan, Deliang and Zhang, Junshan},
  booktitle={Proceedings of the 9th International Conference on Learning Representations},
  year={2021}
}

@inproceedings{konishi2023parameter,
  title={Parameter-Level Soft-Masking for Continual Learning},
  author={Konishi, Tatsuya and Kurokawa, Mori and Ono, Chihiro and Ke, Zixuan and Kim, Gyuhak and Liu, Bing},
  booktitle={Proceedings of the 40th International Conference on Machine Learning},
  pages={17492--17505},
  year={2023}
}

@inproceedings{7.2.7,
  author    = {Hao Liu and
               Huaping Liu},
  title     = {Continual Learning with Recursive Gradient Optimization},
  booktitle = {Proceedings of the 10th International Conference on Learning Representations},
  year      = {2022},
}

@inproceedings{9.3,
  author    = {Zixuan Ke and
               Bing Liu and
               Xingchang Huang},
  title     = {Continual Learning of a Mixed Sequence of Similar and Dissimilar Tasks},
  booktitle = {Advances in Neural Information Processing Systems},
  year      = {2020},
}

@inproceedings{4.8,
  author    = {Qi Qin and
               Wenpeng Hu and
               Han Peng and
               Dongyan Zhao and
               Bing Liu},
  title     = {{BNS:} Building Network Structures Dynamically for Continual Learning},
  booktitle = {Advances in Neural Information Processing Systems},
  pages     = {20608--20620},
  year      = {2021},
}

@inproceedings{
6.1,
author = {James Kirkpatrick  and Razvan Pascanu  and Neil Rabinowitz  and Joel Veness  and Guillaume Desjardins  and Andrei A. Rusu  and Kieran Milan  and John Quan  and Tiago Ramalho  and Agnieszka Grabska-Barwinska  and Demis Hassabis  and Claudia Clopath  and Dharshan Kumaran  and Raia Hadsell },
title = {Overcoming catastrophic forgetting in neural networks},
booktitle = {Proceedings of the National Academy of Sciences},
volume = {114},
number = {13},
pages = {3521-3526},
year = {2017},
}

@inproceedings{7.2.2,
  author    = {Mehrdad Farajtabar and
               Navid Azizan and
               Alex Mott and
               Ang Li},
  title     = {Orthogonal Gradient Descent for Continual Learning},
  booktitle = {Proceedings of the 23rd International Conference on Artificial Intelligence and Statistics},
  volume    = {108},
  pages     = {3762--3773},
  publisher = {{PMLR}},
  year      = {2020},
}

@inproceedings{6.3,
  author    = {Rahaf Aljundi and
               Francesca Babiloni and
               Mohamed Elhoseiny and
               Marcus Rohrbach and
               Tinne Tuytelaars},
  title     = {Memory Aware Synapses: Learning What (not) to Forget},
  booktitle = {Proceedings of the 15th European Conference on Computer Vision},
  volume    = {11207},
  pages     = {144--161},
  year      = {2018},
}

@inproceedings{7.2.4,
  author    = {Gobinda Saha and
               Isha Garg and
               Kaushik Roy},
  title     = {Gradient Projection Memory for Continual Learning},
  booktitle = {Proceedings of the 9th International Conference on Learning Representations},
  year      = {2021},
}

@article{7.2.1,
  author    = {Guanxiong Zeng and
               Yang Chen and
               Bo Cui and
               Shan Yu},
  title     = {Continual learning of context-dependent processing in neural networks},
  journal   = {Nature Machine Intelligence},
  volume    = {1},
  number    = {8},
  pages     = {364--372},
  year      = {2019},
}

@Inbook{8.1.1,
author="Thrun, Sebastian",
title="Lifelong Learning Algorithms",
bookTitle="Learning to Learn",
year="1998",
publisher="Springer US",
address="Boston, MA",
pages="181--209",
}

@article{4.10,
  author    = {Zhizhong Li and
               Derek Hoiem},
  title     = {Learning without Forgetting},
  journal   = {{IEEE} Transactions on Pattern Analysis
and Machine Intelligence},
  volume    = {40},
  number    = {12},
  pages     = {2935--2947},
  year      = {2018},
}

@inproceedings{8.1.2,
  author    = {Paul Ruvolo and
               Eric Eaton},
  title     = {{ELLA:} An Efficient Lifelong Learning Algorithm},
  booktitle = {Proceedings of the 30th International Conference on Machine Learning},
  volume    = {28},
  pages     = {507--515},
  year      = {2013},
}

@incollection{3.1,
title = {Catastrophic Interference in Connectionist Networks: The Sequential Learning Problem},
series = {Psychology of Learning and Motivation},
publisher = {Academic Press},
volume = {24},
pages = {109-165},
year = {1989},
issn = {0079-7421},
author = {Michael McCloskey and Neal J. Cohen}
}

@article{3.2,
  title={Connectionist models of recognition memory: constraints imposed by learning and forgetting functions.},
  author={Ratcliff, Roger},
  journal={Psychological review},
  volume={97},
  number={2},
  pages={285},
  year={1990},
  publisher={American Psychological Association}
}

@article{1.2,
  author    = {German Ignacio Parisi and
               Ronald Kemker and
               Jose L. Part and
               Christopher Kanan and
               Stefan Wermter},
  title     = {Continual lifelong learning with neural networks: {A} review},
  journal   = {Neural Networks},
  volume    = {113},
  pages     = {54--71},
  year      = {2019},
}

@article{1.3,
  author    = {Matthias De Lange and
               Rahaf Aljundi and
               Marc Masana and
               Sarah Parisot and
               Xu Jia and
               Ales Leonardis and
               Gregory G. Slabaugh and
               Tinne Tuytelaars},
  title     = {A Continual Learning Survey: Defying Forgetting in Classification
               Tasks},
  journal   = {IEEE Transactions on Pattern Analysis and Machine Intelligence},
  volume    = {44},
  number    = {7},
  pages     = {3366--3385},
  year      = {2022},
}

@article{1.4,
  author    = {Gido M. van de Ven and
               Andreas S. Tolias},
  title     = {Three scenarios for continual learning},
  journal   = {CoRR},
  volume    = {abs/1904.07734},
  year      = {2019},
}

@article{8.1.4,
  author    = {Tom M. Mitchell and
               William W. Cohen and
               Estevam R. Hruschka Jr. and
               Partha P. Talukdar and
               Bo Yang and
               Justin Betteridge and
               Andrew Carlson and
               Bhavana Dalvi Mishra and
               Matt Gardner and
               Bryan Kisiel and
               Jayant Krishnamurthy and
               Ni Lao and
               Kathryn Mazaitis and
               Thahir Mohamed and
               Ndapandula Nakashole and
               Emmanouil A. Platanios and
               Alan Ritter and
               Mehdi Samadi and
               Burr Settles and
               Richard C. Wang and
               Derry Wijaya and
               Abhinav Gupta and
               Xinlei Chen and
               Abulhair Saparov and
               Malcolm Greaves and
               Joel Welling},
  title     = {Never-ending learning},
  journal   = {Commun. {ACM}},
  volume    = {61},
  number    = {5},
  pages     = {103--115},
  year      = {2018},
}

@article{4.1,
  title={Progressive neural networks},
  author={Rusu, Andrei A and Rabinowitz, Neil C and Desjardins, Guillaume and Soyer, Hubert and Kirkpatrick, James and Kavukcuoglu, Koray and Pascanu, Razvan and Hadsell, Raia},
  journal={arXiv preprint arXiv:1606.04671},
  year={2016}
}

@inproceedings{6.4,
  author    = {Hongjoon Ahn and
               Sungmin Cha and
               Donggyu Lee and
               Taesup Moon},
  title     = {Uncertainty-based Continual Learning with Adaptive Regularization},
  booktitle = {Advances in Neural Information Processing Systems},
  pages     = {4394--4404},
  year      = {2019},
}

@inproceedings{5.1.3,
  author    = {Jathushan Rajasegaran and
               Salman H. Khan and
               Munawar Hayat and
               Fahad Shahbaz Khan and
               Mubarak Shah},
  title     = {iTAML: An Incremental Task-Agnostic Meta-learning Approach},
  booktitle = {Proceedings of the 2020 {IEEE/CVF} Conference on Computer Vision and Pattern Recognition},
  pages     = {13585--13594},
  year      = {2020},
}

@inproceedings{7.1.6,
  author    = {Arslan Chaudhry and
               Marc'Aurelio Ranzato and
               Marcus Rohrbach and
               Mohamed Elhoseiny},
  title     = {Efficient Lifelong Learning with {A-GEM}},
  booktitle = {Proceedings of the 7th International Conference on Learning Representations},
  year      = {2019},
}

@inproceedings{7.1.2,
  author    = {David Lopez{-}Paz and
               Marc'Aurelio Ranzato},
  title     = {Gradient Episodic Memory for Continual Learning},
  booktitle = {Advances in Neural Information Processing Systems},
  pages     = {6467--6476},
  year      = {2017},
}

@inproceedings{9.2,
  author    = {Joan Serr{\`{a}} and
               Didac Suris and
               Marius Miron and
               Alexandros Karatzoglou},
  title     = {Overcoming Catastrophic Forgetting with Hard Attention to the Task},
  booktitle = {Proceedings of the 35th International Conference on Machine Learning},
  volume    = {80},
  pages     = {4555--4564},
  publisher = {{PMLR}},
  year      = {2018},
}

@inproceedings{5.1.1,
  author    = {Khurram Javed and
               Martha White},
  title     = {Meta-Learning Representations for Continual Learning},
  booktitle = {Advances in Neural Information Processing Systems},
  pages     = {1818--1828},
  year      = {2019},
}

@inproceedings{5.1.2,
  author    = {Matthew Riemer and
               Ignacio Cases and
               Robert Ajemian and
               Miao Liu and
               Irina Rish and
               Yuhai Tu and
               Gerald Tesauro},
  title     = {Learning to Learn without Forgetting by Maximizing Transfer and Minimizing
               Interference},
  booktitle = {Proceedings of the 7th International Conference on Learning Representations},
  year      = {2019},
}

@inproceedings{5.2.2,
  author    = {Ju Xu and
               Zhanxing Zhu},
  title     = {Reinforced Continual Learning},
  booktitle = {Advances in Neural Information Processing Systems},
  pages     = {907--916},
  year      = {2018},
}

@inproceedings{5.2.3,
  author    = {Christos Kaplanis and
               Murray Shanahan and
               Claudia Clopath},
  title     = {Policy Consolidation for Continual Reinforcement Learning},
  booktitle = {Proceedings of the 36th International Conference on Machine Learning},
  volume    = {97},
  pages     = {3242--3251},
  publisher = {{PMLR}},
  year      = {2019},
}

@article{7.1.1,
  author    = {Anthony V. Robins},
  title     = {Catastrophic Forgetting, Rehearsal and Pseudorehearsal},
  journal   = {Connect. Sci.},
  volume    = {7},
  number    = {2},
  pages     = {123--146},
  year      = {1995},
}

@inproceedings{7.1.3,
  author    = {Sylvestre{-}Alvise Rebuffi and
               Alexander Kolesnikov and
               Georg Sperl and
               Christoph H. Lampert},
  title     = {iCaRL: Incremental Classifier and Representation Learning},
  booktitle = {Proceedings of the 2017 {IEEE} Conference on Computer Vision and Pattern Recognition},
  pages     = {5533--5542},
  publisher = {{IEEE} Computer Society},
  year      = {2017},
}

@article{7.1.10,
  author    = {Arslan Chaudhry and
               Marcus Rohrbach and
               Mohamed Elhoseiny and
               Thalaiyasingam Ajanthan and
               Puneet Kumar Dokania and
               Philip H. S. Torr and
               Marc'Aurelio Ranzato},
  title     = {Continual Learning with Tiny Episodic Memories},
  journal   = {CoRR},
  volume    = {abs/1902.10486},
  year      = {2019}
}

@inproceedings{7.1.11,
  author    = {Arun Mallya and
               Svetlana Lazebnik},
  title     = {PackNet: Adding Multiple Tasks to a Single Network by Iterative Pruning},
  booktitle = {Proceedings of the 2018 {IEEE} Conference on Computer Vision and Pattern Recognition},
  pages     = {7765--7773},
  year      = {2018}
}

@inproceedings{4.3,
  author    = {Jaehong Yoon and
               Eunho Yang and
               Jeongtae Lee and
               Sung Ju Hwang},
  title     = {Lifelong Learning with Dynamically Expandable Networks},
  booktitle = {Proceedings of the 6th International Conference on Learning Representations},
  year      = {2018},
}

@inproceedings{4.7,
  author    = {Jaehong Yoon and
               Saehoon Kim and
               Eunho Yang and
               Sung Ju Hwang},
  title     = {Scalable and Order-robust Continual Learning with Additive Parameter
               Decomposition},
  booktitle = {Proceedings of the 8th International Conference on Learning Representations},
  year      = {2020},
}

@inproceedings{9.12,
  title={Supermasks in superposition},
  author={Wortsman, Mitchell and Ramanujan, Vivek and Liu, Rosanne and Kembhavi, Aniruddha and Rastegari, Mohammad and Yosinski, Jason and Farhadi, Ali},
  journal={Advances in Neural Information Processing Systems},
  volume={33},
  pages={15173--15184},
  year={2020}
}

@article{12.1,
  title={Calculation of the Wasserstein distance between probability distributions on the line},
  author={Vallender, SS},
  journal={Theory of Probability \& Its Applications},
  volume={18},
  number={4},
  pages={784--786},
  year={1974},
  publisher={SIAM}
}

@article{12.2,
  title={Statistical aspects of Wasserstein distances},
  author={Panaretos, Victor M and Zemel, Yoav},
  journal={Annual review of statistics and its application},
  volume={6},
  pages={405--431},
  year={2019},
  publisher={Annual Reviews}
}

@book{13.2,
  title={Mathematics for machine learning},
  author={Deisenroth, Marc Peter and Faisal, A Aldo and Ong, Cheng Soon},
  year={2020},
  publisher={Cambridge University Press}
}

@article{n70,
  title={A survey on negative transfer},
  author={Zhang, Wen and Deng, Lingfei and Zhang, Lei and Wu, Dongrui},
  journal={IEEE/CAA Journal of Automatica Sinica},
  volume={10},
  number={2},
  pages={305--329},
  year={2022},
  publisher={IEEE}
}

@inproceedings{n71,
  title={Beyond not-forgetting: Continual learning with backward knowledge transfer},
  author={Lin, Sen and Yang, Li and Fan, Deliang and Zhang, Junshan and et al.},
  booktitle={Advances in Neural Information Processing Systems},
  volume={35},
  pages={16165--16177},
  year={2022}
}

@inproceedings{n72,
  title={A theory for knowledge transfer in continual learning},
  author={Prado, Diana Benavides and Riddle, Patricia},
  booktitle={Conference on Lifelong Learning Agents},
  pages={647--660},
  year={2022},
  organization={PMLR}
}

@inproceedings{n74,
  title={Afec: Active forgetting of negative transfer in continual learning},
  author={Wang, Liyuan and Zhang, Mingtian and Jia, Zhongfan and Li, Qian and Bao, Chenglong and Ma, Kaisheng and Zhu, Jun and Zhong, Yi},
  booktitle={Advances in Neural Information Processing Systems},
  volume={34},
  pages={22379--22391},
  year={2021}
}

@Techreport{D-2,
  author = {Krizhevsky, Alex and Hinton, Geoffrey},
 address = {Toronto, Ontario},
 institution = {University of Toronto},
 number = {0},
 publisher = {Technical report, University of Toronto},
 title = {Learning multiple layers of features from tiny images},
 year = {2009},
 title_with_no_special_chars = {Learning multiple layers of features from tiny images}
}

@inproceedings{D-3,
  author    = {Oriol Vinyals and
               Charles Blundell and
               Tim Lillicrap and
               Koray Kavukcuoglu and
               Daan Wierstra},
  title     = {Matching Networks for One Shot Learning},
  booktitle = {Advances in Neural Information Processing Systems},
  year      = {2016}
}

@inproceedings{D-4,
  author    = {Sayna Ebrahimi and
               Franziska Meier and
               Roberto Calandra and
               Trevor Darrell and
               Marcus Rohrbach},
  title     = {Adversarial Continual Learning},
  booktitle = {Proceedings of the 16th European Conference on Computer Vision},
  year      = {2020}
}

@inproceedings{D-7,
title	= {Reading Digits in Natural Images with Unsupervised Feature Learning},
author	= {Yuval Netzer and Tao Wang and Adam Coates and Alessandro Bissacco and Bo Wu and Andrew Y. Ng},
year	= {2011},
booktitle	= {Advances in Neural Information Processing Systems Workshop}
}

@techreport{D-8,
  title={Notmnist dataset. Google (Books/OCR)},
  author={Bulatov, Yaroslav},
  year={2011},
  institution={Tech. Rep.}
}

@article{D-9,
  title={Fashion-mnist: a novel image dataset for benchmarking machine learning algorithms},
  author={Xiao, Han and Rasul, Kashif and Vollgraf, Roland},
  journal={arXiv preprint arXiv:1708.07747},
  year={2017}
}

@inproceedings{N-5,
  author    = {Yen{-}Chang Hsu and
               Yen{-}Cheng Liu and
               Zsolt Kira},
  title     = {Re-evaluating Continual Learning Scenarios: {A} Categorization and
               Case for Strong Baselines},
  booktitle   = {Advances in Neural Information Processing Systems Workshop},
  year      = {2018},
}

@article{N-6,
  author    = {Sebastian Farquhar and
               Yarin Gal},
  title     = {Towards Robust Evaluations of Continual Learning},
  journal={arXiv preprint arXiv:1805.09733},
  year      = {2018},
}

@article{n84,
  title={Leaf: A benchmark for federated settings},
  author={Caldas, Sebastian and Duddu, Sai Meher Karthik and Wu, Peter and Li, Tian and Kone{\v{c}}n{\`y}, Jakub and McMahan, H Brendan and Smith, Virginia and Talwalkar, Ameet},
  journal={arXiv preprint arXiv:1812.01097},
  year={2018}
}

@article{n85,
  title={Gradient-based learning applied to document recognition},
  author={LeCun, Yann and Bottou, L{\'e}on and Bengio, Yoshua and Haffner, Patrick},
  journal={Proceedings of the IEEE},
  volume={86},
  number={11},
  pages={2278--2324},
  year={1998},
  publisher={Ieee}
}

@inproceedings{n91,
  title={Piggyback: Adapting a single network to multiple tasks by learning to mask weights},
  author={Mallya, Arun and Davis, Dillon and Lazebnik, Svetlana},
  booktitle={Proceedings of the European conference on computer vision},
  pages={67--82},
  year={2018}
}

@inproceedings{n93,
  title={Lifelong Learning of Task-Parameter Relationships for Knowledge Transfer},
  author={Srivastava, Shikhar and Yaqub, Mohammad and Nandakumar, Karthik},
  booktitle={Proceedings of the IEEE/CVF Conference on Computer Vision and Pattern Recognition},
  pages={2524--2533},
  year={2023}
}

@inproceedings{n94,
  title={The Lottery Ticket Hypothesis: Finding Sparse, Trainable Neural Networks},
  author={Frankle, Jonathan and Carbin, Michael},
  booktitle={Proceedings of the 6th International Conference on Learning Representations},
  year={2018}
}

@inproceedings{n95,
  title={Forget-free continual learning with winning subnetworks},
  author={Kang, Haeyong and Mina, Rusty John Lloyd and Madjid, Sultan Rizky Hikmawan and Yoon, Jaehong and Hasegawa-Johnson, Mark and Hwang, Sung Ju and Yoo, Chang D},
  booktitle={Proceedings of the 39th International Conference on Machine Learning},
  pages={10734--10750},
  year={2022},
}

@article{n96,
  title={Estimating or propagating gradients through stochastic neurons for conditional computation},
  author={Bengio, Yoshua and L{\'e}onard, Nicholas and Courville, Aaron},
  journal={arXiv preprint arXiv:1308.3432},
  year={2013}
}

@inproceedings{n97,
  title={Imagenet classification with deep convolutional neural networks},
  author={Krizhevsky, Alex and Sutskever, Ilya and Hinton, Geoffrey E},
  booktitle={Advances in neural information processing systems},
  volume={25},
  year={2012}
}

@inproceedings{n98,
  title={CoSCL: Cooperation of small continual learners is stronger than a big one},
  author={Wang, Liyuan and Zhang, Xingxing and Li, Qian and Zhu, Jun and Zhong, Yi},
  booktitle={European Conference on Computer Vision},
  pages={254--271},
  year={2022},
  organization={Springer}
}

@article{n99,
  title={Incorporating neuro-inspired adaptability for continual learning in artificial intelligence},
  author={Wang, Liyuan and Zhang, Xingxing and Li, Qian and Zhang, Mingtian and Su, Hang and Zhu, Jun and Zhong, Yi},
  journal={Nature Machine Intelligence},
  volume={5},
  number={12},
  pages={1356--1368},
  year={2023},
  publisher={Nature Publishing Group UK London}
}

@article{n100,
  title={A comprehensive survey of continual learning: Theory, method and application},
  author={Wang, Liyuan and Zhang, Xingxing and Su, Hang and Zhu, Jun},
  journal={IEEE Transactions on Pattern Analysis and Machine Intelligence},
  number={01},
  pages={1--20},
  year={2024},
}

@article{n101,
  author       = {Thang Doan and
                  Seyed{-}Iman Mirzadeh and
                  Joelle Pineau and
                  Mehrdad Farajtabar},
  title        = {Efficient Continual Learning Ensembles in Neural Network Subspaces},
  journal      ={arXiv preprint arXiv:2202.09826},
  year         = {2022},
}

@article{n300,
  title={Class-incremental learning: survey and performance evaluation on image classification},
  author={Masana, Marc and Liu, Xialei and Twardowski, Bart{\l}omiej and Menta, Mikel and Bagdanov, Andrew D and Van De Weijer, Joost},
  journal={IEEE Transactions on Pattern Analysis and Machine Intelligence},
  volume={45},
  number={5},
  pages={5513--5533},
  year={2022},
  publisher={IEEE}
}
\vspace{-1.2cm}
\begin{IEEEbiography}[{\includegraphics[width=1in,height=1.25in,clip,keepaspectratio]{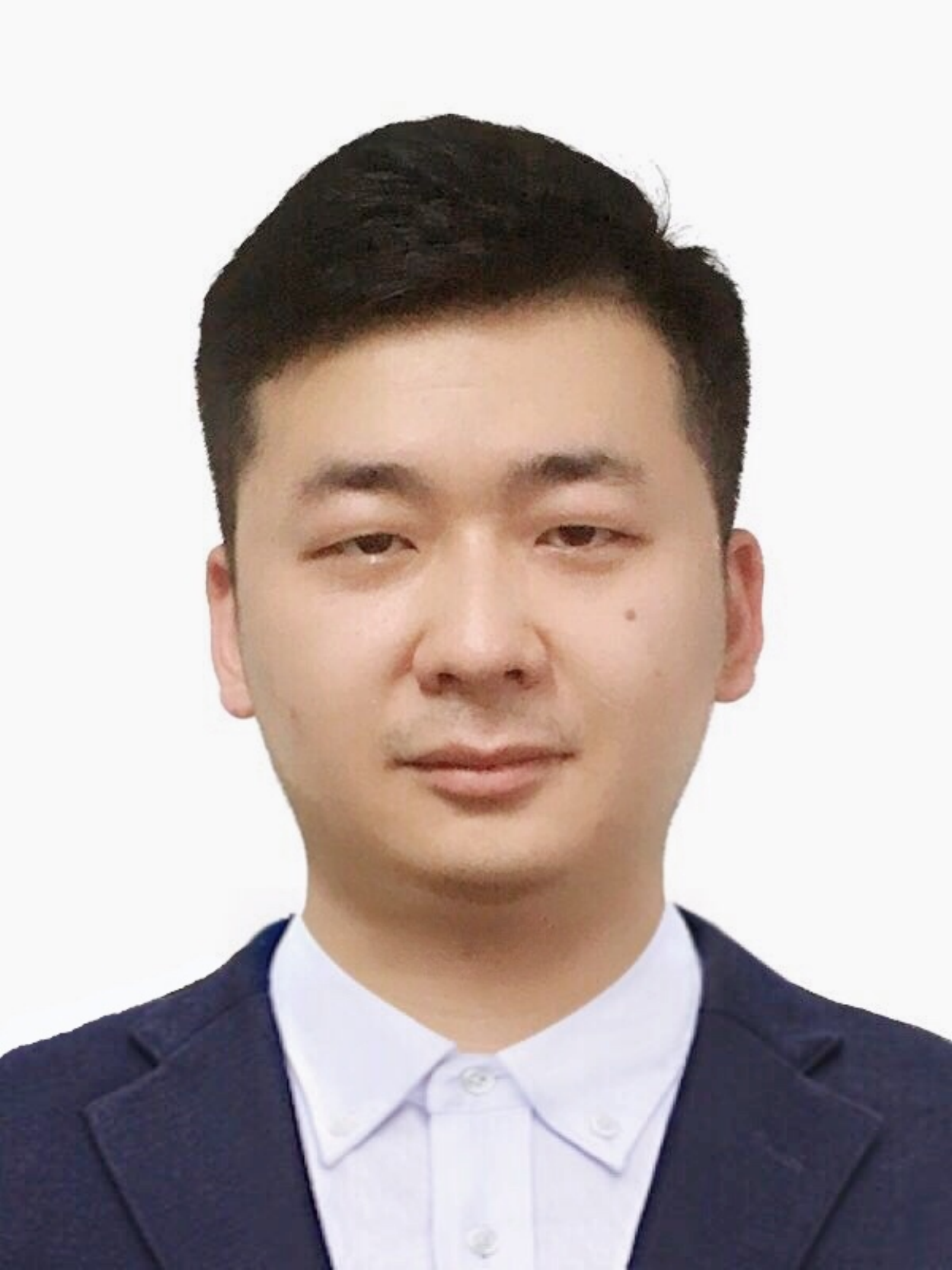}}]{Zhi~Wang} received the double BS in computer science and mathematics from Zhaoqing University, Zhaoqing, China,  and received the MS in computer technology from Xidian University, Xi’an, China in 2018. He is currently working toward a PhD degree at the School of Computer Science and Technology, Xidian University, China. His current research interests include data mining, machine learning, lifelong/continual learning, representation and clustering of multivariate time series.
\end{IEEEbiography}

\begin{IEEEbiography}[{\includegraphics[width=1in,height=1.25in,clip,keepaspectratio]{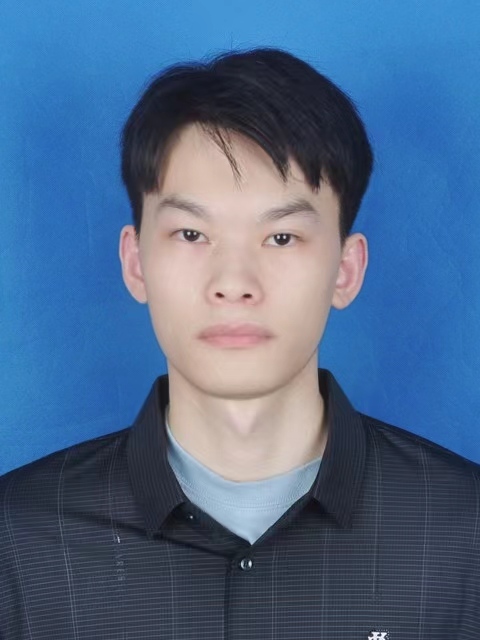}}]{Zhongbin~Wu} received the BS in software engineering from Nanchang University, Jiangxi, China. He is currently working toward an MS degree at the School of Computer Science and Technology, Xidian University, China. His current research interests include data mining, machine learning, lifelong/continual learning, representation and clustering of multivariate time series.
\end{IEEEbiography}

\begin{IEEEbiography}[{\includegraphics[width=1in,height=1.25in,clip,keepaspectratio]{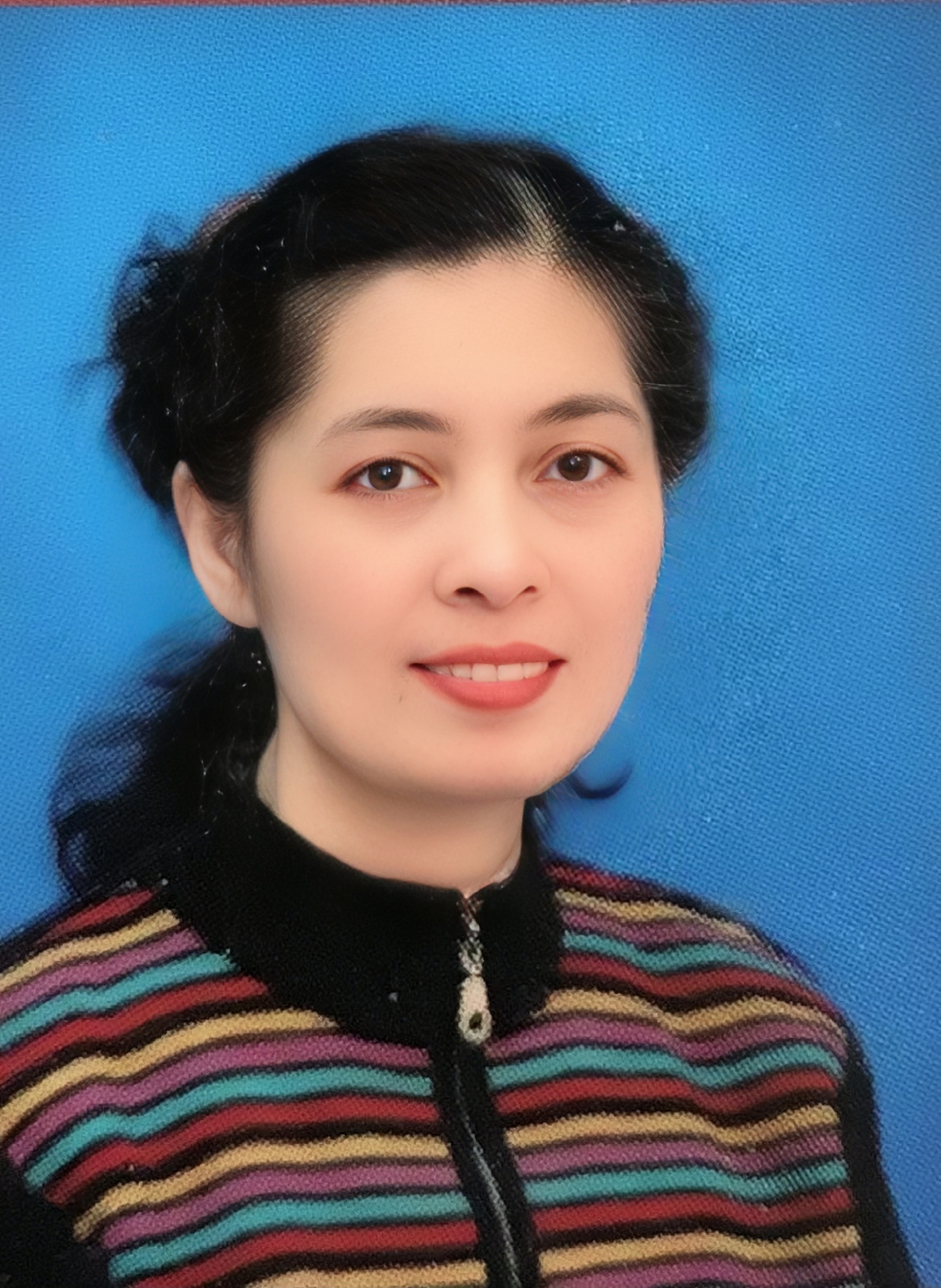}}]{Yanni~Li} received the MS and PhD degrees in computer science and technology from Xidian University, Xi’an, China. She is a professor at the School of Computer Science and Technology of Xidian University. Her current research interests include big data analysis, data mining, machine learning, lifelong/continual learning, large-scale combinatorial optimization, etc. In the past five years, she has published more than 20 papers as the first author at top academic conferences or in journals.
\end{IEEEbiography}

\begin{IEEEbiography}[{\includegraphics[width=1in,height=1.25in,clip,keepaspectratio]{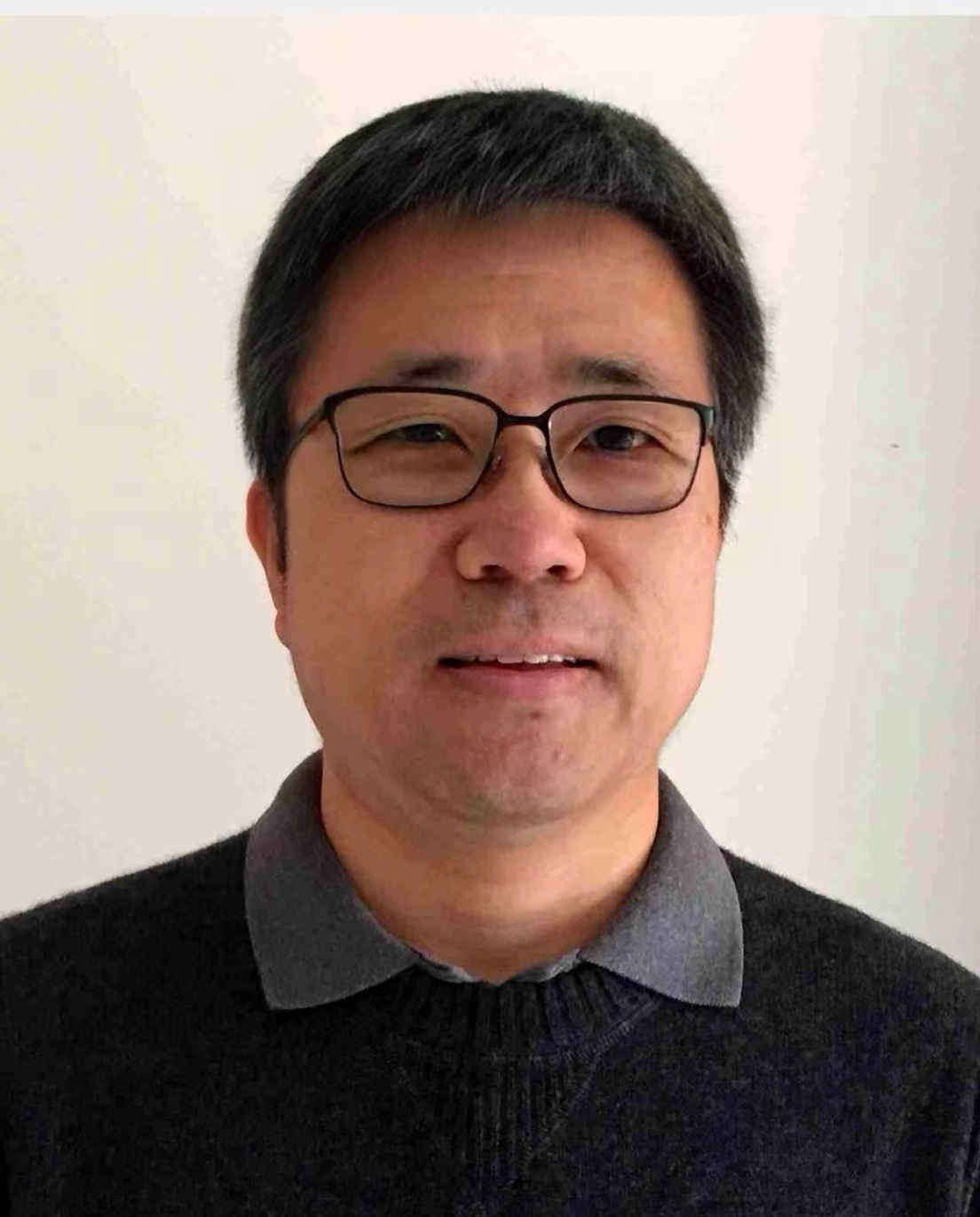}}]{Bing~Liu} (Fellow, IEEE) received the PhD degree in artificial intelligence from the University of Edinburgh. He is a distinguished professor at the University of Illinois Chicago. His research interests include lifelong/continual machine learning, sentiment analysis and opinion mining, data mining, machine learning, and natural language processing. He has published extensively at top conferences and in journals in these areas. Two of his papers have received 10-year test-of-time awards from KDD, the premier conference on data mining and data science. He also authored five books: one on lifelong machine learning, one on Web data mining, two on sentiment analysis, and one on lifelong dialogue systems. Some of his work has been widely reported in the press, including a front-page article in the New York Times. On professional services, he served as the Chair of ACM SIGKDD from 2013–2017. He has served as program chair of many leading data mining conferences, including KDD, ICDM, CIKM, WSDM, SDM, and PAKDD. He is a Fellow of ACM, IEEE, and AAAI.
\end{IEEEbiography}


\vspace{-5.5cm}
\begin{IEEEbiography}[{\includegraphics[width=1.2in,height=1.2in,clip,keepaspectratio]{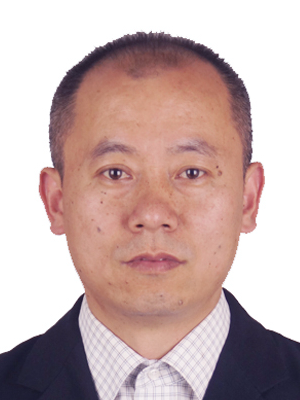}}]{Guangxin Li} received the M.S. degree from the School of Mechano-Electronic Engineering, and the Ph.D. degree from the School of Compute Science and Technology, Xidian University, Xi’an, China, respectively. He was a Visiting Scholar with the School of Electrical and Computer Engineering, Oklahoma State University, Stillwater, OK, USA. He is currently an Associate Professor with the School of Compute Science and Technology, Xidian University. His research interests include data analysis, image processing, computer graphics, etc.
\end{IEEEbiography}

\vspace{-5cm}
\begin{IEEEbiography}[{\includegraphics[width=1.2in,height=1.2in,clip,keepaspectratio]{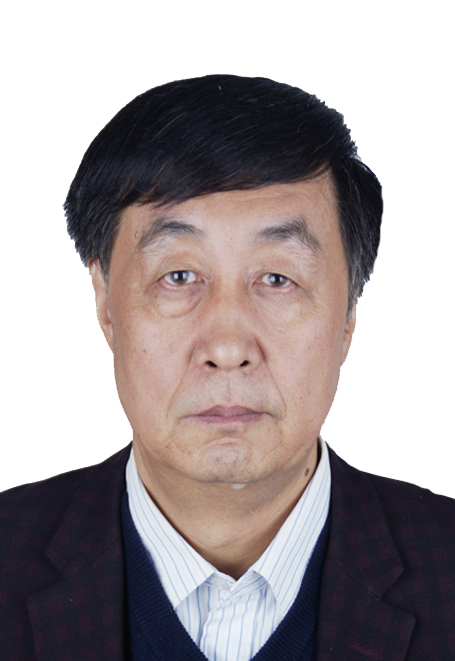}}]{Yuping Wang} received the Ph.D degree in Computation Mathematics from Xi’an Jiaotong University, Xi’an, China, in 1993. He is a distinguished professor at Xidian University, Xi’an, China. He is a senior member of IEEE. His current research interests include machine learning, optimization algorithms and modeling for engineering problems, network task scheduling. He has published more than 200 papers.
\end{IEEEbiography}

\newpage

 
\vspace{11pt}


\vspace{11pt}


\clearpage
\onecolumn
\appendices
\section{Theorem 1 and The Proof}
\textbf{Theorem 1.} The theoretical bounds for FWT and BWT of tasks $i$ and $t$ ($i < t$, $i \in [1,t-1]$, $t \in [2, T]$)  in TIL are as follows.
\begin{equation*}
\left\{\begin{matrix}
\begin{array}{l}
\begin{aligned}
&FWT: \epsilon_{t}(h)\leq \epsilon_{i}(h)+d(\mathcal{D'}_{i},\mathcal{D'}_{t}) + \textup{min}\{\mathbb{E}_{\mathbf{x} \sim \mathcal{D'}_{i}}\left [|l_{i}(\mathbf{x})-l_{t}(\mathbf{x})|\right ],\mathbb{E}_{\mathbf{x} \sim \mathcal{D'}_{t}}\left [|l_{i}(\mathbf{x})-l_{t}(\mathbf{x})|\right ]\}\\
&\;\\
&BWT: \epsilon_{i}'(h)\leq \epsilon_{t}(h)+d(\mathcal{D'}_{i},\mathcal{D'}_{t}) + \textup{min}\{\mathbb{E}_{\mathbf{x} \sim \mathcal{D'}_{i}}\left [|l_{i}(\mathbf{x})-l_{t}(\mathbf{x})|\right ],\mathbb{E}_{\mathbf{x} \sim \mathcal{D'}_{t}}\left [|l_{i}(\mathbf{x})-l_{t}(\mathbf{x})|\right ]\}
\end{aligned}
\end{array}
\end{matrix}\right.
\end{equation*}

\begin{proof}
{\color{black}Recall that $\epsilon_{t}(h)=\epsilon_{t}(h, l_t)$ and $\epsilon_{i}(h)=\epsilon_{i}(h, l_i)$. Let $m_i$ and $m_t$ be the density functions of $\mathcal{D'}_i$ and $\mathcal{D'}_t$ respectively. For the theoretical bound of  FWT,} 
\begin{equation}
\begin{split}
\label{eq15}
\epsilon_{t}(h) &= \epsilon_{t}(h) + \epsilon_{i}(h) - \epsilon_{i}(h) + \epsilon_{i}(h, l_t) - \epsilon_{i}(h, l_t) \\
&\leq \epsilon_{i}(h) + |\epsilon_{i}(h, l_t)-\epsilon_{i}(h, l_i)| + |\epsilon_{t}(h, l_t)-\epsilon_{i}(h, l_t)| \\
&\leq \epsilon_{i}(h) + \mathbb{E}_{\mathbf{x} \sim \mathcal{D'}_{i}}[|l_{i}(\mathbf{x})-l_{t}(\mathbf{x})|] + |\epsilon_{t}(h, l_t)-\epsilon_{i}(h, l_t)| \\
&\leq \epsilon_{i}(h) + \mathbb{E}_{\mathbf{x} \sim \mathcal{D'}_{i}}[|l_{i}(\mathbf{x})-l_{t}(\mathbf{x})|] + \int |m _i(\mathbf{x})-m _t(\mathbf{x})||h(\mathbf{x})-l_t(\mathbf{x})|d\mathbf{x} \\
&\leq \epsilon_{i}(h) + \mathbb{E}_{\mathbf{x} \sim \mathcal{D'}_{i}}[|l_{i}(\mathbf{x})-l_{t}(\mathbf{x})|] + d(\mathcal{D'}_{i},\mathcal{D'}_{t}).
\end{split}
\end{equation}

{\color{black}For the theoretical bound of BWT, in the first line of the above Eq. (\ref{eq15}), we can instead choose to add and subtract $\epsilon_{t}(h, l_t)$ rather than $\epsilon_{i}(h, l_t)$, which would result in the same bound only with the expectation taken with respect to $\mathcal{D'}_{t}$ instead of $\mathcal{D'}_{i}$. Choosing the smaller of the two gives us the bound of BWT.}
\end{proof}

\begin{table}[ht]
\centering
\caption{Datasets, network architectures and hyperparameters of the proposed ETCL.}\label{ETCL-T4}
\renewcommand\arraystretch{1.6}
\resizebox{0.7\linewidth}{!}{
\begin{threeparttable}
    \begin{tabular}{|l|c|c|c|c|c|c|p{1cm}<{\centering}|}
    \hline
    \normalsize{\textbf{Datasets}} & \textbf{Backbone} & \textbf{Batch Size} &  \textbf{Epochs} & \textbf{$\lambda$} & \textbf{Optimizer}  & $c$& \multicolumn{1}{c|}{\boldmath$\delta$} \\\hline
    PMNIST & 3-Layers FCN & 10 & 5 & 0.001 & SGD & 0.5 & 0.80\\\hline
    CIFAR 100  & AlexNet    & 64 & 100   & 0.001 & SGD & 0.5 & 0.70\\\hline
    CIFAR 100 Sup& LeNet-5 & 64 & 100   & 0.001   & SGD & 0.5 & 0.60 \\\hline
    MiniImageNet    & ResNet 18  & 64 & 200   & 0.100  & SGD & 0.5  & 0.95  \\\hline
    5-Datasets & ResNet 18 & 64 & 100   & 0.100  &  SGD & 0.5  & 0.50  \\\hline
    F-EMNIST-1  & ResNet 18 & 64 & 50 & 0.100 & SGD &  0.4 & 0.57 \\\hline
    F-EMNIST-2  & ResNet 18 & 64 & 50 & 0.100 & SGD &  0.4  & 0.58 \\\hline
    F-CelebA-1  & ResNet 18 & 16 & 100 & 0.100 & SGD  & 0.4 & 0.65 \\\hline
    F-CelebA-2  & ResNet 18 & 16 & 100 & 0.100 & SGD & 0.4 & 0.77\\\hline
    \multirow{2}{*}{(EMNIST, F-EMNIST-1)} & 3-Layers FCN & 64 & 50 & 0.010 & SGD & 0.5 & 0.58 \\ \cline{2-8}
     & AlexNet & 128 & 200 & 0.001 & SGD  & 0.5 & 0.60 \\\hline
    \multirow{2}{*}{(CIFAR 100, F-CelebA-1)} & 3-Layers FCN & 64 & 50 & 0.010 & SGD & 0.2 & 0.70 \\ \cline{2-8}
     & AlexNet & 64 & 200 & 0.001 & SGD  & 0.4 & 0.60 \\\hline
    
    \end{tabular}
\begin{tablenotes}
\small
\item[1]$\lambda$ is the learning rate, and $c$ is the target/model layer-wise capacity ratio.
\item[2]$\delta$ is the empirical distance threshold in Eq.(13) on each dataset.
\end{tablenotes}
\end{threeparttable}}
\end{table}

\section{Network Architecture and Hyperparameters of ETCL on Various Datasets}

\subsection{Network Architecture and Hyperparameters of the Proposed ETCL}

To test the efficacy and scalability of our method, we use various DNN models/backbones on the 11 image classification benchmark datasets. We use a 3-Layer fully connected network (FCN) with two hidden layers of 100 units each for PMINIST, (EMNIST, F-EMNIST-1) and (CIFAR 100, F-CelebA-1) following \cite{7.1.2} and \cite{n85}. For experiments with CIFAR 100 we use a 5-Layer AlexNet following \cite{9.2}. For experiments with CIFAR-100 Sup, we use a 5-Layer LeNet-5 \cite{4.7}. For experiments with MiniImageNet, 5-Datasets, F-EMNIST-1 and F-EMNIST-2, F-CelebA-1, and F-CelebA-2, similar to \cite{7.1.10}, we use a reduced ResNet-18 architecture \cite{7.1.11}. For PMNIST, F-CelebA-1, and F-CelebA-2, we evaluate and compare our ETCL in the `single-head' setting \cite{N-5,N-6} where all tasks share the final classifier layer and inference is performed without a task hint. For all other experiments, we evaluate our ETCL in the `multi-head' setting, where each task has a separate head or classifier. The correspondence between the training dataset and its network structure, as well as the training hyperparameters used by each network structure, are shown in Table \ref{ETCL-T4}. Moreover, Table \ref{ETCL-T4} reveals: 1) the hyperparameter $c$~(\%) in our ETCL has good stability for different backbones and datasets; 2) all the values of hyperparameter $\delta$ are greater than or equal to 0.5 ($\delta \in [0,1])$ on the 11 different datasets, which experimentally verifies Theorem 2.

\subsection{Computing Platform} 

All of the experiments were conducted on the platform:  Intel(R) Xeon(R) Gold 6230 CPU 2.10GHz, 251GB RAM, and GPU - GeForce RTX 2080 Ti with 12GB MC (graphics card Memory Capacity). And all the experimental results are averages with standard deviation values over 5 different runs with 5 random seeds. 

\section{Datesets Details}

{\color{black}Eleven benchmark image classification datasets are used in our experiments, which are divided into the following categories:

\textbf{Dissimilar tasks datasets}. (1) PMNIST  (Permuted MNIST, 10 tasks) \cite{n85}. It is a variant of the MNIST dataset where each task is considered as a random permutation of the original MNIST pixels. We create 10 sequential tasks using different permutations where each task has 10 classes. (2) CIFAR-100 (10 tasks) \cite{D-2}. It is constructed by randomly splitting 100 classes of CIFAR-100 \cite{D-2} into 10 tasks with 10 classes per task. (3) CIFAR 100 Sup 
(20 tasks) \cite{D-2}: It is constructed by splitting 100 classes of CIFAR 100 into 20 tasks with 5 classes of the same attributes per task. (4) MiniImageNet (20 tasks) \cite{D-3}: It is constructed by splitting 100 classes of miniImageNet into 20 sequential tasks where each task has 5 classes. (5) 5-Datasets (5 tasks) \cite{D-4}: It includes CIFAR-10, MINIST, SVHN \cite{D-7}, notMNIST \cite{D-8} and Fashion MNIST \cite{D-9}, where the classification of each dataset is considered as a task.

\textbf{Similar tasks datasets}. (1) F-EMINIST-1 (10 tasks) and (2) F-EMINIST-2 (35 tasks). They are similar task datasets from \textit{federated learning}, which are constructed by randomly choosing 10/35 tasks from two publicly available federated learning datasets \cite{n84}. (3) F-CelebA-1 (10 tasks) and (4) F-CelebA-2 (20 tasks). They are also similar task datasets from \textit{federated learning}, which are constructed by randomly choosing 10/20 tasks from two publicly available federated learning datasets \cite{n84}. Each of the 10/20 tasks contains images of a celebrity labeled by whether he/she is smiling or not. Note that for the four similar tasks datasets (1)-(4), the training and testing sets are already provided in \cite{n84}. We further split about 10\% of the original training set and kept it for validation purposes.

\textbf{Mixed tasks datasets.} (1) (EMNIST, F-EMNIST-1) (20 tasks). It is a randomly mixed sequence of similar and dissimilar tasks constructed from EMNIST \cite{n85} and F-EMNIST-1. (2) (CIFAR-100, F-CelebA-1)(20 tasks). It is a randomly mixed sequence of similar and dissimilar tasks constructed from CIFAR-100 (10 tasks) and F-EMNIST-1 (10 tasks).

The sample sizes of the training/{\color{black}validation}/testing are as follows: (1) PMNIST 6000 / 300 / 700, (2) CIFAR 100 5000/300/700, (3) CIFAR 100 Sup 5000 / 300 / 700, (4) MiniImageNet 5000 / 200 / 800, and (5) 5-Datasets, which has 5 tasks in total, with the samples of each task being 50000 / 10000 / 10000, 50000 / 10000 / 10000, 63257 / 10000 / 26032, 50000 / 10000 / 10000, and 10000/6854/1872 respectively.}

{\color{black}
\section{Related Baselines Details}

CAF-MAS \cite{n99}: CAF-MAS is the best-performing combined model of CoSCL\cite{n98} or CAF\cite{n99} with MAS \cite{6.3}, in which the mechanisms of method CAF are embedded within the representative experience-replay method MAS.

GPM \cite{7.2.4}: It is an OG-based TIL method where a neural network learns new tasks by taking gradient steps in the orthogonal direction to the gradient sub-spaces deemed important for the past tasks. It finds the bases of these sub-spaces by analyzing network representations after learning each task with Singular Value Decomposition (SVD) in a single shot manner and storing them in the memory as Gradient Projection Memory (GPM). Qualitative and quantitative analyses show that such an orthogonal gradient descent induces minimum to no interference with past tasks, thereby mitigating forgetting.

HAT \cite{9.2}: HAT proposes a task-based hard attention mechanism that preserves previous tasks’ information without affecting the current task’s learning. A hard attention mask is learned concurrently with each task, through a stochastic gradient descent, and previous masks are exploited to condition such learning. It is shown that the proposed mechanism is effective in  reducing catastrophic forgetting. 

CAT \cite{9.3}: CAT uses binary masks of neurons in HAT to achieve CF prevention and employs a separate model to perform task similarity detection for its FKT and BKT. Specifically, CAT proposes a new TIL method to learn both types of tasks in the same network. For dissimilar tasks, CAT focuses on dealing with forgetting, and for similar tasks, CAT focuses on selectively transferring the knowledge learned from some similar previous tasks to improve the new task learning.

WSN \cite{n95}: It is a new TIL method referred to as Winning SubNetworks (WSN), which jointly learns the model weights and task-adaptive binary masks pertaining to sub-networks associated with each task, and by reusing weights of the prior sub-networks, WSN achieves forgetting-free and FKT.

TRGP \cite{7.2.5}: Based on the GPM method, TRGP is the OG-based method and it selects the most related old tasks within the ``trust region" for the new task, and then reuses the frozen weights in layer-wise scaling matrices to jointly optimize the matrices and model to achieve its FKT.

CUBER \cite{n71}: On the basis of GPM and TRGP, the OG-based method CUBER first analyzes the conditions under which updating the learned model of old tasks could lead to BKT. It then proposes a new method for FKT and BKT.}

\section{Additional Experimental Results}
\subsection{Time and Space Comparisons}

\begin{figure}[h]
  \centering
  \includegraphics[width=0.85\textwidth]{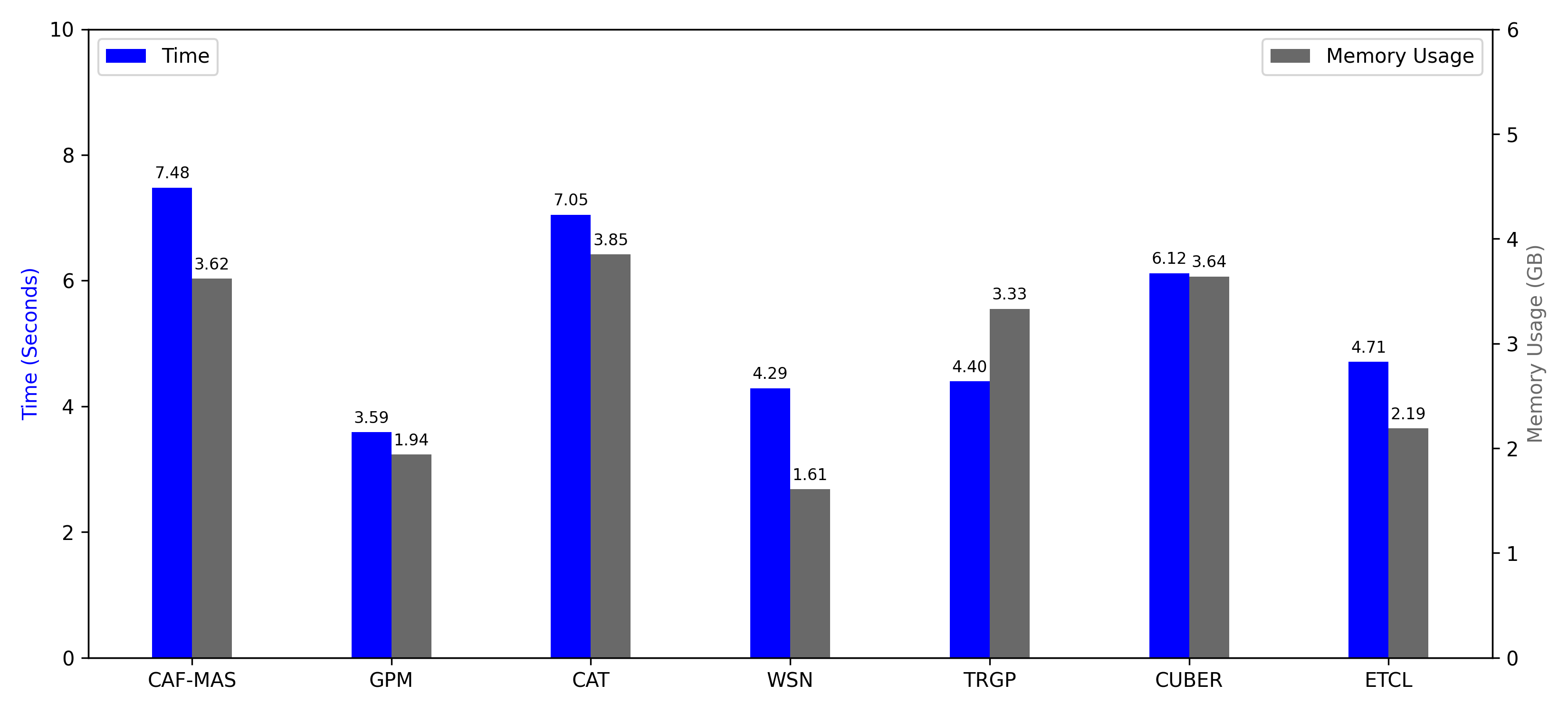}
  \caption{The average time and memory usage comparisons of ETCL and SOTA baselines on 11 benchmark datasets.}
  \label{TSC}
\end{figure}

To verify and compare the efficiency of the proposed ETCL that of with baselines in time and memory required for the model training, we conducted the efficiency comparison experiments in terms of the time spent per epoch, and the amount of memory used by the baselines. The time and space comparisons of ETCL with some SOTA with/without KT baselines are shown in Table \ref{ETCL-TS}, and the average time and memory usage comparisons of ETCL and SOTA baselines on 11 benchmark datasets are shown in Figure \ref{TSC}. 

Note that although the OG-based method GPM has no explicit KT mechanism, GPM can perform KT (see Tables \ref{KT1}-\ref{KT2}). Both TRGP and CUBER are OG-based methods built upon GPM, where TRGP only has the FKT function, while CUBER can perform both forward and backward KT. Both CAT and WSN are mask-based methods, but WSN cannot do BKT. The masks in ETCL are similar to those of CAT and WSN for dealing with CF. However, ETCL's main techniques for achieving positive KT: task similarity detection, either positive  FKT or BKT mechanisms are different from those of the above KT methods. As simultaneous processing of FKT and BKT requires more processing time and memory, Table \ref{ETCL-TS} shows that ETCL is much better than CAT and CUBER with simultaneous FKT and BKT functions, in terms of time and space training performances. It is worth noting that as CAF-MAS is an ensemble model based TIL method using $k$ CL learners ($k=5$), the experimental results show that CAF-MAS has high time and space complexities. 

\begin{table}[h]
\caption{The efficiency and memory comparisons of ETCL and SOTA baselines with/without the KT mechanism.}\label{ETCL-TS}
\renewcommand\arraystretch{1.6}
\centering
\small
\setlength{\tabcolsep}{3mm}{\resizebox{\textwidth}{!}{
\begin{tabular}{|l?r|r?r|r?r|r?r|r?r|r?r|r?r|r|}
\hline
\multicolumn{1}{|c?}{\multirow{2}{*}{ {\textbf{Datasets}}}}  & \multicolumn{2}{c?}{CAF-MAS} & \multicolumn{2}{c?}{GPM} & \multicolumn{2}{c?}{CAT} & \multicolumn{2}{c?}{WSN} & \multicolumn{2}{c?}{TRGP} & \multicolumn{2}{c?}{CUBER}& \multicolumn{2}{c|}{ETCL(\textbf{Ours})} \\\cline{2-15}
\multicolumn{1}{|c?}{}     & \multicolumn{1}{c|}{{ T(S)}} & \multicolumn{1}{c?}{{ M(G)}} & \multicolumn{1}{c|}{{ T(S)}} & \multicolumn{1}{c?}{{ M(G)}} & \multicolumn{1}{c|}{{ T(S)}} & \multicolumn{1}{c?}{{ M(G)}} & \multicolumn{1}{c|}{{ T(S)}} & \multicolumn{1}{c?}{{ M(G)}}& \multicolumn{1}{c|}{{ T(S)}} & \multicolumn{1}{c?}{{ M(G)}}& \multicolumn{1}{c|}{{ T(S)}} & \multicolumn{1}{c?}{{ M(G)}}& \multicolumn{1}{c|}{{ T(S)}} & \multicolumn{1}{c|}{{ M(G)}}\\\hline
PMNIST  & 22.5 & 3.34 & \textbf{10.17} & 1.29 & 15.32 & 3.92& 21.38& \textbf{0.59}& 13.22 & 1.36 & 16.68 & 1.52 &  21.61 & 0.60  \\\hline
CIFAR100  & 6.67 & 2.55  & \textbf{2.51} & 1.59 & 3.28   & 4.22 & 4.96 & \textbf{1.33} & 2.82 & 4.91 & 2.91 & 5.11 &  6.36 & 3.93 \\\hline
CIFAR100 Sup  & 3.42 & 2.67 & 1.59 & 1.53 & 2.41 & 4.22 &1.26 & \textbf{0.96}& 1.62 & 1.61 & 1.87 & 1.92  & \textbf{0.85} & 1.57 \\\hline
MiniImageNet   & 5.87 & 11.92  & \textbf{3.52} & 1.91 & 3.54 & 4.41 & \textbf{3.52}& \textbf{1.75} & 4.72 & 2.97 & 5.53 & 3.53&  4.16  &  2.18 \\\hline
5-Datasets   & 26.11 & 3.08 & 14.74 & 8.49 & 12.54 & 4.22 & \textbf{10.37}& \textbf{0.94}& 15.73 & 11.58 & 26.64 & 11.98&  10.47 & 1.37 \\\hline
F-EMNIST-1   & 1.73 & 2.45 & 1.08 & 1.06 & 2.16 & 3.62 & \textbf{0.44}& \textbf{0.94}& 1.37 & 1.93 & 1.56 & 2.14& 0.96 & 1.22\\\hline
F-EMNIST-2   & 1.96 & 2.45 & 3.06& 1.25 & 19.46 & 3.11 & \textbf{0.44} & \textbf{0.99}& 3.76 & 2.25 & 4.09 & 2.56& 0.70 & 1.62\\\hline
F-CelebA-1  & 0.85 & 2.67 & 0.14 & \textbf{0.82} & 0.18 & 3.25 & \textbf{0.08}& 2.13& 0.20 & 2.40 & 0.37 & 2.63&  0.14 & 2.28\\\hline
F-CelebA-2   & 0.93 & 2.67 & 0.17 & \textbf{0.87} & 0.74 & 3.78 &\textbf{ 0.09}& 3.07& 0.48 & 1.53 & 0.67 & 1.71&  0.44 & 3.81\\\hline
(EMNIST, F-EMNIST-1)  & 5.49 & 2.53 & 1.26 & \textbf{1.73} & 8.68 & 4.24 &\textbf{0.57} &1.98 & 2.16 & 2.71 & 3.35 & 3.26 &0.95 & 2.28\\\hline
(CIFAR 100, CelebA-1)  & 6.71 & 3.51 & \textbf{1.30}&\textbf{0.82}& 9.26 & 3.17 & 4.13 & 3.08& 2.35& 3.43& 3.69& 3.71&  5.13 & 3.19\\\hline
\textbf{Average} & 7.48 & 3.62 &  \textbf{3.59}&1.94&7.05&3.83 & 4.29 & \textbf{1.61}&4.40&3.33&6.12&3.64& 4.71& 2.19 \\\hline
\end{tabular}}
\begin{tablenotes}
        \scriptsize
        \item[1] T(S)--Time (Seconds); M(G)--Memory (GB). \\The bold numbers on each row indicate that they have the best performance values on the dataset corresponding to that row.
\end{tablenotes}}
\end{table}

\subsection{Model Scalability Comparisons}
\begin{figure}[h]
  \centering
  \includegraphics[width=0.8\textwidth]{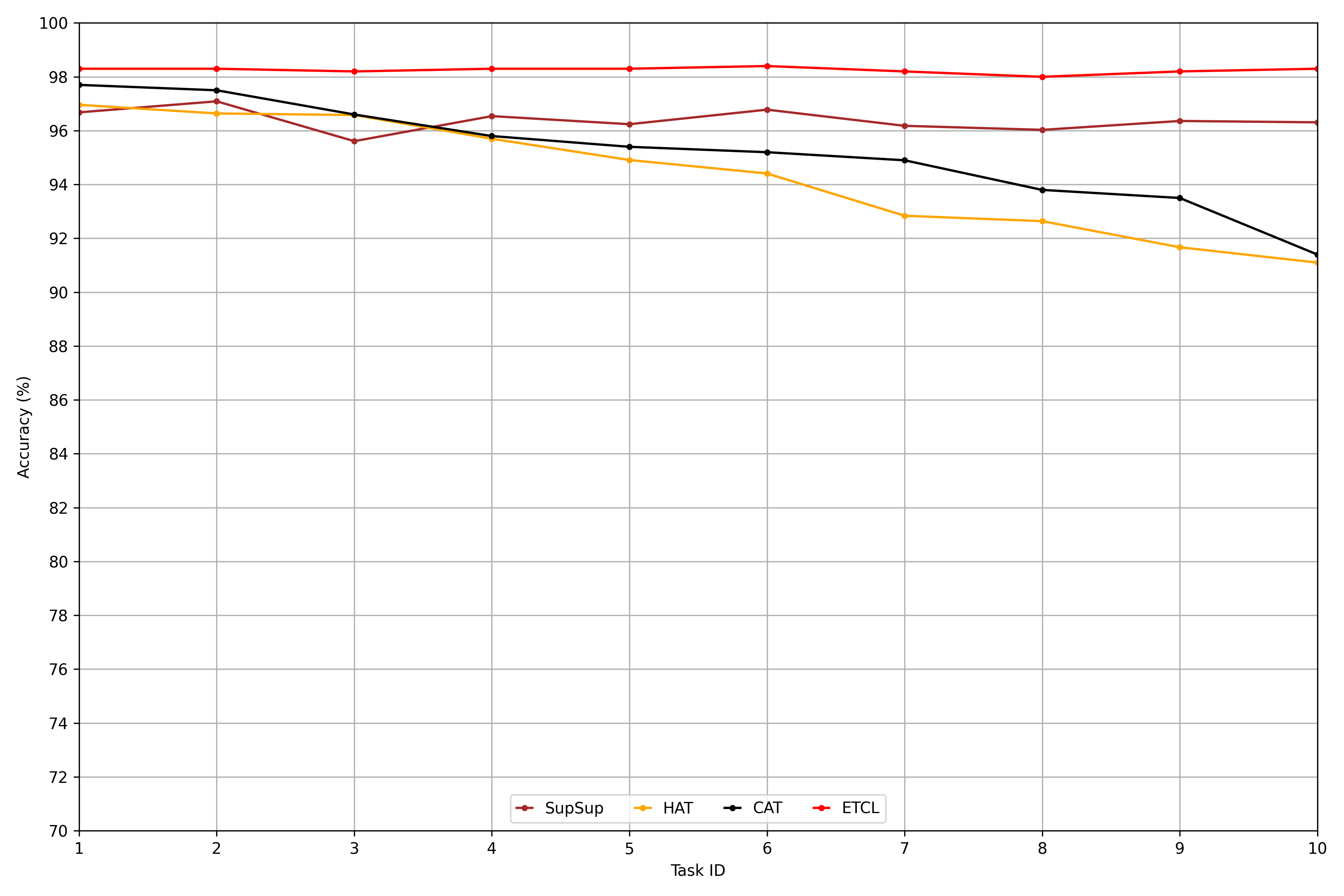}
  \caption{Scalability experimental results on dataset NPMNIST (10 Tasks).}
  \label{Sca1}
\end{figure}

\begin{figure}[h]
  \centering
  \includegraphics[width=0.8\textwidth]{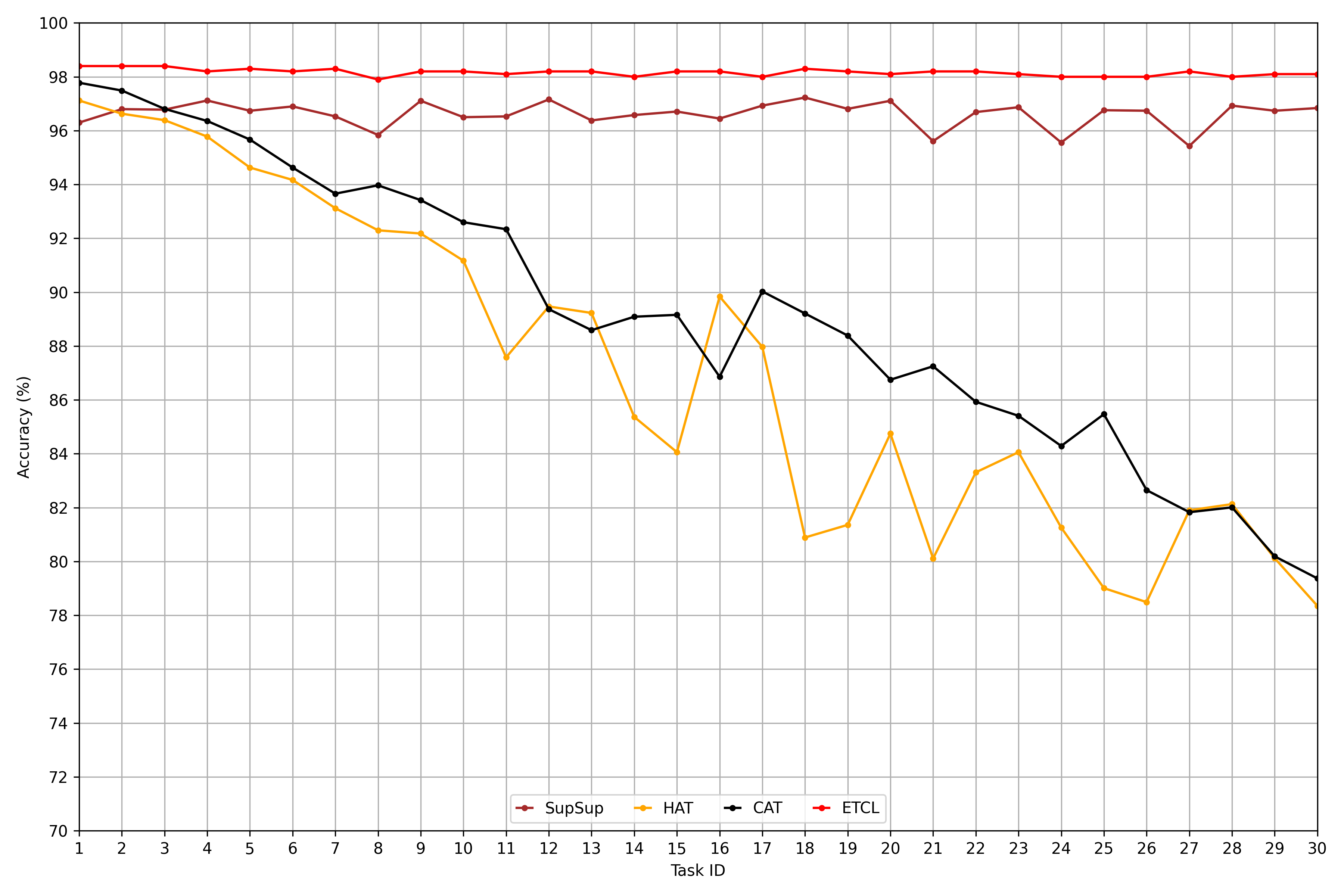}
  \caption{Scalability experimental results on dataset NPMNIST (30 Tasks).}
  \label{Sca2}
\end{figure}

\begin{figure}[h]
  \centering
  \includegraphics[width=0.8\textwidth]{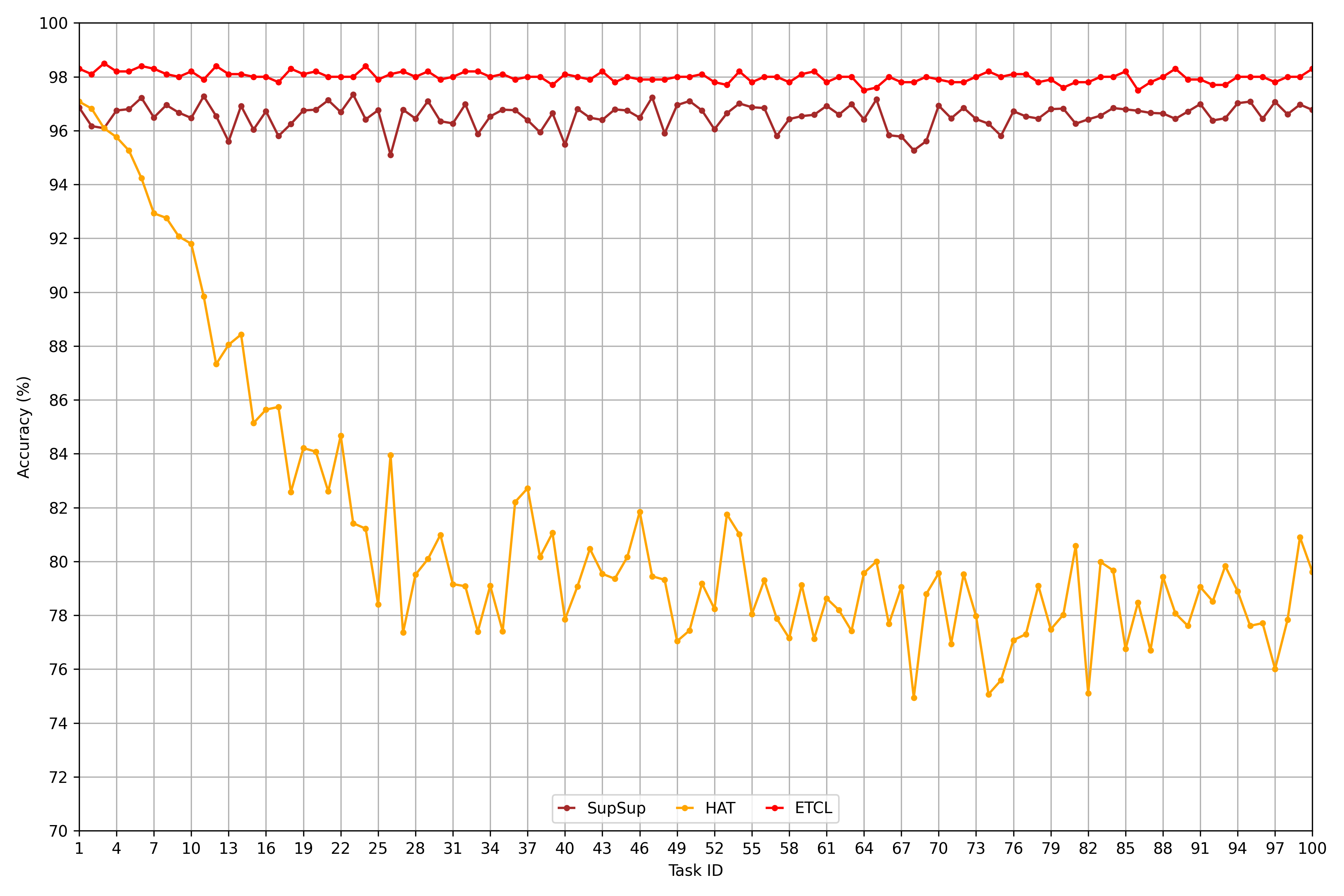}
  \caption{Scalability experimental results on dataset NPMNIST (100 Tasks).}
  \label{Sca3}
\end{figure}

\begin{figure}[h]
  \centering
  \includegraphics[width=0.8\textwidth]{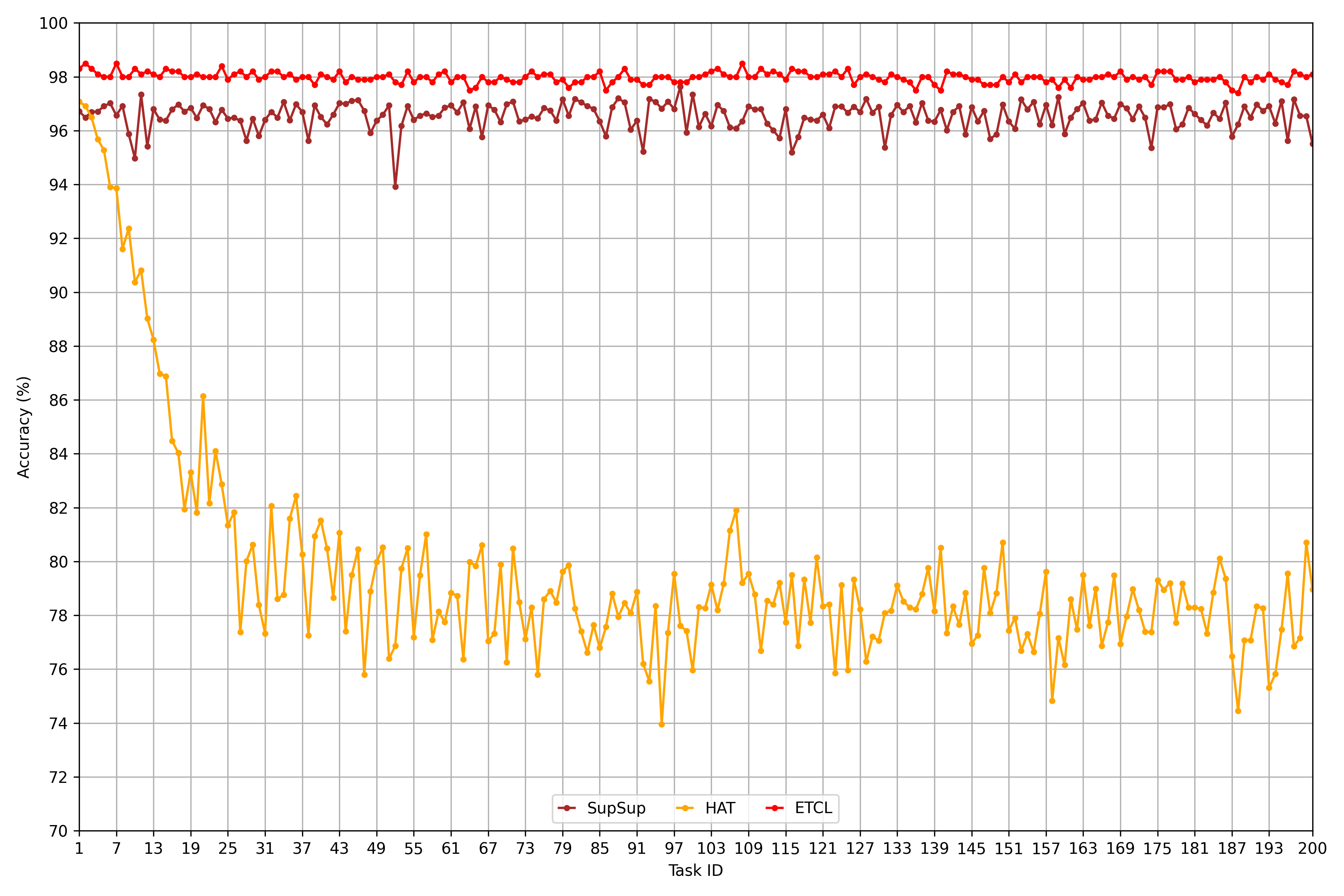}
  \caption{Scalability experimental results on dataset NPMNIST (200 Tasks).}
  \label{Sca4}
\end{figure}

Using the data enhancement technique to randomly shuffle the pixel on each image in the benchmark dataset PMNIST (10 Tasks), we constructed a set of the new datasets NPMNIST (New PMNIST) that has a total of 200 tasks. Then on the new dataset NPMNIST, we conducted a set of experiments to assess the model scalability with different numbers of tasks of the proposed ETCL and three SOTA mask-based baselines: CAT (with FKT and BKT mechanisms), HAT and SupSup (both HAT and SupSup have no explicit KT mechanisms). We use a 3-Layer FCN network with two hidden layers of 1000 units each on NPMNIST for all methods, where the hyperparameters of the model on NPMNIST are the same as the ones used on PMNIST (see Table \ref{ETCL-T4} for details). The experimental results of the model scalability comparisons are shown in Figures \ref{Sca1}-\ref{Sca4}.

Figures \ref{Sca1}-\ref{Sca4} demonstrate that with the increase in the number of tasks, our ETCL has the best ACC and scalability performances than the other three baselines. Although SupSup also has better model scalability than CAT and HAT, its average ACC performance of all tasks is not only lower than that of our ETCL but also its ACC on each task is undulating. CAT and HAT do have the drawback of poor model scalability. Due to the poor time performance of CAT, we were unable to run it with more tasks, e.g., 100 tasks and 200 tasks. Figures \ref{Sca1}-\ref{Sca4} clearly show that as the number of tasks increases, the scalability of the model HAT becomes worse and worse.

\end{document}